\documentclass[dvipsnames]{article} %
\usepackage{colm2024_conference}

\usepackage{booktabs}
\usepackage{enumitem}
\usepackage{wrapfig}
\usepackage{algorithm}
\usepackage{algpseudocode}
\usepackage{graphicx}
\graphicspath{{figures/}}
\usepackage[misc]{ifsym}
\usepackage{multicol}
\usepackage{amssymb}
\usepackage{microtype}
\usepackage{amsmath}
\usepackage{colortbl}
\usepackage[utf8]{inputenc} 
\usepackage[T1]{fontenc}
\definecolor{lightgray}{rgb}{0.9,0.9,0.9}
\usepackage{caption}
\usepackage{subcaption}
\usepackage{setspace}
\usepackage{url}
\usepackage{multirow}
\usepackage{tabularx}
\usepackage{blindtext}
\usepackage{pgfplots}
\pgfplotsset{compat=1.18} 
\usepackage{tikz}
\usetikzlibrary{er,positioning,bayesnet}
\usepackage{makecell}
\usepackage{tipa}
\usepackage{siunitx}
\usepackage{nicefrac}
\usepackage{listings}
\usepackage[raster,skins, most]{tcolorbox} %
\usepackage{xltabular}
\usepackage{adjustbox}
\usepackage{xurl}
\usepackage{rotating}
\usepackage[normalem]{ulem}
\usepackage{caption}    
\usepackage{subcaption} 
\usepackage[subfigure]{tocloft}
\usepackage{fontawesome}
\usepackage{arydshln}  
\usepackage{array}
\usepackage[table]{xcolor}

\useunder{\uline}{\ul}{}

\usepackage{amsmath,amsfonts,bm}
\usepackage{pifont}

\def\eqref#1{equation~\ref{#1}}

\def\1{\bm{1}}

\DeclareMathAlphabet{\mathsfit}{\encodingdefault}{\sfdefault}{m}{sl}
\SetMathAlphabet{\mathsfit}{bold}{\encodingdefault}{\sfdefault}{bx}{n}

\renewcommand{\ghlink}{tongyi-mai.github.io/Qwen-UI-Agent}

\newcommand*\justify{%
  \fontdimen2\font=0.4em%
  \fontdimen3\font=0.2em%
  \fontdimen4\font=0.1em%
  \fontdimen7\font=0.1em%
  \hyphenchar\font=`\-%
}

\renewcommand{\texttt}[1]{%
  \begingroup
  \ttfamily
  \begingroup\lccode`~=`/\lowercase{\endgroup\def~}{/\discretionary{}{}{}}%
  \begingroup\lccode`~=`[\lowercase{\endgroup\def~}{[\discretionary{}{}{}}%
  \begingroup\lccode`~=`.\lowercase{\endgroup\def~}{.\discretionary{}{}{}}%
  \catcode`/=\active\catcode`[=\active\catcode`.=\active
  \justify\scantokens{#1\noexpand}%
  \endgroup
}

\usepackage{makecell}
\usetikzlibrary{tikzmark}
\makeatletter
\newcommand*\myfontsize{%
  \@setfontsize\myfontsize{7}{8}%
}
\makeatother

\definecolor{lightblue}{RGB}{173,216,230}
\definecolor{uclablue}{RGB}{159, 195, 224}

\definecolor{uclagold}{RGB}{255, 240, 180}

\definecolor{aliceblue}{RGB}{255, 238, 241}

\definecolor{cadmiumgreen}{rgb}{0.0, 0.42, 0.24}

\definecolor{myred}{rgb}{0.7, 0.3, 0.0}
\definecolor{myblue}{rgb}{0.2, 0.3, 0.6}
\definecolor{babygreen}{rgb}{0.85, 0.97, 0.85}

\definecolor{purple1}{RGB}{126, 107, 196}
\definecolor{purple2}{RGB}{199, 158, 207}
\definecolor{purple3}{RGB}{214, 200, 255}
\definecolor{purple4}{RGB}{254, 240, 255}

\definecolor{deepblue}{RGB}{48, 58, 82}
\definecolor{light_purple}{RGB}{228, 228, 251}
\definecolor{deep_purple}{RGB}{234, 224, 240}
\definecolor{deep_purple2}{RGB}{102, 48, 226}

\newcommand{\symboletongyi}{\raisebox{0pt}{~\includegraphics[scale=0.012]{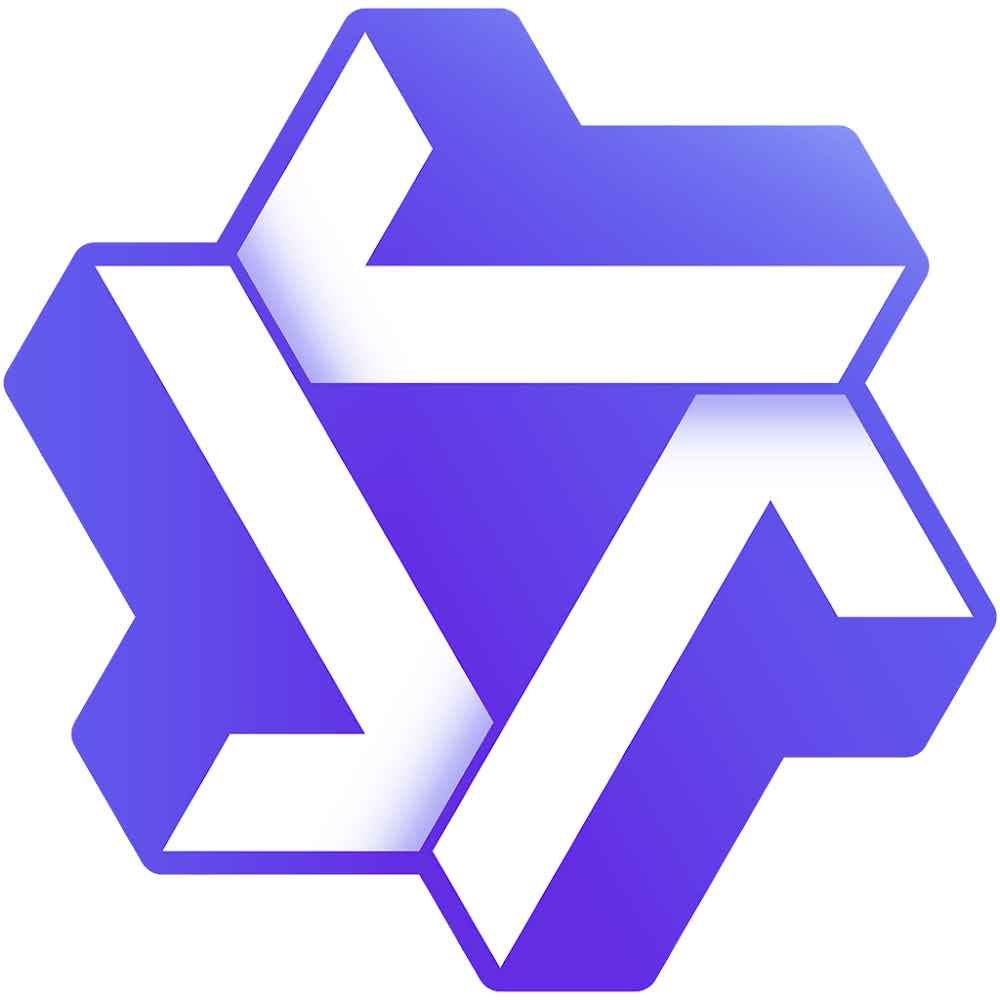}}~}

\definecolor{deepPurple}{HTML}{330066}

\definecolor{uclablue_old}{rgb}{0.15, 0.45, 0.68}
\hypersetup{
    breaklinks,
    citecolor=uclablue_old,
    colorlinks=true,
}
\usepackage[most]{tcolorbox} 
\newtcbtheorem{promptbox}{Prompt}{
    breakable,
    colframe=deepPurple!70,
}{box}

\newtcolorbox{mybox}[2][]
  {colback = black!5!white, colframe = black!75!black, fonttitle = \bfseries,
    colbacktitle = black!100!black, enhanced, before upper={\fontsize{8}{11}\obeyspaces\obeylines\selectfont}, fontupper=\selectfont,
    attach boxed title to top left={yshift=-2.2mm,xshift=4mm},
    title=#2,#1}

\newcommand{\modelname}{Qwen-UI-Agent}
\title{
    \modelname{} Technical Report: Toward Next-Generation Real-World Centric Foundation GUI Agents
}
\author{%
{MAI-UI Team\thanks{Qwen-UI-Agent is a continuation of our previous work, MAI-UI~\citep{maiui}. See \hyperref[contribution]{Contributions} section for a full author list.}}
  \\[0.35em]
  {\fontsize{10pt}{11pt}\selectfont
Alibaba Token Hub\symboletongyi, Alibaba Group}
}

\definecolor{LeiColor}{HTML}{E85D45}

\begin{document}

\maketitle

\begin{abstract}
\vspace{-0.3em}
GUI agents have the potential to become a general purpose executor over existing digital devices. To advance them toward real-world use, we envision agents that operate reliably on real devices, execute workflows across platforms, combine GUI interaction with CLI execution, complete long-horizon tasks, proactively initiate useful services, and autonomously improve their capabilities with minimal human effort. Guided by this vision, we present \textbf{\modelname{}}, a real-world centric foundation GUI agent spanning mobile, computer-use, web, and DeepSearch environments. \modelname{} combines diverse sandbox environments with a large-scale real-device mobile runtime. Its unified action space interleaves GUI operations with CLI execution and generates batched actions in a single model turn. An AutoResearch-style data flywheel uses agents to construct tasks and environments, diagnose failures, and plan subsequent iterations. Online RL support training on trajectories exceeding 100 turns, with over 10,000 concurrent environments accelerating rollout.  A lightweight harness layer supports proactive service initiation and stateful workflows across mobile and computer. \\ 
\vspace{+0.3em}
Across a broad suite of evaluations, \modelname{} sets state-of-the-art performance on mobile-use benchmarks while delivering competitive performance on computer- and browser-use  tasks against frontier models, including Opus 4.8, Gemini 3.1 Pro, and GPT-5.6 Sol. On mobile use, it achieves \textbf{82.1\%} on MobileWorld, \textbf{92.2\%} on MobileWorld-Real, and \textbf{97.5\%} on AndroidDaily. On computer use, it achieves \textbf{79.5\%} on OSWorld-Verified and a \textbf{40.0\%} partial-progress score on OSWorld-v2. On browser use and GUI grounding, it achieves \textbf{73.6\%} on WebArena and \textbf{81.5\%} on ScreenSpot-Pro, respectively.

\end{abstract}

\begin{figure}[H]
\vspace{-1.1em}
    \centering  \includegraphics[width=0.92\linewidth]{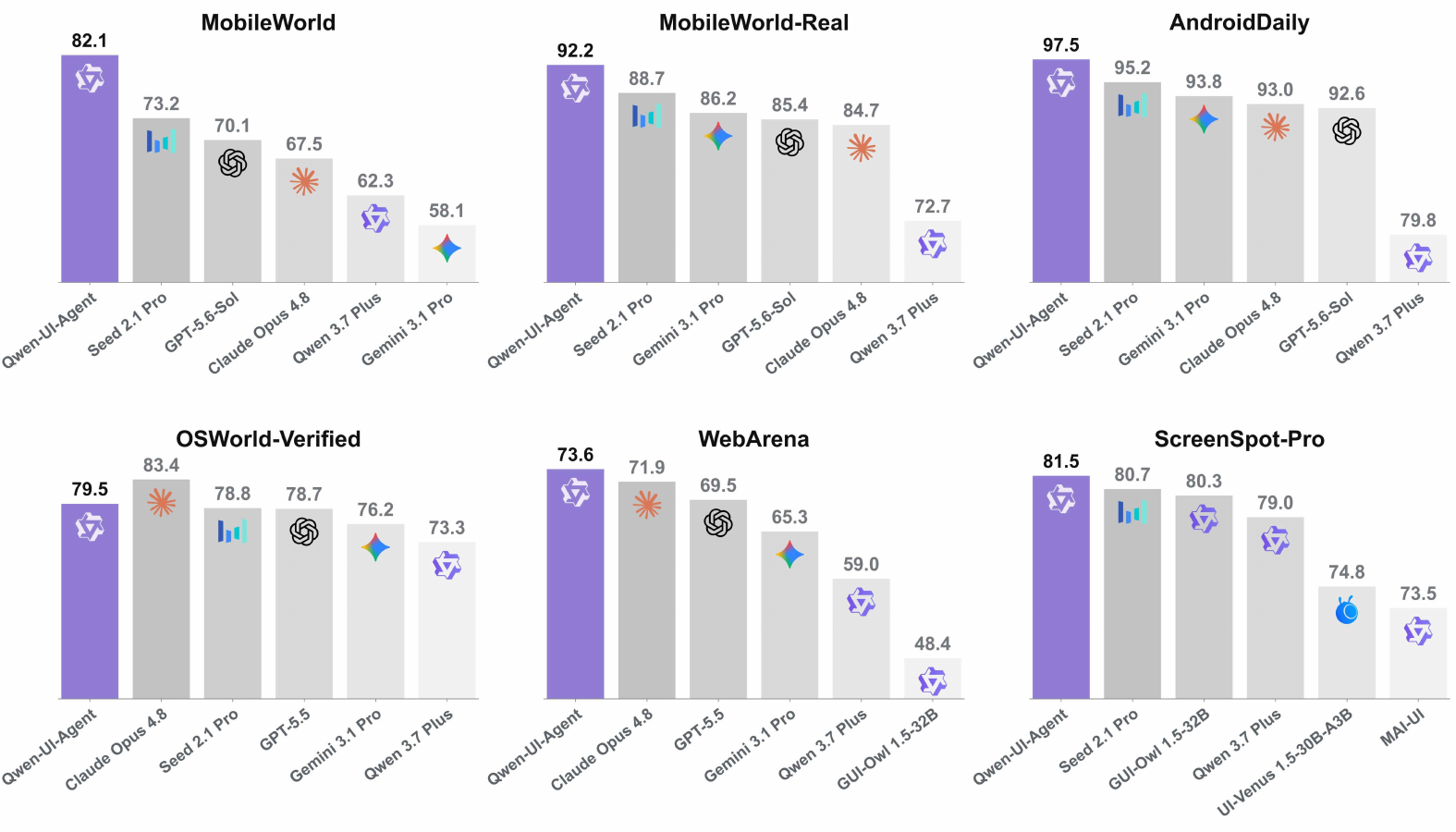}
    \vspace{-1.1em}
    \caption{\modelname{} demonstrates leading or competitive performance across diverse GUI settings.}
    \label{fig:teaser}
\end{figure}

\renewcommand{\thefootnote}{*}
\renewcommand{\thefootnote}{\arabic{footnote}}

\clearpage
\tableofcontents
\clearpage
\section{Introduction}
\vspace{-0.5em}
Graphical user interfaces (GUIs) are the primary interface through which humans access digital services and applications. 
A capable GUI agent \citep{maiui, uitars2, uitars, Mobile-agent-v3.5} therefore has the potential to become a general purpose executor over existing application ecosystems: it can understand user intent, perceive application states, operate interfaces, and complete tasks without requiring every service to expose a dedicated API. 
Recent progress in multimodal foundation models has made this direction increasingly practical, leading to rapid improvements in mobile and desktop GUI agents \citep{cao2026xiaomi, uivenus15, evocua15, hymobileagent}.

Despite this progress, GUI agents remain largely optimized for simulated benchmarks, leaving a substantial gap between benchmark performance and real-world utility. To bridge this gap, we envision a next generation of GUI agents that operate reliably on real devices, complete long-horizon tasks, carry workflows across platforms, proactively initiate useful services, combine GUI interaction with CLI execution, and continuously evolve their capabilities with less human effort. To realize this vision, we identify six key transitions:
\textbf{(1) From simulated environments to real-device execution.}
Many GUI agents, especially mobile agents, are optimized for simulated benchmarks, while real mobile devices involve much more complex application functions, varied UI layouts, interruptions, permissions, and changing environments. 
A model that performs well in simulation often fails on real devices. 
\textbf{(2) From isolated domains to cross-domain, cross-platform workflows.}
Existing agents are often built for mobile, web, or desktop separately, while real-world tasks may span several domains and continue across phones and computers. 
\textbf{(3) From GUI-only actions to hybrid GUI+CLI and batched actions.}
 GUI interaction provides universal access to visual interfaces, while the growing coding capabilities of foundation models make CLI a natural complement for structured tasks such as data processing and file manipulation. 
By combining these action types and supporting batched actions, agents can better match the execution method to the nature of tasks, improving execution efficiency and expanding task coverage. 
\textbf{(4) From short-horizon tasks to reliable long-horizon task completion.} 
The practical value of a GUI agent grows with the complexity of the user objectives it can complete and the number of execution steps it can perform on the user's behalf.
However, completing long-horizon tasks requires sustained planning, state tracking, intermediate verification, and recovery from execution errors, making long-horizon reliability a challenging capability.
\textbf{(5) From human-intensive training pipelines to AutoResearch-style one.}
Current GUI agent development still relies heavily on manually constructing tasks, collecting data, training models, running evaluations, and analyzing failures. As agent capabilities improve, this process should become increasingly automated.
\textbf{(6) From reactive execution to proactive service initiation.}
Most agents wait for explicit user instructions, even though many useful tasks can arise directly from digital signals on phones or other devices. An agent that recognizes such actionable moments and proposes appropriate next steps can provide a fundamentally different user experience.

In this report, we introduce \modelname{}, a real-world-centric foundation GUI agent developed around these five transitions, representing our step toward next-generation GUI agents for real-world utility.
Specifically, \modelname{} delivers the following core features:
\begin{itemize}[leftmargin=*]
\vspace{-0.5em}
    \item \textbf{Real-device Mobile GUI Foundation}. We find that the simulation-to-real gap is particularly severe in mobile environments. 
    To address this gap, we build a real-device mobile environment comprising over 100 physical devices and supporting more than 150 applications. We use it for realistic task design, trajectory collection, online reinforcement learning, and evaluation on real devices. To improve practical usability, we further support user takeover for high-risk actions, expand coverage to both high-frequency workflows and less commonly used app functions. 
    Together, these efforts enable \modelname{} to substantially outperform existing models on real phones.
    \item \textbf{Multi-domain Capabilities and Cross-Domain Workflows.} \modelname{} supports task execution across mobile, web, computer-use, and DeepSearch environments. Rather than treating these capabilities as isolated modules, our harness layer composes them within a unified workflow, preserving context and task state as execution moves seamlessly across devices and environments. 
    For example, GUI workflows can invoke DeepSearch to efficiently retrieve and verify external information, avoiding time-consuming GUI-based web navigation before seamlessly resuming interaction.
    This turns multi-domain coverage into integrated task execution across heterogeneous digital environments.
    \item \textbf{Hybrid GUI+CLI and Batched Action Spaces.} 
    We equip \modelname{} with a unified action space that supports GUI operations, bash-based CLI commands, and batched actions. The model can select and compose these actions within a single trajectory, using GUI interaction and CLI execution where each is most suitable, while batching compatible actions to reduce unnecessary steps. In computer-use tasks, CLI commands and GUI clicks emerge as the two dominant action types, and over 40\% of action outputs are batched, substantially shortening execution trajectories.
    \item \textbf{Scalable Online RL for Long-horizon Tasks.} Hybrid GUI+CLI and batched actions improve local execution efficiency, but completing long-horizon tasks also requires reliable planning, state tracking, and intermediate decision-making across the full trajectory. We therefore scale verifier-guided online RL to trajectories exceeding 100 interaction steps. We further enable approximately 10,000 simulated environments to run concurrently to accelerate rollout generation while a model-adaptive curriculum prioritizes tasks with intermediate success rates and replaces mastered tasks with harder ones.
    \item \textbf{AutoResearch-style Capability Acquisition.} We build an agent-driven iteration loop in which agents generate and assess candidate tasks for data flywheel, process rollout trajectories, analyze failures, and propose plans for the next iteration. Rather than relying on humans to manually drive every stage, the system automates much of the capability development process, with humans mainly providing supervision and making targeted revisions. This approach reduces human effort and accelerates the acquisition and refinement of new GUI agent capabilities.
    \item \textbf{Proactive Mobile Service.} Consider a flight cancellation notification arriving on a user’s phone. Instead of waiting for the user to notice it and issue an instruction, the agent can identify the affected itinerary, gather alternative travel options, check against user’s schedule, and present a proposed plan for confirmation. To support such experiences, we build a proactive service harness based on mobile notifications.
    Beyond initiating tasks, the harness forms a coherent understanding of real-world situations, generates personalized tasks, executes approved actions, and continuously evolves from user responses and long-term interaction history. 
\end{itemize}
\vspace{-0.5em}
Empirically, \modelname{} achieves state-of-the-art performance across a broad range of evaluation settings, covering real-device and benchmark-based mobile use, computer use, browser use, DeepSearch, and GUI grounding. Across these evaluations, \modelname{} generally outperforms leading foundation models, including Opus 4.8 \citep{anthropic2026claudeopus48}, GPT-5.6 Sol \citep{openai2026gpt56}, Gemini 3.5 Flash \citep{google2026gemini35flash}, and Seed2.1 Pro \citep{bytedanceseed2026seed21}, as well as specialized GUI agents. We highlight main results below:
\begin{itemize}[leftmargin=*]
\vspace{-0.5em}
\item \textbf{Mobile Use.} On MobileWorld-Real, our real-device benchmark comprising more than 400 tasks across over 100 apps, \modelname{} achieves a \textbf{92.2\%} success rate. It outperforms frontier closed-source models Gemini 3.1 Pro, Claude Opus 4.8, GPT-5.6 Sol, and Seed 2.1 Pro by {6.0, 7.5, 6.8, and 3.5} percentage points, respectively.
\modelname{} further achieves a near-perfect success rate of \textbf{97.5\%} on AndroidDaily, another real-device benchmark.
On MobileWorld, \modelname{} scores \textbf{82.1\%}, surpassing Opus 4.8, GPT-5.6 Sol, and Seed 2.1 Pro by 14.6, 12.0, and 8.9 percentage points.
\item \textbf{Computer Use.} On OSWorld-Verified, \modelname{} achieves \textbf{79.5\%}, ranking second overall and outperforming GPT-5.5, Gemini 3.1 Pro, and Seed 2.1 Pro. On the more challenging OSWorld-v2, \modelname{} obtains the second-best binary success rate (\textbf{13.9\%}), the third-best partial-progress score (\textbf{40.0\%}). Its partial-progress score exceeds those of MiniMax M3 and Qwen 3.7 Plus by \textbf{17.7} and \textbf{18.5} percentage points, respectively, while requiring \textbf{58.4\%} and \textbf{21.7\%} fewer steps per task.

\item \textbf{Browser Use and DeepSearch.} On WebArena, \modelname{} achieves {\textbf{73.6\%}}, outperforming Claude Opus 4.8, GPT-5.5 and Gemini 3.1 Pro by 1.7, 4.1 and 8.3 points. On DeepSearch, it scores {\textbf{64.1\%}} on BrowseComp and {\textbf{75.0\%}} on BrowseComp-ZH, surpassing Qwen3.5-397B-A17B and UI-TARS-2.

\item \textbf{GUI Grounding.} \modelname{} reaches {\textbf{81.5\%}} on ScreenSpot-Pro under the zoom-in setting, outperforming Seed 2.1 Pro, GUI-Owl-1.5, and UI-Venus-1.5. It further sets the best results among baseline models on {UI-Vision ({70.0\%}), OSWorld-G-Refined (78.5\%), MMBench-GUI L2 ({92.6\%}), and ScreenSpot-V2 ({97.5\%}).}

\item \textbf{General and Agentic Capabilities.} To preserve the model's broad utility in real-world settings, we retain \modelname{}'s general reasoning and agentic capabilities. We evaluate \modelname{} on a suite of 13 benchmarks, including MMMU-Pro, MMLU-Pro, Terminal-Bench 2.0, and Claw-Eval. 
The results show that \modelname{}  outperforms the Qwen base model on agentic tasks while remaining comparable on general reasoning tasks, and significantly outperforms GUI-specialized models across both categories.

\end{itemize}

\section{\modelname{}}
\begin{figure*}[t]
    \centering
    \includegraphics[width=\textwidth]{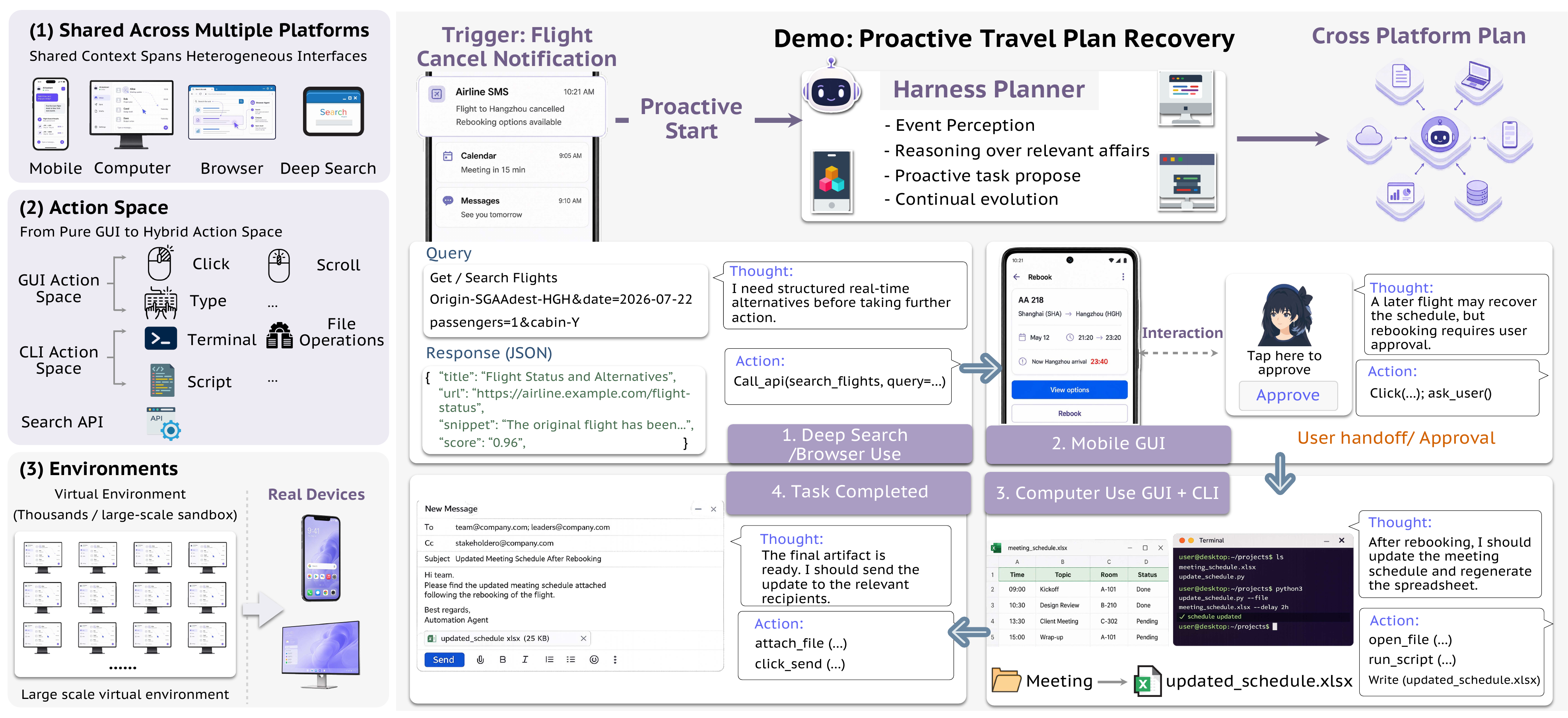}
    \caption{
    An illustrative trajectory of \modelname{} for proactive cross-platform task execution. The left panel summarizes the system capabilities underlying the trajectory, including cross-platform execution, a hybrid action space spanning GUI, CLI, and API operations, and environments ranging from large-scale sandboxes to real devices. The right panel illustrates a travel-recovery scenario triggered by a flight-cancellation notification. After identifying the affected tasks and commitments, the agent searches for alternative flights through an API, requests user approval before rebooking via a mobile GUI, updates the affected meeting schedule through desktop GUI and CLI actions, and sends the revised artifact to the relevant recipients.
    }
    \label{fig:demo_traj}
    \vspace{-1em}
\end{figure*}

\vspace{-0.5em}
This section presents the methodology of \modelname{} as an integrated system for building advanced foundation GUI agents. The system comprises four components: (1) an environment infrastructure spanning both sandboxed and real-world environments that supports trajectory collection and training across mobile-use, computer-use, browser-use, and deep-search tasks under a unified action space covering GUI, CLI, and batched actions; (2) an automated data flywheel that reduces human intervention and closes the loop from data synthesis to training, evaluation, failure analysis, and targeted data generation for the next iteration; (3) a unified training framework that combines supervised fine-tuning with reinforcement learning; and (4) a harness layer that connects the trained agent to user context and enables proactive services and cross-platform task execution. Figure \ref{fig:demo_traj} illustrates a demo trajectory.%
\subsection{System Overview}
\label{sec:system_overview}

\modelname{} is designed as a real-world-centric foundation GUI agent that operates across diverse digital environments, including mobile, desktop, web, and information-seeking systems. This section first defines the task scope of \modelname{}, then introduces its unified observation and action spaces, and finally summarizes the capability landscape supported by this formulation.

\subsubsection{Task Formulation}

\modelname{} is designed for interactive digital task execution. We define a task as:
\begin{equation}
    \tau = \left(I, \mathcal{E}_{\tau}\right),
\end{equation}
where $I$ is the user instruction and $\mathcal{E}_{\tau}$ is the set of
digital environments available to the task. Depending on the task,
$\mathcal{E}_{\tau}$ may include mobile devices, web browsers,
computer systems, DeepSearch systems, or a
combination of them. This formulation covers both tasks completed
within a single environment and workflows that span multiple platforms~\citep{Mobile-agent-v3.5, uimopd2026}.

At decision step $t$, the agent receives a multichannel observation
$o_t$ that may combine different forms of environment feedback:
\begin{equation}
    o_t =
    \left(
    o_t^{\mathrm{GUI}},
    o_t^{\mathrm{CLI}},
    o_t^{\mathrm{API}}
    \right).
\end{equation}
Here, $o_t^{\mathrm{GUI}}$ represents the current screenshot, $o_t^{\mathrm{CLI}}$ contains command execution results, and $o_t^{\mathrm{API}}$ contains structured responses from external services, such as search API. A Component is set to empty when the corresponding channel is unavailable or unnecessary for the current task. This multichannel observation allows the agent to jointly reason over visual interface states and programmatic execution feedback~\citep{toolcua2026, phoneharness2026}.

The agent predicts an intermediate reasoning output $r_t$ and an
executable action output $\mathbf{a}_t$:
\begin{equation}
    \left(r_t,\mathbf{a}_t\right)
    =
    \pi_{\theta}
    \left(
    I,o_t,h_t
    \right),
\end{equation}
where
$
    h_t =
    \left(
    o_1,r_1,\mathbf{a}_1,
    \ldots,
    o_{t-1},r_{t-1},\mathbf{a}_{t-1}
    \right)
$
denotes the preceding interaction history.

Notably, one model decision step may produce either a single action or
an ordered sequence of actions:
\begin{equation}
    \mathbf{a}_t =
    \left(
    a_t^{(1)},\ldots,a_t^{(K_t)}
    \right),
    \qquad
    a_t^{(k)} \in \mathcal{A}_t,
\end{equation}
where $\mathcal{A}_t$ is the set of actions available in the current
environment. When $K_t=1$, the model performs single-action execution.
When $K_t>1$, it produces a batched action sequence. Actions within a
batch are executed consecutively in this decision step,
reducing unnecessary inference and observation steps when multiple
operations can be completed without additional environment feedback.

\subsubsection{Action Space}
\label{sec:action-space}
We design the action space of \modelname{} around three requirements: supporting GUI interaction across different platforms, extending execution beyond GUI manipulation, and preserving user control over consequential operations. The GUI action space is defined as the union of actions necessary for mobile, web, and desktop environments, with each environment exposing the subset supported by its native interaction mechanisms. To complement GUI interaction, we introduce \texttt{cli\_command}, which enables direct bash command execution, and \texttt{api\_call}, which invokes external services with structured arguments. We further include \texttt{ask\_user}, allowing the agent to request missing information or obtain explicit confirmation before handling sensitive data, payments, or other consequential operations. Together, these actions provide a unified interface for cross-platform GUI interaction, hybrid GUI+CLI execution, and user-controlled operation in real-world tasks. The full action space is shown in Table~\ref{tab:action-space}.

\begin{table}[t]
    \centering
    \caption{Action Space in \modelname{}.}
    \label{tab:action-space}
    \small
    \setlength{\tabcolsep}{5pt}
    \begin{tabular}{@{}p{0.25\linewidth}p{0.67\linewidth}@{}}
        \toprule
        Action & Definition \\
        \midrule
        \multicolumn{2}{@{}l}{\textbf{GUI Actions}} \\
        \texttt{click} & Clicks at coordinates $(x,y)$. \\
        \texttt{double\_click} & Double-clicks at coordinates $(x,y)$. \\
        \texttt{long\_press} & Long-presses at coordinates $(x,y)$. \\
        \texttt{type} & Types the specified text content. \\
        \texttt{open} & Opens the specified app.  \\
        \texttt{drag} & Drags from start coordinates $(x_1,y_1)$ to end coordinates $(x_2,y_2)$. \\
        \texttt{system\_button} & Presses a system button, selected from \texttt{back}, \texttt{home}, \texttt{menu}, and \texttt{enter}. \\
        \texttt{wait} & Waits for a specified duration in seconds. \\
        \midrule
        \multicolumn{2}{@{}l}{\textbf{CLI Actions}} \\
        \texttt{cli\_command} & Executes a bash command in the active CLI environment. \\
        \midrule
        \multicolumn{2}{@{}l}{\textbf{API Actions}} \\
        \texttt{api\_call} & Invokes an API with specified arguments. \\
        \midrule
        \multicolumn{2}{@{}l}{\textbf{Interaction and Control Actions}} \\
        \texttt{ask\_user} & Interact with the user to complete the task. \\
        \texttt{terminate} & Ends the task with a final answer and a status of \texttt{success} or \texttt{failed}. \\
        \bottomrule
    \end{tabular}
\end{table}

\subsubsection{Capability Landscape}
\label{sec:capability_landscape}

The unified formulation supports four complementary capabilities of \modelname{} centered on real-world task execution. (1) \textbf{Real-device mobile execution.} By building a real-device mobile environment and using it throughout model capability development and evaluation, \modelname{} substantially eliminates the simulation-to-reality gap and achieves leading real-world mobile execution performance over both specialized GUI agents and frontier closed-source models. (2) \textbf{Multi-domain task execution.} \modelname{} operates across mobile, desktop, web, and DeepSearch environments, moving beyond domain-specific policies toward a foundation GUI agent for heterogeneous digital tasks. (3) \textbf{Hybrid and efficient interaction.} \modelname{}'s unified action space enables the model to select and interleave GUI operations with bash-based CLI commands within the same trajectory, while batched actions largely enhance the execution efficiency. (4) \textbf{Harness-enabled services.} A lightweight harness layer extends the core agent beyond isolated task execution by enabling proactive service initiation and allowing a single workflow to span mobile and computer systems.

\subsection{Environment Infrastructure}

\begin{figure*}[t]
    \centering
    \includegraphics[width=\textwidth]{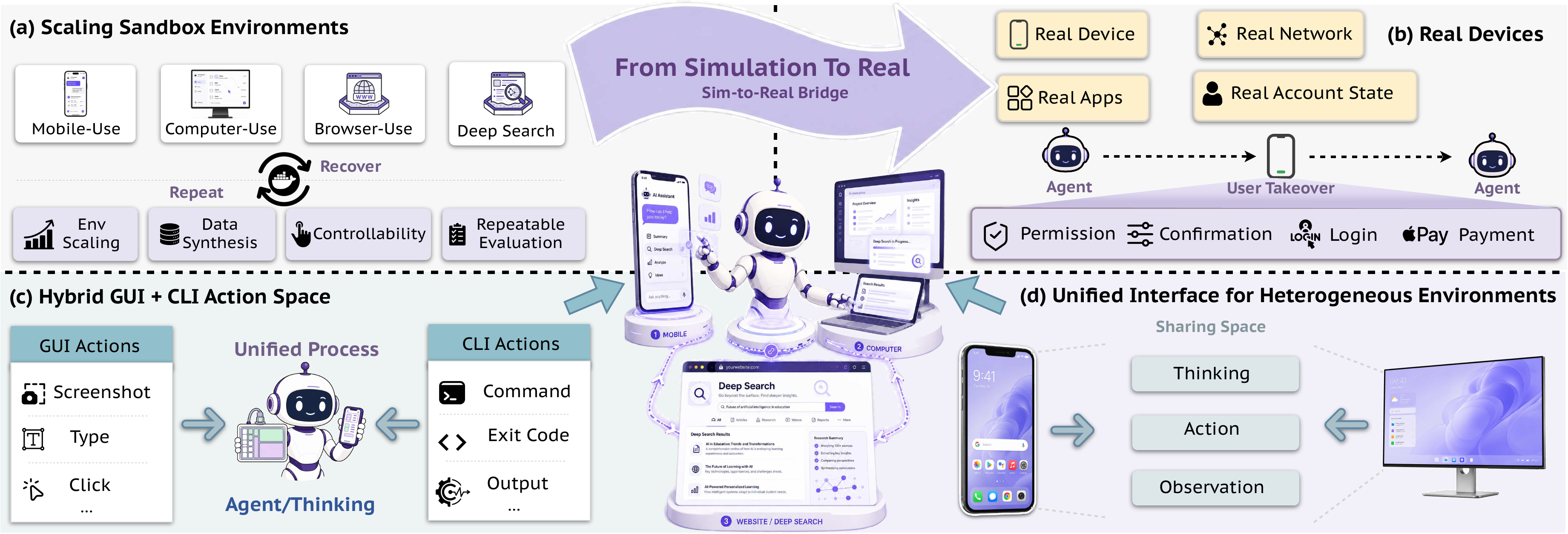}
    \caption{The environment infrastructure of \modelname{}. (a) Scalable sandbox environments spanning mobile-use, computer-use, browser-use, and DeepSearch provide controllability and repeatable evaluation for data synthesis and training. (b) A sim-to-real bridge extends agents to real devices with real applications, networks, and account states, supporting user takeover for login, payment, permission, and confirmation. (c) A hybrid GUI+CLI action space interleaves graphical operations with direct command execution. (d) A unified interface standardizes the thinking--action--observation loop across heterogeneous environments.}
    \label{fig:environment-infrastructure}
\end{figure*}

The environment defines the capability boundary of an agent: it determines what the agent can perceive, what actions it can take, and what experience it can acquire through interaction. 
Therefore, moving GUI agents from benchmark-centric optimization toward real-world utility requires not only stronger models, but also broader and more realistic environments in which agents can learn, act, and adapt. 
As illustrated in Figure~\ref{fig:environment-infrastructure}, our environment infrastructure is organized around four core components. 
First, \textbf{simulated and sandbox environments} provide scalable, controllable, and reproducible settings for data generation, agent learning, and evaluation. 
Second, extending these simulated environments to \textbf{real devices} enables agents to learn and operate under realistic deployment conditions, including dynamic device states, system constraints, and user interaction. 
Third, a \textbf{hybrid GUI and CLI action space} supports more efficient and adaptive execution by allowing the agent to select the most appropriate interaction modality for each task. 
Finally, a \textbf{unified cross-platform environment} provides a unified pipeline across mobile-use, computer-use, browser-use, and DeepSearch, while preserving context and task state across platforms.
Together, these components make the environment a core part of the agent system, directly shaping the range of tasks the agent can learn and accomplish.

\subsubsection{Scaling Sandbox Environments}
\label{sec:scaling-sandbox-environments}
Sandbox environments provide controlled and resettable task states for training, evaluation, and ablation. Our suite spans mobile-use, computer-use, browser-use, and DeepSearch tasks. Each domain retains its native applications, environment-specific context, and state-based verification mechanisms. A key strength of our infrastructure is its scalability: it supports up to \textbf{10,000} isolated sandbox environments running in parallel, enabling large-scale rollout and agent training.
\vspace{-0.5em}
\begin{itemize}[leftmargin=*]
    \item \textbf{Mobile Use.} We build our mobile sandbox on the environment provided by MobileWorld, which offers a resettable Android system with 20 applications and supports GUI task execution, agent-user interaction, and MCP tool use \citep{kong2026mobileworld}. To address the scaling bottleneck of KVM-based emulators, we rebuilt the MobileWorld environment on redroid~\citep{redroid}, which runs Android as a container on the host kernel without QEMU or nested KVM. This design enables complete device-and-backend environments to be replicated across ordinary container hosts. Its containerized device instances and self-hosted application backends provide controllable initial states and reproducible state transitions. Building on this infrastructure, we construct diverse task-relevant environment contexts, synthesize tasks, and generate state-based verifiers. The resulting sandbox provides a stable and scalable environment for trajectory collection and agent training.

\item \textbf{Computer Use.} We build our computer-use sandbox on the Ubuntu virtual-machine environment provided by OSWorld, which supports real desktop applications, operating-system operations, and cross-application workflows~\citep{OSWorld}. While the standard OSWorld interaction interface is GUI-based, we extend it with direct bash execution, allowing the agent to interleave GUI operations and CLI commands within the same trajectory. For our newly constructed tasks, we further create task-specific initial environment states and state-based verifiers for some of these tasks. Together, these extensions provide a controllable and scalable environment for collecting trajectories and training agents under both GUI-only and hybrid GUI+CLI interaction.

 \item \textbf{Browser Use.} Our web environment is a self-contained browser runtime implemented with FastAPI, Playwright~\citep{playwright}, and Chromium. Each episode runs in a fresh Playwright \texttt{BrowserContext} with isolated cookies, cache, and local storage, and begins from a task-specific page and state. 
 The agent receives rendered screenshots and interacts through browser-native GUI actions. The JavaScript-based verifiers inspect the DOM and persistent application state, allowing alternative action sequences to receive credit when they reach the required state.

 \item \textbf{DeepSearch.} The DeepSearch environment complements interactive browser control with structured information-access tools. Serper~\citep{serper} provides ranked search results, while Jina Reader~\citep{jinareader} converts selected web pages and documents into model-consumable text. The agent can iteratively reformulate queries, select sources, retrieve full content, and synthesize evidence across pages. This environment therefore targets open-ended, multi-source information seeking in which the principal observations are search results and retrieved documents rather than rendered GUI states.
\end{itemize}

\subsubsection{Real-Device Mobile Environment: From Simulated Sandboxes to Real-World Learning}
\label{sec:real-device-mobile}

 \begin{figure}[t]
    \centering
    \includegraphics[width=\linewidth]{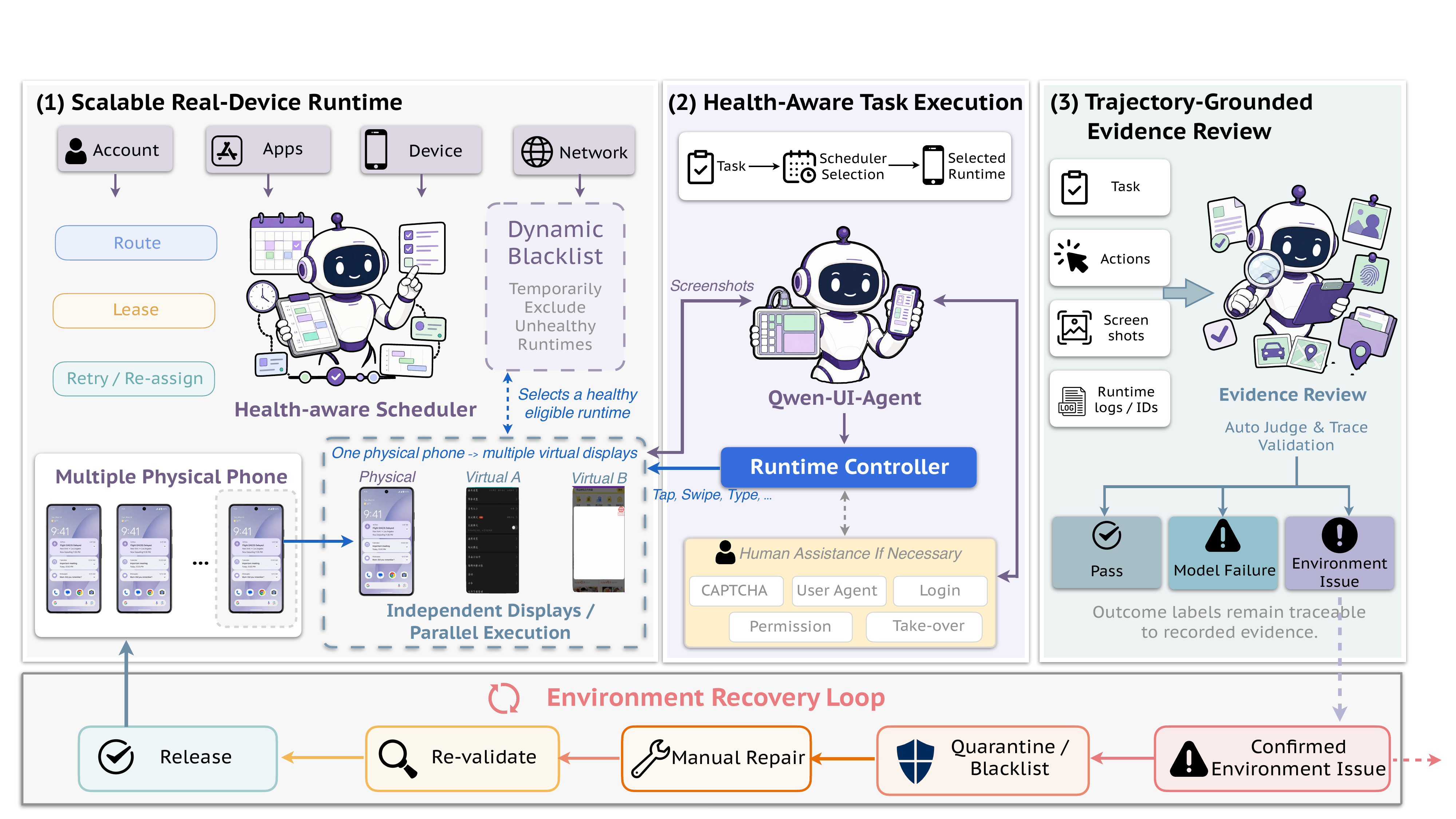}
    \caption{Real-device mobile runtime with closed-loop environment governance. The health-aware scheduler routes each task to an eligible phone, App/account, and display; unhealthy targets remain blacklisted until repair and revalidation. Virtual displays allow one phone to run multiple Apps concurrently. Evidence-based review separates task success, model failure, and environment failure from the complete trajectory, and confirmed environment issues are fed back to the scheduler.}
    \label{fig:real-device-loop}
\end{figure}

\paragraph{Why Real Devices Matter?}
Sandbox environments provide stable and resettable task states, making them essential for large-scale rollout and training. However, they cannot fully capture the evolving application states, permission-related constraints, and or execution-time disruptions such as unexpected pop-ups and network instability.
On real devices, application interfaces and content change over time, permissions and CAPTCHA interrupt execution, and network or device failures can occur at any step. The Chinese mobile ecosystem further amplifies these challenges: super Apps, dense interfaces, and frequent pop-ups make it impossible to reproduce such execution paths in a sandbox. To make \modelname{} work reliably under these conditions, we build a real-device mobile runtime and use it for model training and evaluation. We further construct MobileWorld-Real, which is introduced in Section~\ref{sec:MobileWorld-Real}, to systematically evaluate agent performance on real devices.

\paragraph{Real-device Mobile Runtime.}
 We build a real-device mobile runtime comprising over 100 physical devices and over 150 applications. At this scale, assigning each task a reliable and usable execution device becomes a central systems challenge. As shown in Figure~\ref{fig:real-device-loop}, the health-aware scheduler continuously tracks the health and availability of each device, application, account, network connection, and display. When a task enters the queue, it selects an eligible execution target, leases the required resources to the task, and reroutes execution to another target when failures occur. This health-aware orchestration keeps the environment productive despite environment-side failures caused by the inherent instability of real phones, applications, and network connections.

To further scale real-device environment, we leverage the virtual-screen mechanism~\citep{scrcpy} that allows a single phone to host multiple application sessions concurrently on separate displays. The runtime controller binds each virtual display to a corresponding agent session, ensuring that observations are delivered to the correct agent and that its actions are executed on the intended display. This design increases the number of active execution environments without requiring a proportional increase in physical devices. Additionally, real-world tasks often involve either sensitive operations that require explicit user approval or missing information that must be provided by the user. We therefore introduce a dedicated User Agent that can provide missing information, obtain explicit confirmation for data-sensitive or payment-related operations, and hand control to the user when an action requires direct human intervention such as CAPTCHA. After the required input or operation is completed, the task resumes from the updated environment state.

A real-device runtime also needs to distinguish model failures from failures of the execution environment itself. Application restrictions, unavailable services, unstable devices, and network errors may interrupt a trajectory even when the model's decisions are valid. The scheduler maintains a dynamic blacklist over unhealthy devices, removes them from routing, and restores them only after manual inspection. We employ a VLM-based judge to examine the complete trajectory and to distinguish among task success, model failure, and environment failure. Confirmed environment failures update the Scheduler and maintenance queue, while validated trajectories are retained for training.

\subsubsection{Hybrid GUI+CLI Action Space}
Real-device environments expand the range of states an agent can encounter, but real-world utility also depends on how effectively the agent can act on them. 
GUI interaction provides universal access to user-facing applications and visual context, but pure GUI execution may turn structured operations into long sequences of grounding, clicking, and typing. 
CLI execution offers a direct programmatic access to files, code, system settings, and batch operations, but is less suitable when tasks depend on visual understanding, application-specific interfaces, or services without programmable access. GUI and CLI therefore offer complementary forms of environment access: GUI enables broad application coverage and visually grounded interaction, while CLI provides efficient execution over programmatic state.

We therefore equip \modelname{} with the unified GUI+CLI action space defined in Section~\ref{sec:action-space}, and our environments expose both forms of access through a single execution interface. In the computer-use environment, every action is dispatched to a lightweight execution service inside the VM: GUI actions are translated into native input events, while \texttt{cli\_command} runs the command in a non-interactive shell, without visually locating and operating a terminal application. The resulting stdout, stderr, and exit status are returned as structured observations alongside the post-action screenshot, allowing the model to condition its next decision jointly on programmatic and visual state.

Reliable CLI execution requires environment-side handling beyond spawning a shell. Command execution is bounded by a timeout budgeted for slow operations such as dependency installation, while failures like non-zero exits and timeouts are returned as error observations rather than aborting the episode, allowing the model to diagnose and recover within the same trajectory. In batched actions, GUI and CLI sub-actions execute in order with their CLI outputs aggregated into a single observation. Executed commands are also recorded into the shell history, so that verifiers inspecting terminal state credit CLI-based solutions as well.

\subsubsection{Unified Interface for Heterogeneous Environments}

Mobile-use, computer-use, browser-use, and DeepSearch environments expose different resource models, action spaces, observation formats, and verification procedures. 
We therefore introduce a unified environment interface that standardizes the environment lifecycle and agent-facing inputs and outputs, while delegating execution, reset, and verification to environment-specific adapters.
This allows the same agent and training pipeline to operate across heterogeneous environments without imposing identical runtime semantics.

The interface exposes a common asynchronous lifecycle comprising \texttt{acquire}, \texttt{reset}, \texttt{step}, \texttt{evaluate}, \texttt{tear\_down}, and \texttt{release}. Each acquired environment is represented as a leased session bound to a backend and, where necessary, a display identifier, which consistently routes screenshots, actions, and evaluations to the same isolated context. Backend adapters implement resource initialization, observation capture, action execution, and evaluation, while task adapters preserve platform-specific reset and verification logic. All environments return a common transition format containing observations, a termination signal, and reward information.

The policy and environment independently declare their native action spaces. When they differ, supported GUI operations are translated through a canonical intermediate representation, with coordinates normalized according to the target platform. The hybrid GUI, CLI, and API actions defined in Section~\ref{sec:action-space} remain available through native adapters, and their visual or structured results are returned through the same observation stream. This design provides a shared rollout, data-collection, and evaluation pipeline while retaining the native interaction semantics of each platform.

\subsection{Agent-Driven Data Flywheel}
\label{sec:data-flywheel}

\begin{figure}[t]
    \centering
    \includegraphics[width=0.9\textwidth]{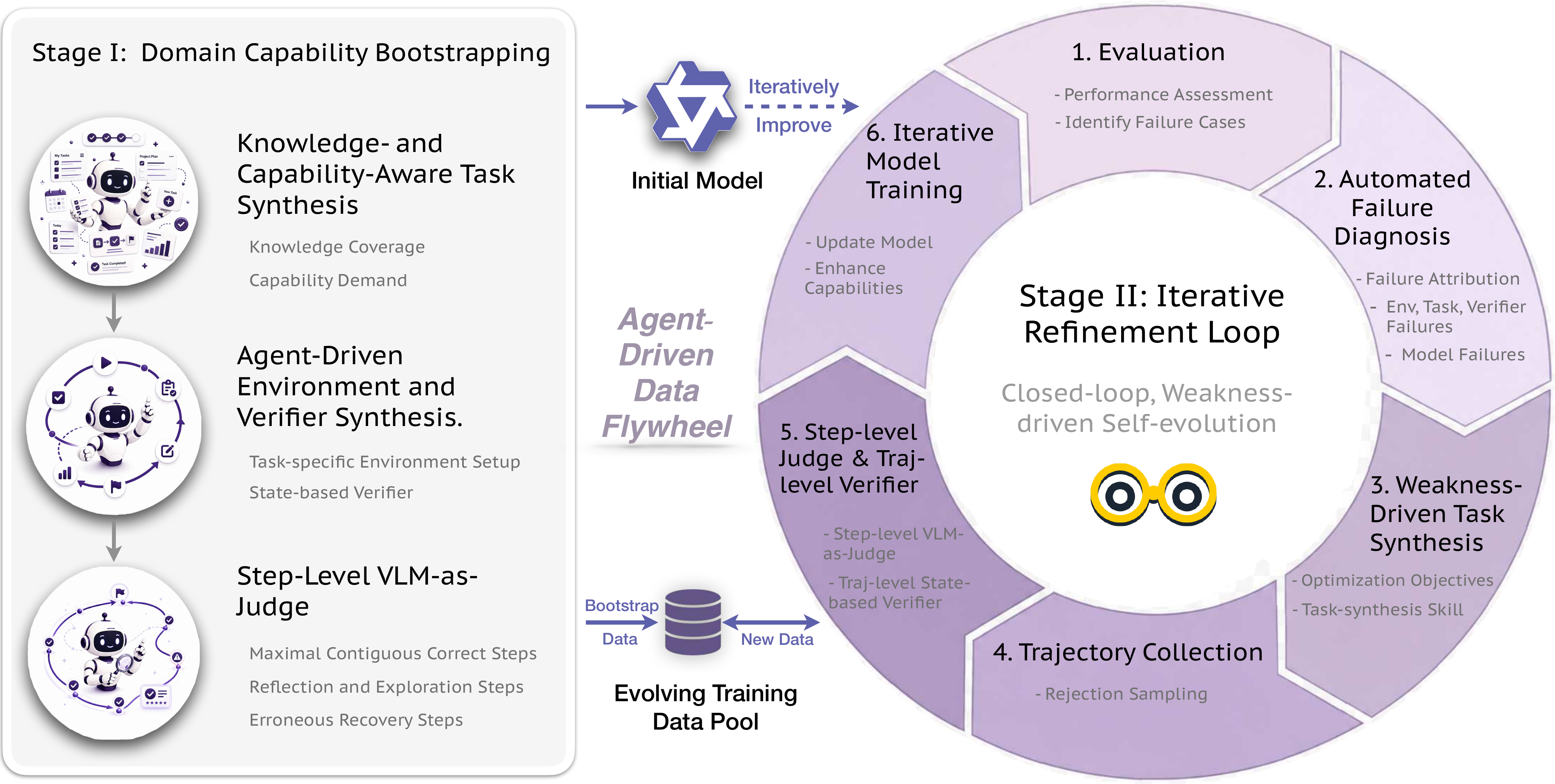}
    \caption{The data flywheel of \modelname{}. Domain capability bootstrapping initializes training, iterative refinement loop identifies capability weaknesses and generate targeted tasks, and the resulting data improve the next training iteration.}
    \label{fig:data-flywheel}
\end{figure}

Continually improving GUI agents requires an iterative, closed-loop process spanning task design, environment construction, verifier development, data curation, model training, evaluation, and error analysis.
When humans drive these stages, iteration becomes slow and difficult to scale as the coverage expands across applications, platforms, and capability dimensions. 
To address this issue, we therefore introduce an AutoResearch-style, agent-driven data flywheel that shifts human involvement from executing each stage to high-level oversight and targeted intervention.
\vspace{-0.7em}
\paragraph{Overall Procedure.}
As illustrated in Figure~\ref{fig:data-flywheel}, we organize our agent driven data flywheel into two stages: domain capability bootstrapping and an iterative refinement loop. During bootstrapping, we use strong foundation models to analyze the domain knowledge and capabilities required for the various domain covering mobile, desktop and web. Guided by this analysis, the agents generate an initial pool of tasks and corresponding environment contexts, from which candidate trajectories are collected.  
Following MAI-UI~\citep{maiui}, we first perform multiple rounds of rejection sampling and aggregate the accepted trajectories into a unified SFT corpus. This process yields a strong initial policy with broad coverage of the target domain. Building on this policy, our agent-driven failure analysis identifies the model's remaining weaknesses and maps them to targeted goals for the next cycle. These goals guide the generation of new tasks, environment configurations, and, where applicable, task-specific verifiers.
The accepted trajectories are then incorporated into the training corpus to further improve the model, thereby closing the loop and initiating the next iteration. Across both the initial policy construction and iterative improvement stages, strong foundation models generate tasks and environment configurations, synthesize task-specific verifiers, evaluate trajectories, analyze failures, and plan the optimization and execution of subsequent iterations. Human involvement primarily focuses on designing and implementing the overall system, overseeing its operation, and intervening or revising the workflow when necessary. We next describe four design choices that make this agent-driven data flywheel scalable and effective.
\vspace{-0.7em}
\paragraph{Knowledge- and Capability-Aware Task Synthesis.}
Training tasks define the learning signals available to the model and therefore shape the knowledge and behaviors it can acquire.
We organize task synthesis along two conceptually distinct dimensions: \emph{knowledge coverage}, which determines what the agent needs to know, and \emph{capability demand}, which determines how the agent must reason and act to complete the task. Knowledge coverage specifies the domain-specific operational knowledge, including application functions, interface conventions, tool usage, and common workflows. Capability demand captures the transferable reasoning and interaction abilities, such as long-horizon state tracking, precise constraint following, numerical and temporal reasoning, and error recovery. For each target domain, we use strong foundation models to construct an initial hierarchical function tree for knowledge coverage and a capability profile for capability demand. We then compile them into a reusable task-synthesis skill containing composition rules, difficulty controls, environment prerequisites, and representative examples. Conditioned on the objective of the current iteration, this skill lets the agent automatically generate task toward the required knowledge and capabilities.
\vspace{-0.7em}
\paragraph{Agent-Driven Environment and Verifier Synthesis.} Task synthesis determines the capabilities that a training instance should elicit, but interactive training additionally requires a reproducible environment in which the task is feasible and executable. 
Given a synthesized task, our agents use \emph{Environment State Synthesis}, detailed in Section~\ref{sec:online}, to construct the application state, files, records, accounts, and cross-application context needed for execution.
For tasks with reliably checkable outcomes, agents further synthesize state-based verifiers that inspect the resulting application, file, database, or system state, yielding executable task--verifier pairs   .

\vspace{-0.7em}
\paragraph{Step-Level Judging for Scalable SFT and Executable Verification for Online RL.} Executable verifiers provide high-precision outcome signals, but constructing and validating one is costly. Even with stronger foundation models, we find that synthesizing and validating executable verifiers remains time-consuming, whereas the accuracy of model-based judges improves alongside model development. We therefore extend the step-level trajectory judge used in MAI-UI~\citep{maiui} and apply a VLM-based judge to mobile, web, and computer-use trajectories, including hybrid GUI+CLI interactions. Given the instruction, trajectory, and visual or structured observations, the judge extracts three forms of useful supervision: maximal contiguous steps that correctly advance the task, the first step that initiates each reflection or exploration phase, and recovery segments that return execution from an erroneous state to a valid path. Retaining only the first reflection or exploration step captures the useful decision to reconsider without over-keeping a potentially noisy branch. This fine-grained procedure recovers effective training signal from both successful and failed trajectories. Empirically, we find that step-level data filtering yields SFT performance comparable to or better than training on complete trajectories selected by executable verifiers. We therefore use fine-grained VLM judgment to scale process supervision for SFT, while constructing executable verifiers for the high-precision outcome signals required by online RL.
\vspace{-0.7em}
\paragraph{Failure Analysis Driven Iteration.} 
We construct a broad and diagnostic evaluation suite for each domain to track model performance and define the optimization objectives of the next iteration. For each failed task, an analysis agent examines the complete trajectory and distinguishes model failures from environment, task, or verifier failures. Model failures are then mapped to a structured cause, such as missing application knowledge, constraint violations, state-tracking errors, or insufficient verification. The system aggregates these task-level diagnoses to identify the most prevalent and consequential weaknesses of the current policy and converts them into a prioritized set of optimization objectives. Conditioned on these objectives and the task-synthesis skill, agents generate targeted tasks and corresponding environment setups and, where applicable, task-specific verifiers for the next cycle. Non-model issues are routed to environment maintenance or task and verifier revision, where agents diagnose and address the identified problems accordingly.

\subsection{Training}
\label{sec:training}

\subsubsection{Supervised Fine-Tuning (SFT)}
\label{sec:sft}
\modelname{} is designed to support task execution across mobile, computer-use, web, and DeepSearch settings. This scope poses two coupled challenges for SFT: jointly acquiring strong capabilities across all target domains within a single model, and preserving the general reasoning and agentic competence needed for robust real-world task completion.
\vspace{-0.5em}
\paragraph{Domain-Conditioned Expert Training.}
Across mobile, desktop, and web, each domain places different demands on the agent. Mobile tasks require reliable execution in real-device environments, where interfaces are app-specific, workflows are frequently interrupted by pop-ups, and safety-sensitive actions must remain under user control. Desktop tasks often combine GUI control with Bash-based file and system operations. Web tasks require interaction with dynamic webpages, while DeepSearch tasks require the agent to retrieve, verify, and synthesize evidence through Search APIs.

We train a specialized expert for each domain and then consolidate their capabilities. Each expert is trained mainly on data from its target domain, together with a controlled mixture of data from other domains. This cross-domain mixture helps each expert retain capabilities that transfer across domains and reduces overfitting to its target domain. Practical deployment requires a single model that can operate across all supported settings. We therefore merge the domain-expert checkpoints into a unified model. Model merging combines their complementary strengths within a single checkpoint, preserving the capabilities acquired by the individual experts with little additional training overhead.

\vspace{-0.5em}

\paragraph{Preserving General and Agentic Capabilities with In-Distribution Data.}General reasoning and agentic capabilities are essential for real-world use, where agents need to respond to diverse and open-ended user requests. Some requests may even be unrelated to GUI tasks, yet the model should still respond appropriately. Even within GUI tasks, real-world deployment exposes the agent to out-of-distribution (OOD) states and objectives that may require knowledge, reasoning, instruction following, retrieval, coding, or tool use not anticipated during training. We therefore seek to preserve and, where possible, strengthen the general reasoning and agentic capabilities inherited from the base model, providing an important foundation for handling unfamiliar situations and maintaining a consistent user experience.

\vspace{-0.3em}
We find that the most effective strategy is to use in-distribution data that reinforces successful behaviors that the starting model can produce. We construct a broad query pool covering general question answering, mathematics, coding, visual understanding, search, and agentic tool use, sample responses from the starting model, and retain verified correct examples after format normalization. A small proportion of these examples is mixed with GUI trajectories during SFT. In our experiments, examples that the starting model is capable of solving are substantially more effective for capability preservation than challenging examples on which it fails. The former rehearse established capabilities using targets aligned with the model's own response distribution, whereas the latter shift training toward capability acquisition and can introduce competing optimization signals.

\vspace{-0.5em}
\paragraph{Sliding-Window Training for Efficient Long-Trajectory SFT.}
Long-trajectory SFT becomes expensive when every action is packed with its visual and textual context as a separate training sample, causing highly overlapping context to be processed repeatedly across adjacent actions. We therefore divide each trajectory into windows of $n=5$ consecutive steps and advance each window by $n-1=4$ steps, leaving a one-step overlap between adjacent windows. This overlap ensures that the first newly supervised action in each subsequent window can condition on at least two screenshots. In each subsequent window, the shared boundary step remains in the input context, but its loss is masked because it has already been supervised in the preceding window. Each full subsequent window therefore supervises four new actions. Screenshots before the current window are omitted from the visual input, while the earlier textual trajectory history is retained as context. Joint supervision reduces this repeated context processing and increases the number of supervised action targets per processed context, thereby improving training efficiency.

\subsubsection{Action RL: Correcting Recurring Action Errors}
A single local error in a long-horizon GUI task can move the interface into an incorrect state and compromise subsequent decisions.
SFT learns from successful demonstrations but does not explicitly suppress recurring action errors.
We therefore introduce \textbf{Action RL}, an action-level reinforcement learning stage that combines targeted error-pattern data with action-aware rewards to improve local decision reliability.
\vspace{-0.5em}
\paragraph{Recurring Action-Error Patterns.}
Across applications, we identify six recurring patterns:
\begin{itemize}[leftmargin=*]
\vspace{-0.5em}
    \item \emph{Confusable-Element Grounding.}
    When a screen contains visually or semantically similar icons, controls, or list items, the model may ground the action to a nearby distractor rather than the correct one.

    \item \emph{Sorting and Ranking.}
    The model may misinterpret ordering requirements such as ordinal positions, top-$k$ selection, newest-versus-oldest ordering, or rankings conditioned on multiple attributes, leading it to select the wrong item or process items in the wrong order.

    \item \emph{Quantity and Multi-Target Completeness.}
    Tasks that specify an exact quantity or require actions over multiple related targets are prone to partial completion. The model may process too few or too many objects, or omit one component of a compound objective.

    \item \emph{Premature Completion.}
    The model may declare success after preparing the intended result but before executing the final state-changing action, such as saving an edit, sending a message, submitting a form, or publishing content.

    \item \textbf{Repetitive Action Loops.}
    The model may repeatedly execute the same action or a short action sequence without making task-relevant progress, often because it fails to recognize that the interface state has not changed or cannot identify a recovery action.

    \item \textbf{Long-Tail Action Selection Failures.}
    The model may fail to invoke infrequent but critical actions, such as \texttt{open}, \texttt{ask\_user}, or \texttt{long\_press}, and instead fall back on unsuitable frequent actions such as \texttt{click}. 
\end{itemize}
\vspace{-0.5em}
The shared structure of these failures makes them reusable targets for action-level optimization across applications.
Their importance, however, is not reflected by their frequency in naturally collected trajectories: rare but consequential patterns can be overwhelmed by more common actions.
\vspace{-0.5em}
\paragraph{Targeted Error-Pattern Data Construction.}
To correct this coverage imbalance, we combine historical-trajectory mining with active environment interaction.
From existing trajectories, we localize the actions that initiate each failure; for repetitive behavior, a visual judge distinguishes unproductive loops from purposeful exploration.
For patterns with insufficient historical coverage, agents actively explore executable environments, identify relevant interface structures, construct tasks around them, and collect new rollouts.
The resulting corpus preserves realistic failure contexts from the current policy while deliberately increasing the representation of rare but important errors.
\vspace{-0.5em}
\paragraph{Action-Aware Reward and Training.}
Using this targeted corpus, action RL assigns a structured reward to each predicted action at step $t$:
\begin{equation}
r_t =
F_t
\left(
w_{\mathrm{type}}C_t
+
w_{\mathrm{arg}}C_tQ_t
-
\lambda_{\mathrm{sens}}S_t
-
\lambda_{\mathrm{rep}}L_t
\right).
\end{equation}
where $F_t\in\{0,1\}$ indicates format validity, $C_t$ measures action-type correctness, and $Q_t$ measures argument quality using pixel distance, lexical similarity, tag matching, or LLM-based judgment.
The terms $S_t$ and $L_t$ penalize incorrect sensitive actions and repetitions in the action history, respectively.
During training, we observe declining token entropy and progressively shorter reasoning traces.
We therefore apply entropy regularization together with lower and upper bounds on reasoning length to prevent policy collapse while avoiding unnecessary verbosity.

\subsubsection{Online RL: Learning Long-Horizon Decision Making}
\label{sec:online}
Long-horizon GUI tasks are not simply collections of locally correct actions.
An action may appear reasonable in the current state yet lead the agent into a future state from which the task becomes difficult or impossible to complete.
Action RL improves local decision reliability, but does not fully capture such delayed trajectory-level consequences; we therefore introduce Online RL to optimize end-to-end task success through environment interaction.

\paragraph{Unified Environment Infrastructure for Scalable Rollouts.}
Learning from complete interactions requires large-scale and reliable rollout generation across heterogeneous GUI environments.
Building on the sandbox infrastructure described in Section~\ref{sec:scaling-sandbox-environments}, we develop a unified environment infrastructure that manages environment allocation, reset, interaction, and evaluation across domains.
On Alibaba Cloud, this infrastructure deploys sandbox environments for mobile, computer use, web, and DeepSearch, supporting up to $10{,}000$ concurrent rollouts.
Each domain retains its native applications, action spaces, and verification mechanisms, while the shared infrastructure exposes a common environment lifecycle and rollout interface to the training pipeline.
For tasks that must be executed on physical phones, we use the virtual-screen mechanism described in Section~\ref{sec:real-device-mobile}, allowing multiple application sessions to run concurrently on separate displays of a single device.
Across our real-device cluster, this mechanism increases aggregate rollout throughput by approximately $20\times$ compared with running the same cluster without virtual screens.
Together, this infrastructure provides scalable rollout capacity across both sandbox and real-device environments for Online RL.

\paragraph{Task--Verifier Synthesis.}
Building on this rollout infrastructure, we automatically construct executable task--verifier pairs that provide reliable outcome supervision through three phases:
\begin{itemize}[leftmargin=*]
    \item \textbf{Environment State Synthesis.}
    Coding agents analyze the codebases underlying each sandbox and distill reusable data-injection skills for their applications.
    Using these skills, the agents construct coherent initial environment states spanning multiple applications.
    For example, an environment state representing a computer science PhD student preparing a paper submission may contain mutually consistent emails, photos, files, calendar events, and social-media records across the student's phone.
    Environment state synthesis is critical because it largely determines the difficulty and diversity of downstream tasks.
    Generating tasks directly from benchmark-provided states confines task synthesis to the application states and data already available in the benchmark.

    \item \textbf{Task Synthesis.}
    Conditioned on each synthesized environment state, an LLM generates multiple tasks at different difficulty levels, including workflows spanning multiple applications.
    An independent LLM judge then filters the generated tasks for feasibility and consistency with the initialized environment state.

    \item \textbf{Verifier Synthesis.}
    Using the unified environment infrastructure, coding agents autonomously launch corresponding sandboxes, inject synthesized environment states, and construct task-specific code-based verifiers.
    Each verifier is validated through rollouts from multiple agent models, while a VLM-based judge assesses whether verifier outputs are consistent with observed rollout evidence.
\end{itemize}
This automated synthesis and execution-validation pipeline produces approximately $10{,}000$ validated task--verifier pairs for Online RL.

\paragraph{Online RL with a Model-Adaptive Task Curriculum.}
Given the validated task--verifier pairs, we use verifier-guided Online RL to improve long-horizon decision making.
Following MAI-UI~\citep{maiui}, we adopt a tailored variant of Group Relative Policy Optimization (GRPO) for GUI interaction trajectories.
For each task, the current policy samples a group of $K$ complete interaction trajectories $\{\tau_i\}_{i=1}^{K}$.
The verifier evaluates the final environment state $s_T^{(i)}$ of each trajectory.
We compute its binary outcome reward and group-relative advantage as
\begin{equation}
r_i = v_x\bigl(s_{\mathrm{final}}^{(i)}\bigr) \in \{0,1\},\quad
\bar{r}_x = \frac{1}{K}\sum_{j=1}^{K} r_j,\quad
\hat{A}_i = \frac{r_i - \bar{r}_x}{\operatorname{Std}(r_1,\dots,r_K) + \epsilon}.
\end{equation}
where the mean and standard deviation are computed over the \(K\) rollouts sampled for the same task \(x\).
GRPO updates the policy using these relative advantages.
Unlike action RL, which provides localized action-level feedback, this stage optimizes complete trajectories using terminal verifier outcomes, targeting long-horizon planning, state tracking, and intermediate decision making.

The training value of a task is not fixed: a task that is initially too difficult may become learnable as the policy acquires related capabilities from other tasks.
Under the group-relative objective, tasks whose rollouts all fail or all succeed produce no reward variation and therefore provide little immediate learning signal.
However, permanently removing difficult tasks would miss the point at which they enter the policy's learning frontier.
We therefore treat task difficulty as a dynamic property of the current policy and organize the curriculum into an active training pool and a monitoring pool.
Tasks with intermediate empirical success rates enter the active pool and receive the full rollout budget, while currently unsolvable tasks remain in the monitoring pool with a smaller budget.
Once a difficult task begins to produce successful rollouts, it is promoted to the active pool for full-scale training.
Mastered tasks are also monitored with a small budget and reactivated if their performance declines.
This closed-loop curriculum continually identifies newly learnable tasks while concentrating rollout computation where it provides the most informative relative rewards.

\subsection{Harness Layer: Proactive Service and Cross-Platform Task Execution}
\label{sec:harness}
\begin{figure}[t]
    \centering
    \includegraphics[width=\textwidth]{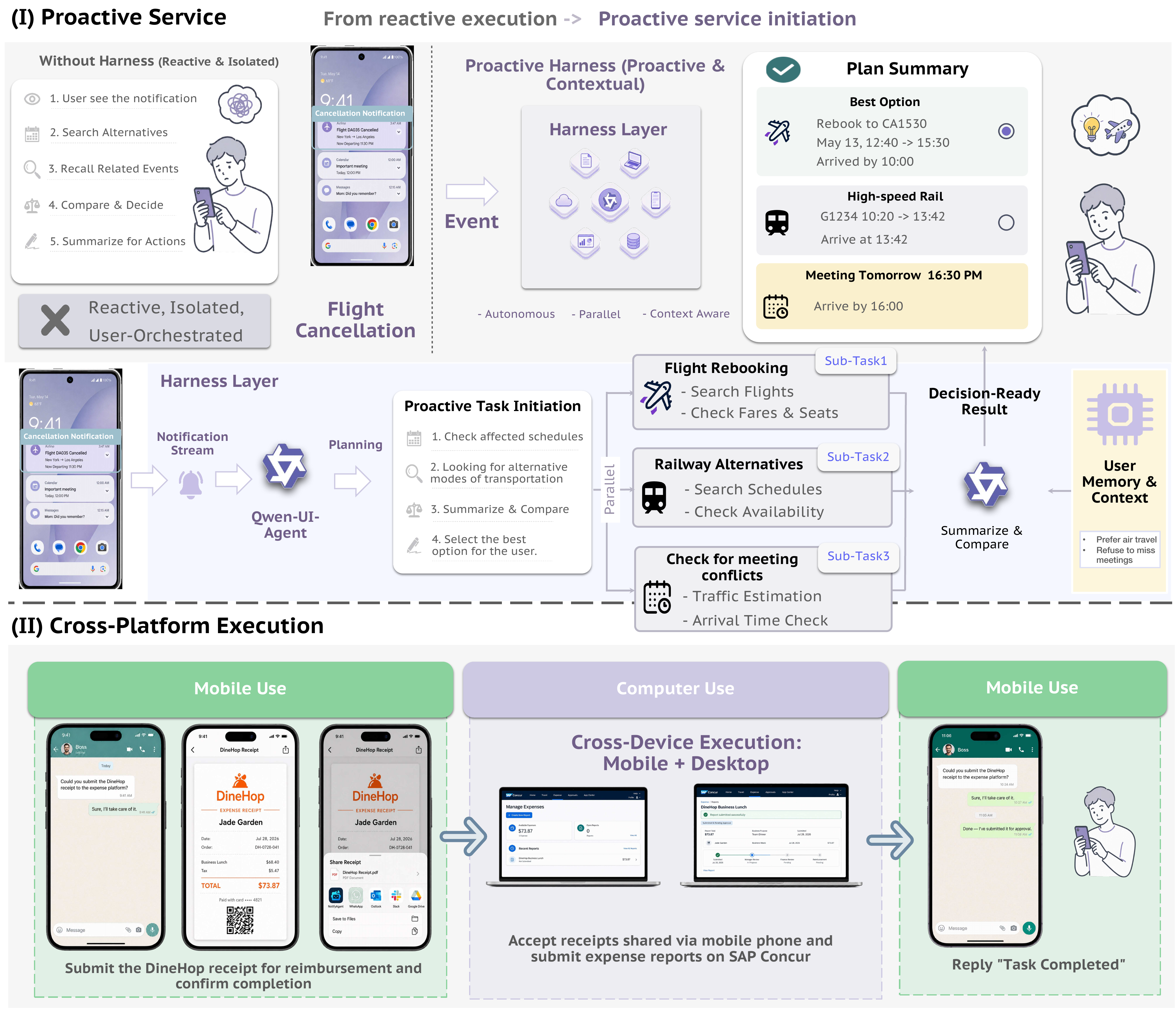}
\caption{Overview of our harness for proactive service initiation and cross-platform execution. (I) Without the harness, users must manually interpret notifications, retrieve related context, and provide instructions for the agent. The proactive harness instead detects a flight cancellation, reasons over relevant affairs, proactively implement flight rebooking, railway alternatives, and meeting-conflict checks, and report an actionable plan. (II) The harness also maintains task state and support cross-platform execution. In the restaurant-selection example, \modelname{} discovers candidates on mobile, organizes them in a desktop spreadsheet, sends the artifact for approval, and saves the selected locations back to mobile.}
    \label{fig:harness}
\end{figure}
Strong task execution is necessary, but it is not sufficient for GUI agents to move beyond isolated task execution and become effective assistants in users’ everyday digital lives. Existing agents are largely reactive: users must recognize a need, formulate an instruction, and invoke the agent, even when that need is already evident from signals in their digital environment. Workflows may also span devices, with relevant information appearing on a phone while subsequent actions must be completed on a computer. This creates two fundamental gaps: \emph{when} should an agent initiate assistance, and \emph{how} can it carry the resulting workflow across devices without losing context? We address both through a shared harness layer above the underlying GUI agents. The harness connects signals from the user's digital environment, maintains a unified representation of ongoing activities, and uses this context to initiate and coordinate tasks across mobile and computer environments.

Built on this layer, we develop two capabilities. \emph{Proactive service} addresses the first gap by shifting task initiation from explicit instructions to to actionable events inferred from the user's digital signals, enabling the agent to identify and prepare useful assistance before the user asks. \emph{Cross-platform task execution} addresses the second by carrying workflows across phones and computers while preserving shared context, dependencies, and task state. Together, these capabilities extend GUI agents along two complementary dimensions: \emph{when} a task can begin and \emph{where} it can be carried forward.
\vspace{-0.5em}
\subsubsection{Proactive Service Based on Mobile Notifications}
\paragraph{Mobile Notifications as a High-Value Signal Source.}
A proactive agent needs signals that are timely, broadly available, and controllable by the user. We use mobile notifications as this interface. A phone aggregates signals from communication, travel, finance, logistics, and scheduling, and notifications expose many events at the moment they become actionable, such as a flight cancellation, a payment deadline, or a schedule change. Compared with continuous screen, audio, or video monitoring, notification access is lightweight and governed by explicit per-application permissions. Notifications are nevertheless partial and noisy: they report only what an application chooses to surface. The harness therefore treats them as observations rather than complete world state, and gathers additional information only when needed.
\vspace{-0.7em}
\paragraph{From a Notification to a Proactive Workflow.}
Consider a late-night notification that the user's flight the following morning has been canceled. A conventional GUI agent would wait for an explicit instruction before searching for alternatives. We instead envision a proactive workflow that goes beyond a simple search. A flight cancellation is rarely an isolated event: it may jeopardize an important meeting, require changes to a hotel reservation, and force a choice between rebooking and rail travel. The key question is whether an agent can anticipate and prepare these interdependent follow-up actions before the user asks. Our harness does so by reconstructing the broader context before requesting the users attention. It parses the cancellation into a structured event, associates it with the ongoing trip and the relevant meeting deadline, retrieves user preferences from memory, queries feasible alternatives, evaluates them against the user's underlying objective, and presents a decision-ready proposal. Actions with external consequences, such as rebooking or modifying a reservation, remain subject to user confirmation. Figure~\ref{fig:proactive_demo_a} and Figure~\ref{fig:proactive_demo_b} of Appendix~\ref{app:examples} show two representative examples of this behavior: a scheduled morning brief that combines the user's commitments with live weather and commute conditions, and a proactive travel-recovery plan triggered by a flight-cancellation notification.

The harness implements this workflow as a stateful pipeline that parses notifications into structured events, associates them with persistent affairs, derives executable tasks, and uses execution outcomes to update memory and future behavior.
Its central abstraction is the \emph{affair}, a persistent representation of an ongoing real-world matter. An \emph{event} records what happened at a particular time; an \emph{affair} captures what is still unfolding across events, applications, and days; and a \emph{task} specifies what should be done next. This separation allows the harness to reason about the user's underlying goals.

\begin{itemize}[leftmargin=*]
\vspace{-0.5em}

\item \textbf{Event Perception.}
Incoming notifications are processed in short windows. The event parser removes low-value noise and converts useful notifications into structured events containing their source, entities, timestamp, urgency, and available actions. These events are stored with their temporal order and provenance, providing an evidence base for subsequent reasoning.

\item \textbf{Affair State and Memory.}
For each new event, the harness retrieves relevant recent events, active affairs, the current context state, and profile memory. It then associates the event with an existing affair or creates a new one, updating the affair’s lifecycle, key entities, deadlines, unresolved gaps, and supporting evidence. Affairs retain the evolving state of individual trips, purchases, meetings, or family matters, while profile memory stores more stable knowledge such as user preferences, relationships, and behavioral patterns. Together, these two forms of memory combine situation-specific reasoning with long-term personalization.

\item \textbf{Affair-level Reasoning and Task Formation.}
The harness reasons over updated affairs to identify expected next steps, missing information, unfinished actions, time pressure, and dependencies or conflicts across affairs. Conclusions are grounded in observed events or stored user information and are associated with confidence estimates. The task generator then evaluates whether intervening now would provide sufficient value, considering factors such as urgency, consequence, evidential support, and the amount of user effort saved. Depending on this assessment, a task is proposed immediately, placed in a passive to-do list, or suppressed.

\item \textbf{Proactive Preparation and Controlled Execution.}
For a selected task, the harness chooses an appropriate executor, such as an information-seeking tool, a mobile GUI agent, or a computer-use agent. It may perform low-risk preparatory actions before interrupting the user, such as retrieving alternative trains, comparing travel times, or checking refund status. The resulting task card therefore presents concrete, decision-ready options rather than another reminder. Operations with external consequences, including payments, bookings, and sending messages, remain subject to user confirmation.

\item \textbf{Personalization and Continual Evolution.}
The harness continuously refines both its understanding of the user and its proactive behavior. Profile memory is built from explicit user input, evidence-supported observations, allowing the system to learn preferences, relationships, and routines. Feedback memory records how users respond to proposed tasks, including approval, modification, rejection, ignoring, and completion. These signals reinforce or revise profile entries and calibrate future task value, timing, and degree of automation. Evolution therefore occurs at two levels: profile memory learns \emph{what} matters to the user, while feedback memory learns \emph{when and how} the agent should intervene, making proactive assistance increasingly personalized and useful.

\end{itemize}

\subsubsection{Cross-Platform Task Execution}
Users' digital activities are distributed primarily across mobile devices and computers. As a result, many real-world workflows span multiple platforms whose interfaces, capabilities, and states differ. For example, a user may ask the agent to identify receipts in the mobile photo gallery, transfer the selected images to a desktop, rename them according to their recognized dates, and generate a structured expense spreadsheet. This workflow combines visual selection on the phone, cross-device artifact transfer, and file processing on the computer. Treating each device as an independent episode would fragment the task state, obscure dependencies and failures, and leave the user to coordinate the handoffs. Figure~\ref{fig:crossplatform_demo_a} and Figure~\ref{fig:crossplatform_demo_b} of Appendix~\ref{app:examples} show two such workflows executed by our harness: the receipt-organization task above, and a parallel restaurant search that operates multiple mobile apps concurrently before summarizing the results on the computer; in both cases, mobile subtasks run on virtual screens without blocking the user's own use of the device. We therefore implement the harness as a hierarchical planner--executor system with device-addressed actions, shared execution state, and explicit cross-platform operations:
\begin{itemize}[leftmargin=*]
\vspace{-0.5em}
\item \textbf{Global Perception and Device Grounding.}
The harness first enumerates available devices, applications, displays, and workspaces. Each observation is tagged with its application, device, and display, and screenshots are assembled into a labeled multi-device view. This allows the planner to inspect mobile and desktop states jointly. The shared state stores the instruction, application--device mappings, recent observations, artifacts, tool outputs, and subtask results. Device-tagged updates change only the corresponding platform state while preserving the remaining context.

\item \textbf{Dependency-aware Planning and Executor Selection.}
We implement an OpenClaw-like planner with a persistent session, workspace, and tool registry \citep{openclaw2026}. The planner decomposes the objective into dependency-aware subtasks and selects an appropriate executor for each. It invokes \modelname{}, exposed as a GUI subagent, for visually grounded application interaction. Search, file, CLI, and API tools handle information seeking, deterministic system operations, and synthesis. Each GUI call specifies its instruction and target application and device. Returned results update the session, allowing subsequent steps to be replanned from the latest state.

\item \textbf{Parallel Multi-agent Execution.}
Independent calls to different applications or environments run concurrently. A unique identifier and observation queue isolate each subtask's actions and feedback. Its action--observation loop is sequential, while different subtasks progress in parallel. On Android, a display manager assigns application-specific subtasks to separate virtual displays on one physical device. Each display provides independent capture and display-addressed input, enabling multiple \modelname{} instances to operate different applications concurrently.

\item \textbf{Hybrid Execution with \modelname{}.}
Each GUI subtask uses the hybrid action space in Section~\ref{sec:action-space}. \modelname{} uses GUI actions for visually grounded interaction and CLI or API actions for structured operations, file processing, and external services. Desktop CLI actions execute shell commands, while Android actions use ADB shell commands; their output, errors, and exit status return as structured observations. Compatible operations can be batched. Platform adapters route actions to the addressed environment and transfer artifacts between devices.

\end{itemize}

\section{Experiments}
\subsection{Experimental Setup}
\paragraph{Models and implementation details.}
We evaluate three variants of \modelname{}, including 27B, 35B-A3B, and 4B models. 
The 27B model serves as our primary variant for end-to-end agent evaluation across mobile, computer-use, browser, DeepSearch, and general agentic benchmarks. 
The models are initialized from their corresponding base checkpoints and trained using the pipeline described in Section~\ref{sec:training}.
Unless otherwise specified, we follow the official interaction budget and evaluation configuration of each benchmark.
\vspace{-0.7em}
\paragraph{Benchmarks.}
We evaluate \modelname{} across five groups of benchmarks covering real-device mobile use, computer use, browser use, DeepSearch, GUI grounding, and general and agentic capabilities.
\begin{itemize}[leftmargin=1.5em, itemsep=0.2em, topsep=0.3em]
\vspace{-0.5em}
\item \textbf{Real-device mobile use.}
We evaluate end-to-end mobile task execution on MobileWorld-Real and AndroidDaily~\citep{androiddaily}.
MobileWorld-Real, introduced in Section~\ref{sec:MobileWorld-Real}, contains complex human-written tasks executed on live Android devices in the Chinese mobile ecosystem.
AndroidDaily focuses on high-frequency everyday tasks on real Android devices and provides a complementary evaluation of common mobile-use scenarios. We further evaluate on MobileWorld~\citep{kong2026mobileworld}, which provides a stable simulated environment for challenging long-horizon, cross-application tasks.

\item \textbf{Computer use.}
We evaluate desktop task execution on OSWorld-Verified~\citep{OSWorld} and OSWorld-v2~\citep{osworld2}.
OSWorld-Verified evaluates task progress across a broad collection of desktop applications and operating-system workflows.
OSWorld-v2 places greater emphasis on longer-horizon tasks and reports both partial progress and binary task completion.
These benchmarks evaluate whether the model can combine visual interaction, command-line operations, and batched actions to complete complex computer-use workflows.
\begin{figure}[t]
    \centering
    \includegraphics[width=\linewidth]{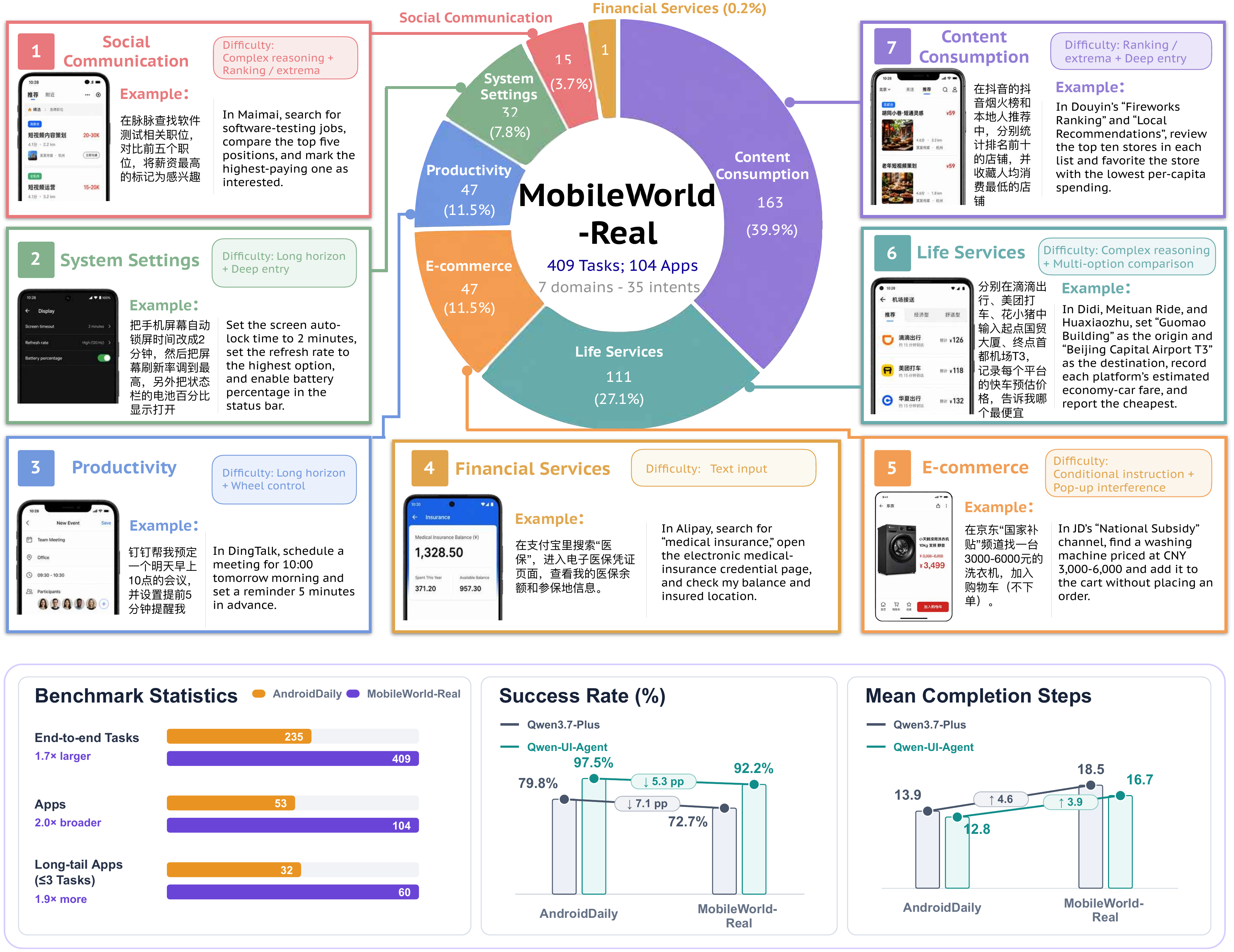}
    \caption{\textbf{Overview of MobileWorld-Real, a real-device benchmark with human-written tasks that reflect the breadth of everyday mobile use.} Representative examples and aggregate statistics show broad domain coverage and a long-tailed App distribution. Matched-model results further show lower success rates and longer trajectories than on AndroidDaily, highlighting the challenge of real-world mobile interaction.}
    \label{fig:MobileWorld-Real-profile}
\end{figure}
\item \textbf{Web navigation and DeepSearch.}
We use WebArena~\citep{zhou2023webarena} to evaluate end-to-end browser interaction on functional websites, where tasks require the model to track evolving page states and complete multi-step user objectives.
We separately evaluate DeepSearch on BrowseComp~\citep{browsecomp} and BrowseComp-ZH~\citep{browsecompzh}, which measure persistent information seeking, cross-source evidence verification, and answer synthesis over the English and Chinese web.
Together, these benchmarks distinguish direct interaction with web interfaces from research-intensive tasks that require information aggregation and verification.

\item \textbf{GUI grounding.}
We evaluate element localization on ScreenSpot-Pro~\citep{li2025screenspotpro}, ScreenSpot-V2~\citep{osaltas_and_screenspot_v2}, MMBench-GUI L2~\citep{mmbenchgui}, {OSWorld-G-Refined}~\citep{osworld_g}, and UI-Vision~\citep{ui_vision}.
These benchmarks cover high-resolution professional software, mobile and web interfaces, desktop applications, text and icon targets, spatial reasoning, functional reasoning, and goal-oriented grounding instructions.

\item \textbf{General and agentic capabilities.}
We evaluate whether GUI post-training preserves the general capabilities of the base model using MMMU-Pro~\citep{mmmupro}, RealWorldQA~\citep{realworldqa}, CharXiv-RQ~\citep{charxiv}, MathVision~\citep{mathvision}, AI2D~\citep{ai2d}, MMLU-Pro~\citep{mmlupro}, and IFEval~\citep{ifeval}.
We further evaluate broader agentic capabilities on Tau2-Bench~\citep{tau2bench}, Terminal-Bench 2.0~\citep{terminalbench}, Claw-Eval~\citep{claweval}, BFCL-v4~\citep{bfcl}, SkillsBench~\citep{skillsbench}, and QwenClawBench~\citep{qwenclawbench}.
These evaluations measure capabilities such as multimodal reasoning, instruction following, tool use, terminal operation, multi-turn interaction, and autonomous task execution beyond routine GUI navigation.

\end{itemize}

\vspace{-0.7em}
\paragraph{Evaluation Protocols.}
We follow the official task sets, environments, and primary scoring protocols unless otherwise stated.
For interactive GUI benchmarks, we use the benchmark-defined task and step limits and report the corresponding success, partial-progress, or binary-completion metrics.
For MobileWorld-Real, AutoJudge assigns each trajectory one of three outcomes: \texttt{pass}, \texttt{failed}, or \texttt{env\_error}.
Environment errors are reported separately and excluded from the success-rate denominator, while average trajectory length is computed over successful runs.
For WebArena, we manually verify incorrect reference answers and identified errors in the official evaluation scripts before evaluation.
\vspace{-0.7em}
\paragraph{Compared Baselines.}
We compare \modelname{} against three groups of systems: frontier proprietary models, strong open-weight foundation models, and specialized GUI agents.
For the general and agentic benchmarks, we additionally compare the 27B model with its base checkpoint to measure capability retention after GUI post-training.
We use officially reported results when the evaluation setting is directly comparable and independently reproduce results when a matched evaluation environment is required.
The source and evaluation setting of each result are indicated in the corresponding table.

\vspace{-0.5em}

\subsection{MobileWorld-Real: Real-world Mobile Use Evaluation Benchmark}
\label{sec:MobileWorld-Real}

Most mobile GUI benchmarks run in sandboxes, where app states can be reset and task outcomes can be verified programmatically~\citep{kong2026mobileworld, android_world}. Such environments support controlled comparison, but they can not fully show whether a GUI agent can handle the changing content, account states, and interruptions found on users' phones.
We introduce MobileWorld-Real, a real-device benchmark for mobile GUI Agents in the Chinese mobile ecosystem. MobileWorld-Real uses everyday tasks written by humans and runs them on live Android devices with real Apps, accounts, content, and networks. It evaluates whether an agent can complete the user's goal while handling the uncertainty and interruptions that arise during real use.

\paragraph{Benchmark Design and Task Coverage.}
MobileWorld-Real contains 409 end-to-end tasks over 104 apps, each derived from an everyday need identified by a human contributor.
These tasks fall into 7 domains of daily mobile use: content consumption, life services, productivity, e-commerce, system settings, financial services, and social communication.
Within each domain, the tasks cover multiple user intents. For example, content consumption includes searching for content, browsing feeds, posting, and sharing.
To evaluate these tasks under realistic conditions, we execute them on the real-device runtime described in Section~\ref{sec:real-device-mobile}, with live account states and dynamically changing online content.
This exposes the GUI agent to pop-ups, expired logins, permission requests, CAPTCHAs, and other conditions rarely encountered in sandbox evaluations.

\begin{table}[!t]
\caption{
Performance comparison on the GUI-only subset of
MobileWorld~\citep{kong2026mobileworld}, consisting of 117 tasks.
We include representative general-purpose VLMs and specialized GUI
models. 
The best result is highlighted in \textbf{bold}, and the second-best result is \underline{underlined}.
}
\label{tab:mobileworld}

\centering
\footnotesize
\renewcommand{\arraystretch}{1.15}

\begin{minipage}{0.95\linewidth}
\centering
\setlength{\tabcolsep}{4.5pt}

\begin{tabularx}{\linewidth}{
  >{\raggedright\arraybackslash}X
  >{\centering\arraybackslash}p{0.21\linewidth}
  >{\centering\arraybackslash}p{0.21\linewidth}
}
\toprule[1.2pt]
\textbf{\textsc{Model}} &
\textbf{\textsc{Access / Size}} &
\textbf{\textsc{Success Rate (\%)}} \\
\midrule

\rowcolor{gray!15}
\multicolumn{3}{l}{\textit{General-purpose VLMs}} \\

Seed~2.1~Pro~\citep{bytedanceseed2026seed21}
    & Closed-source
    & \cellcolor{light_purple}\underline{73.2} \\

GPT-5.6 Sol~\citep{openai2026gpt56}
    & Closed-source
    & \cellcolor{light_purple}70.1 \\

Claude Opus~4.8~\citep{anthropic2026claudeopus48}
    & Closed-source
    & \cellcolor{light_purple}67.5 \\
    
Seed~2.0~Pro~\citep{bytedanceseed2026seed20}
    & Closed-source
    & \cellcolor{light_purple}63.2 \\

Qwen~3.7~Plus~\citep{qwen2026qwen37plus}
    & 397B-A17B
    & \cellcolor{light_purple}62.3 \\

Gemini 3.1 Pro~\citep{google2026gemini31pro}
    & Closed-source
    & \cellcolor{light_purple}58.1 \\

Kimi K2.6~\citep{moonshot2026kimik26}
    & 1T-A32B
    & \cellcolor{light_purple}55.6 \\

\midrule
\rowcolor{gray!15}
\multicolumn{3}{l}{\textit{Specialized GUI Models}} \\

GUI-Owl-1.5-32B-Instruct~\citep{Mobile-agent-v3.5}
    & 32B
    & \cellcolor{light_purple}43.9 \\

MAI-UI-235B-A22B~\citep{maiui}
    & 235B-A22B
    & \cellcolor{light_purple}39.7 \\

UI-Venus-1.5-30B-A3B~\citep{uivenus15}
    & 30B-A3B
    & \cellcolor{light_purple}17.1 \\

\midrule
\rowcolor{gray!15}
\multicolumn{3}{l}{Ours} \\

\modelname{}
    & 27B
    & \cellcolor[HTML]{cfcdfd}\textbf{82.1} \\
\modelname{}
    & 35B-A3B
    & \cellcolor[HTML]{cfcdfd}65.0 \\

\bottomrule[1.2pt]
\end{tabularx}
\end{minipage}
\end{table}

Beyond these environmental challenges, many tasks also involve complex, multi-step workflows. Nearly half are labeled as hard, with common challenges including long-horizon execution, comparison and ranking, reasoning over changing information, navigation to deeply nested app functions, pop-up recovery, and cross-app coordination.
The benchmark further captures the ambiguity of real user requests, which do not always provide all the information needed for successful execution. In such cases, the agent must identify the missing information, use \texttt{ask\_user} to obtain the necessary details, or request user takeover for steps requiring user intervention before continuing the task. 

We balance MobileWorld-Real's realism with the need for repeated evaluation. MobileWorld-Real avoids tasks that depend on a particular past order, cart state, bank card, or other conditions that cannot be set or restored reliably. This ensures high stability of the evaluation results.
Figure~\ref{fig:MobileWorld-Real-profile} summarizes MobileWorld-Real's task coverage, challenge profile, and representative examples.

\vspace{-0.5em}
\paragraph{AutoJudge for Trajectory-level GUI Agent Evaluation.}
Evaluating task success on live devices is challenging. Unlike sandbox environments, real apps usually do not expose programmatic verifiers for their internal states. 
Moreover, an unsuccessful run may result from either agent failure or environmental issues, such as CAPTCHAs, expired login sessions, network failures, unavailable services, or infrastructure errors. The final screen alone is often insufficient to distinguish these cases.

To enable scalable and repeatable evaluation, we develop AutoJudge, a trajectory-level evaluator for real-device interaction. 
The judge prompt specifies explicit decision criteria for assigning each trajectory to one of three outcomes: \texttt{pass}, \texttt{failed}, and \texttt{env\_error}. 
For each run, AutoJudge receives the task instruction and the complete execution trajectory, with every action aligned with its corresponding screenshot. 
Five independent VLM judges examine the same trajectory, and each produces an outcome label with a short rationale. 
The final outcome is determined by majority vote. 
By reporting environment errors separately and excluding them from the success-rate denominator, AutoJudge reduces evaluation noise caused by changing live-device conditions. Appendix~\ref{app:autojudge-validation} evaluates AutoJudge against independent expert annotations on 666 trajectories, and its exact-match accuracy reaches 92.8\%. Figures~\ref{fig:autojudge_pass}, \ref{fig:autojudge_model_failure}, and \ref{fig:autojudge_environment_error} present representative examples of the three outcome types.

MobileWorld-Real is designed to complement rather than replace sandbox benchmarks: it measures whether capabilities learned in controlled environments transfer to live devices. All MobileWorld-Real tasks and collected trajectories are held out from training. Since live Apps and account states change over time, we report environment errors separately and 
retain the complete execution trajectory for auditing and reproducibility.

\begin{table}[t]
\caption{
Performance comparison on real-device mobile benchmarks.
MobileWorld-Real is our proposed Chinese real-device mobile GUI benchmark
(Section~\ref{sec:MobileWorld-Real}), and AndroidDaily~\citep{androiddaily}
covers high-frequency daily scenarios on real Android devices.
}
\label{tab:realdevice_mobile}

\centering
\footnotesize
\renewcommand{\arraystretch}{1.18}

\begin{minipage}{0.95\linewidth}
\centering
\setlength{\tabcolsep}{3.5pt}

\begin{tabularx}{\linewidth}{
  >{\raggedright\arraybackslash}X
  >{\centering\arraybackslash}p{0.16\linewidth}
  >{\centering\arraybackslash}p{0.18\linewidth}
  >{\centering\arraybackslash}p{0.18\linewidth}
}
\toprule[1.2pt]

\textbf{\textsc{Model}} &
\textbf{\textsc{Access / Size}} &
\textbf{\textsc{MobileWorld-Real}} &
\textbf{\textsc{AndroidDaily}} \\
\midrule

\rowcolor{gray!15}
\multicolumn{4}{l}{\textit{Baselines}} \\

Seed 2.1 Pro~\citep{bytedanceseed2026seed21}
    & Closed-source
    & \cellcolor{light_purple}\underline{88.7}
    & \cellcolor{light_purple}\underline{95.2} \\

Gemini 3.1 Pro~\citep{google2026gemini31pro}
    & Closed-source
    & \cellcolor{light_purple}86.2
    & \cellcolor{light_purple}93.8 \\

GPT-5.6 Sol~\citep{openai2026gpt56}
    & Closed-source
    & \cellcolor{light_purple}85.4
    & \cellcolor{light_purple}92.6 \\

Claude Opus 4.8~\citep{anthropic2026claudeopus48}
    & Closed-source
    & \cellcolor{light_purple}84.7
    & \cellcolor{light_purple}93.0 \\

Qwen 3.7 Plus~\citep{qwen2026qwen37plus}
    & Closed-source
    & \cellcolor{light_purple}72.7
    & \cellcolor{light_purple}79.8 \\

Kimi K2.6~\citep{moonshot2026kimik26}
    & 1T-A32B
    & \cellcolor{light_purple}62.6
    & \cellcolor{light_purple}67.6 \\

PhoneBuddy-4B~\citep{tang2026phonebuddytrainingopenmodels}
    & 4B
    & \cellcolor{light_purple}53.5
    & \cellcolor{light_purple}69.0 \\

UI-Venus-1.5-30B-A3B~\citep{uivenus15}
    & 30B-A3B
    & \cellcolor{light_purple}33.0
    & \cellcolor{light_purple}61.7 \\

GUI-Owl-1.5-32B-Instruct~\citep{Mobile-agent-v3.5}
    & 32B
    & \cellcolor{light_purple}32.4
    & \cellcolor{light_purple}60.9 \\
    
GELab-Zero-4B-preview~\citep{step-gui}
    & 4B
    & \cellcolor{light_purple}31.3
    & \cellcolor{light_purple}73.4 \\

\midrule
\rowcolor{gray!15}
\multicolumn{4}{l}{Ours} \\

\modelname{}
    & 27B
    & \cellcolor[HTML]{cfcdfd}\textbf{92.2}
    & \cellcolor[HTML]{cfcdfd}\textbf{97.5} \\

\modelname{}
    & 35B-A3B
    & \cellcolor[HTML]{cfcdfd}{87.4}
    & \cellcolor[HTML]{cfcdfd}93.9 \\

\bottomrule[1.2pt]
\end{tabularx}
\end{minipage}
\end{table}

\subsection{Main Results}
\subsubsection{Mobile-Use Evaluation.}
\label{sec:exp_real_device_mobile}

\begin{figure}[!t]
    \centering
    \includegraphics[width=\linewidth]{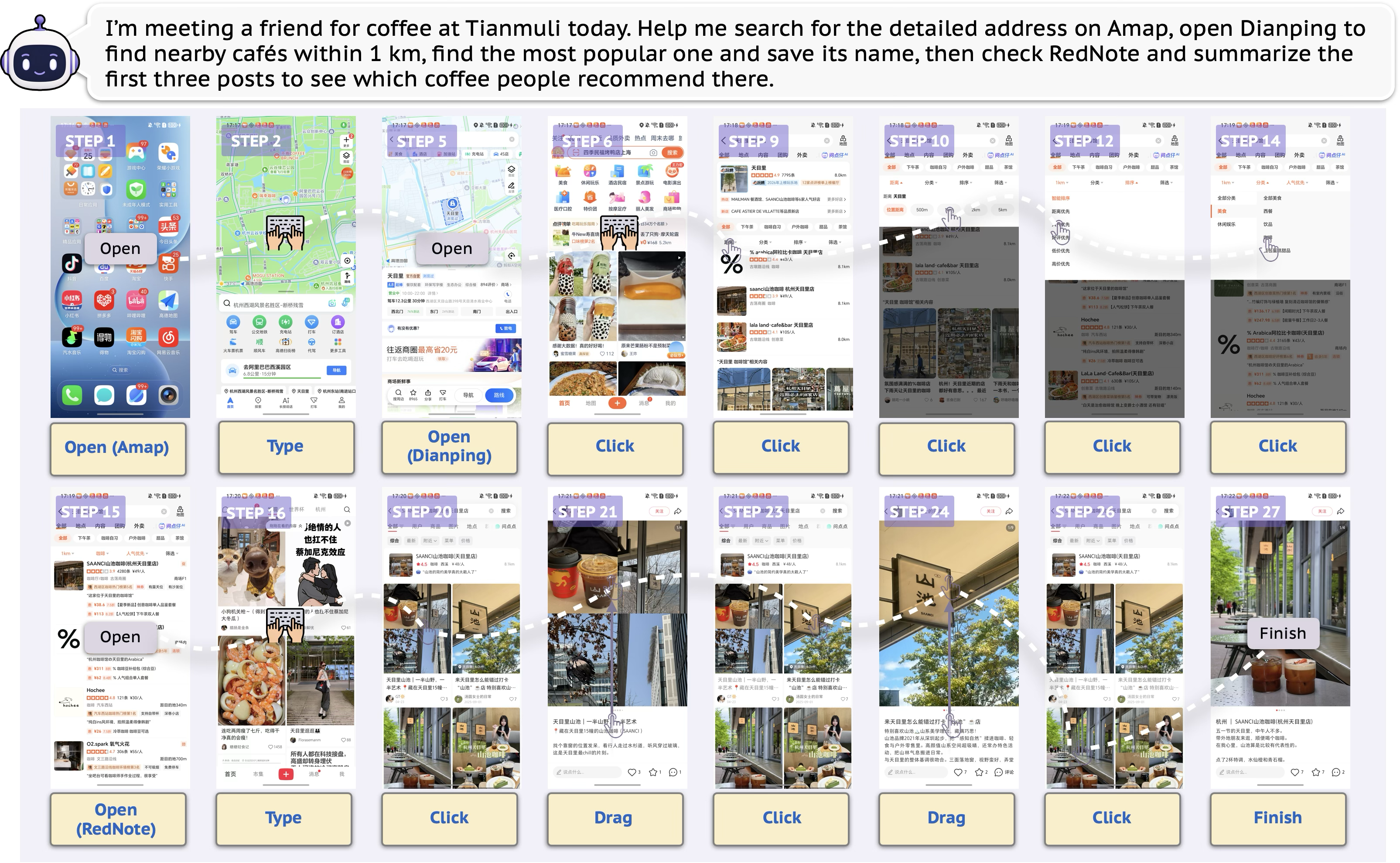}
    \caption{\textbf{Demonstration of real-device mobile GUI execution.} The trajectory is rendered as key frames with the executed action annotated beneath each frame. In this cross-app task, the agent looks up the target address on Amap, finds the most popular caf\'e nearby on Dianping, and posts a summary of the findings on RedNote.}
    \label{fig:real_device_tables}
\end{figure} 

We evaluate \modelname{} on the GUI-only subset of MobileWorld~\citep{kong2026mobileworld}, a challenging and reproducible mobile-use benchmark featuring long-horizon, cross-application workflows across diverse Android applications. The comparison includes both frontier general-purpose VLMs and specialized mobile GUI models from multiple organizations. As shown in Table~\ref{tab:mobileworld}, under the standard 50-step evaluation budget, \modelname{}-27B achieves a success rate of 82.1\%, establishing a new state of the art. It outperforms GPT 5.6 Sol, Opus 4.8, and Seed~2.1~Pro by 12.0, 14.6, and 8.9 percentage points, respectively. It surpasses the strongest specialized GUI baseline, GUI-Owl-1.5-32B-Instruct at 43.9\%, by 38.2 percentage points. \modelname{}-35B-A3B also achieves 65.0\%, outperforming all specialized GUI baselines and all general-purpose VLMs except Seed~2.1~Pro and Claude Opus~4.8. When the maximum number of steps is increased to 100, the success rates of \modelname{}-27B and \modelname{}-35B-A3B further rise to 85.5\% and 68.4\%, respectively. These results demonstrate our strong long-horizon planning, visual understanding, and reliable execution in complex mobile environments.

We further evaluate \modelname{} on two real-device mobile benchmarks: MobileWorld-Real (Section~\ref{sec:MobileWorld-Real}) and AndroidDaily~\citep{androiddaily}. MobileWorld-Real focuses on complex tasks executed on live Android devices in the Chinese mobile ecosystem, while AndroidDaily covers high-frequency daily scenarios on real Android devices.

As shown in Table~\ref{tab:realdevice_mobile}, \modelname{} achieves 92.2\% on MobileWorld-Real, surpassing all baselines, including the top closed-source models Seed~2.1~Pro at 88.7\%, Gemini~3.1~Pro at 86.2\%, and GPT 5.6 Sol at 85.4\%. This result shows that \modelname{} reliably handles long-horizon cross-app execution, deep entry-point navigation within super Apps, and reasoning over changing information on real-world mobile devices. With only 27B parameters, \modelname{} offers a favorable balance among task capability, real-device reliability, and deployment efficiency.

On AndroidDaily, \modelname{} achieves 97.5\%, ranking first among all compared models and outperforming Seed~2.1~Pro at 95.2\% and {Gemini~3.1~Pro} at 93.8\%, showing reliable task completion on high-frequency everyday mobile scenarios. The 35B-A3B variant attains 87.4\% on MobileWorld-Real and 93.9\% on AndroidDaily, while activating only 3B parameters per token, offering substantially higher inference efficiency for deployment.
\begin{table}[!t]
\centering
\captionsetup{skip=3pt}

\caption{Performance comparison on OSWorld-Verified~\citep{OSWorld}.}
\label{tab:osworld_verified}

{\footnotesize
\renewcommand{\arraystretch}{1.15}

\begin{minipage}{0.86\linewidth}
\centering
\setlength{\tabcolsep}{4.5pt}

\begin{tabular}{
>{\raggedright\arraybackslash}p{0.49\linewidth}
>{\centering\arraybackslash}p{0.20\linewidth}
>{\centering\arraybackslash}p{0.23\linewidth}
}
\toprule[1.2pt]
\textbf{\textsc{Model}} &
\textbf{\textsc{Access / Size}} &
\textbf{\textsc{Success Rate (\%)}} \\
\midrule

\rowcolor{gray!15}
\multicolumn{3}{l}{\textit{Baselines}} \\

Claude Opus 4.8~\citep{anthropic2026claudeopus48}
    & Closed-source
    & \cellcolor{light_purple}\textbf{83.4} \\
Seed 2.1 Pro~\citep{bytedanceseed2026seed21}
    & Closed-source
    & \cellcolor{light_purple}78.8 \\
GPT-5.5~\citep{openai2026gpt55}
    & Closed-source
    & \cellcolor{light_purple}78.7 \\
Gemini 3.5 Flash~\citep{google2026gemini35flash}
    & Closed-source
    & \cellcolor{light_purple}78.4 \\
Gemini 3.1 Pro~\citep{google2026gemini31pro}
    & Closed-source
    & \cellcolor{light_purple}76.2 \\
Qwen 3.7 Plus~\citep{qwen2026qwen37plus}
    & Closed-source
    & \cellcolor{light_purple}73.3 \\
MiniMax M3~\citep{minimax2026m3}
    & 428B-A23B
    & \cellcolor{light_purple}75.2 \\
Kimi K2.6~\citep{moonshot2026kimik26}
    & 1T-A32B
    & \cellcolor{light_purple}73.1 \\
GUI-Owl-1.5-32B-Instruct~\citep{Mobile-agent-v3.5}
    & 32B
    & \cellcolor{light_purple}56.5 \\

\midrule
\rowcolor{gray!15}
\multicolumn{3}{l}{Ours} \\

\modelname{}
    & 27B
    & \cellcolor[HTML]{cfcdfd}\underline{79.5} \\

\bottomrule[1.2pt]
\end{tabular}
\end{minipage}
\par
}

\vspace{0.5em}

\caption{Performance comparison on OSWorld-v2~\citep{osworld2}. Partial and Binary denote partial task progress and full task completion, respectively. Lower Steps/task is better. 
}
\label{tab:osworld_v2}

{\footnotesize
\setlength{\tabcolsep}{8pt}
\renewcommand{\arraystretch}{1.18}

\resizebox{\linewidth}{!}{%
\begin{tabular}{lcccc}
\toprule[1.2pt]
\textbf{\textsc{Model}} &
\textbf{\textsc{Partial (\%)}} &
\textbf{\textsc{Binary (\%)}} &
\textbf{\textsc{Steps/task}} &
\textbf{\textsc{Action Mode}} \\
\midrule

\rowcolor{gray!15}
\multicolumn{5}{l}{\textit{Baselines}} \\

Claude Opus 4.8~\citep{anthropic2026claudeopus48}
    & \cellcolor{light_purple}\textbf{54.8}
    & \cellcolor{light_purple}\textbf{20.6}
    & 103.0
    & Batched \\

GPT-5.5~\citep{openai2026gpt55}
    & \cellcolor{light_purple}\underline{49.5}
    & \cellcolor{light_purple}13.0
    & {95.2}
    & Batched \\

MiniMax M3~\citep{minimax2026m3}
    & \cellcolor{light_purple}22.3
    & \cellcolor{light_purple}4.6
    & 326.7
    & Single \\

Kimi K2.6~\citep{moonshot2026kimik26}
    & \cellcolor{light_purple}22.1
    & \cellcolor{light_purple}4.6
    & 179.3
    & Single \\

Qwen 3.7 Plus~\citep{qwen2026qwen37plus}
    & \cellcolor{light_purple}21.5
    & \cellcolor{light_purple}2.8
    & 173.5
    & Single \\

\midrule
\rowcolor{gray!15}
\multicolumn{5}{l}{Ours} \\

\modelname{}
    & \cellcolor[HTML]{cfcdfd}40.0
    & \cellcolor[HTML]{cfcdfd}\underline{13.9}
    & {135.8}
    & Batched \\

\bottomrule[1.2pt]
\end{tabular}%
}
\par
}

\end{table}

\vspace{-0.5em}
\paragraph{Qualitative Examples.}
Figure~\ref{fig:real_device_tables} presents representative real-device execution trajectory of \modelname{}, with the executed action annotated beneath each key frame. It highlights autonomous planning in a cross-app scenario: over 27 steps, the agent navigates directly to the deep function entry points of each app, retrieves the address on Amap, identifies the most popular caf\'e within 1\,km on Dianping, and carries the extracted key information into RedNote as the decision basis for completing the summary.

\subsubsection{Computer-Use Evaluation}\label{sec:exp_computer_use}
We evaluate the computer-use capability of \modelname{} on OSWorld-Verified \citep{OSWorld} and OSWorld-v2 \citep{osworld2}. We report the two benchmarks separately. OSWorld-Verified measures partial progress for 361 computer-use tasks, whereas OSWorld-v2 focuses on longer-horizon workflows and reports both partial progress and binary completion.

As shown in Table~\ref{tab:osworld_verified}, \modelname{} achieves a success rate of 79.5\% on OSWorld-Verified, ranking second among all compared models and trailing only Claude Opus 4.8. It outperforms other leading closed-source foundation models, including Seed 2.1 Pro, GPT-5.5, Gemini 3.5 Flash, and Gemini 3.1 Pro, as well as all evaluated open-weight foundation models. These results demonstrate that \modelname{} is competitive with the strongest closed-source models while establishing a substantial advantage over current open-weight models.

\begin{figure}[!t]
    \centering

    \includegraphics[width=\linewidth]{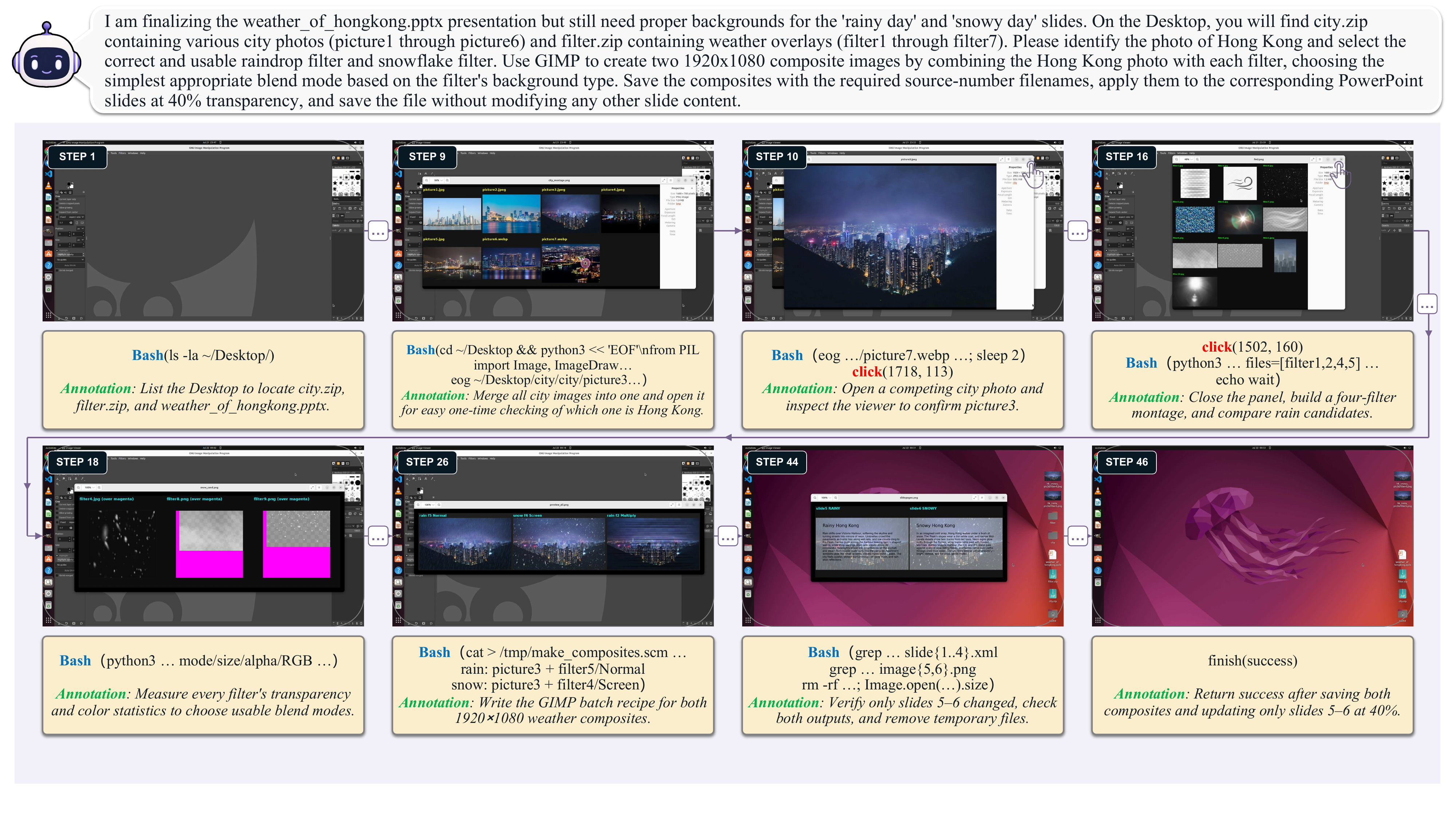}

    \caption{\textbf{Demonstration of hybrid GUI+CLI execution in computer-use tasks.} Key frames are shown with the emitted actions, where CLI commands are highlighted in blue, GUI actions in red, and each step carries a short annotation of its intent.
In this example, the agent selects the target photo and filters by stitching all candidates into a single montage via CLI for one-shot visual inspection, rather than opening each image individually. It then leverages hybrid execution to combine CLI-based processing with GUI-based verification for the final output.}
    \label{fig:cua_demo_a}
\end{figure}

As shown in Table~\ref{tab:osworld_v2}, \modelname{} achieves a partial score of 40.0\% and a binary completion rate of 13.9\% on OSWorld-v2, placing it competitively among frontier closed-source models. In particular, it exceeds GPT-5.5 by 0.9 percentage points in binary completion, while trailing it by 9.5 points in partial score. \modelname{} also substantially outperforms all evaluated open-weight models. Compared with the strongest open-weight baseline, MiniMax M3, \modelname{} improves partial progress by 17.7 percentage points and binary completion by 9.3 points. The margins increase to 17.9/9.3 points over Kimi K2.6 and 18.5/11.1 points over Qwen 3.7 Plus. Additionally, \modelname{} requires an average of only 135.8 steps per task, substantially fewer than MiniMax M3 and Qwen 3.7 Plus. These results indicate that the hybrid GUI+CLI and batched-action design provides both strong task performance and execution efficiency on long-horizon computer-use workflows. 

A key feature of \modelname{} is its unified GUI+CLI action space with batched execution. Rather than treating CLI as an occasional fallback, the model learns to select and interleave GUI and CLI actions within the same trajectory. On OSWorld-Verified and OSWorld-v2, CLI actions constitute \textbf{40.7\%} and \textbf{55.1\%} of all actions, respectively, while the corresponding proportions of actions emitted in batched form are \textbf{39.6\%} and \textbf{41.6\%}. We provide a detailed analysis of how \modelname{} combines GUI and CLI actions in Section~\ref{sec:analysis-cua}.
\vspace{-0.5em}
\paragraph{Qualitative Examples.} 
Figure~\ref{fig:cua_demo_a} and Figure~\ref{fig:cua_demo_b} of Appendix~\ref{app:examples} further illustrate the hybrid execution behavior of \modelname{} in two representative multi-application workflows, where CLI commands (blue) and GUI actions (red) are highlighted beneath each key frame together with a short annotation of the step intent. In Figure~\ref{fig:cua_demo_a}, the agent must identify the target city photo and usable weather filters among many candidate images; pure GUI execution would require opening and inspecting every image individually, whereas the agent stitches all candidates into a single montage with one CLI script and verifies them at a single glance, before returning to GUI actions for compositing and presentation editing. Such complementary use of the two interfaces makes hybrid execution substantially more efficient than either GUI or CLI alone, and the agent further batches compatible operations, such as multi-key hotkey sequences or a command followed by a wait, into a single model turn.

\begin{table}[t]
\caption{Performance comparison on WebArena~\citep{zhou2023webarena}. 
Baseline results marked with $^*$ are obtained from our own evaluation under the same experimental setup.
}
\label{tab:webarena}
\centering
\footnotesize
\renewcommand\arraystretch{1.05}

\begin{minipage}{0.86\linewidth}
\centering
\setlength{\tabcolsep}{4.5pt}

\begin{tabular}{
>{\raggedright\arraybackslash}p{0.5\linewidth}
>{\centering\arraybackslash}p{0.20\linewidth}
>{\centering\arraybackslash}p{0.20\linewidth}
}
\toprule[1.2pt]
\textbf{\textsc{Model}} &
\textbf{\textsc{Access / Size}} &
\textbf{\textsc{Success Rate (\%)}} \\
\midrule
\rowcolor{gray!15}
\multicolumn{3}{l}{\textit{Baselines}} \\
Claude Opus 4.8$^*$~\citep{anthropic2026claudeopus48}
    & Closed-source
    & \cellcolor{light_purple}\underline{71.9}\\
GPT-5.5$^*$~\citep{openai2026gpt55}
    & Closed-source
    & \cellcolor{light_purple}{69.5} \\
Gemini 3.1 Pro$^*$~\citep{gemini3pro}
    & Closed-source
    & \cellcolor{light_purple}{65.3} \\
Qwen 3.7 Plus$^*$~\citep{qwen2026qwen37plus}
    & Closed-source
    & \cellcolor{light_purple}{59.0} \\
\midrule
CUA-GYM-A17B~\citep{cuagym}
    & 397B-A17B
    & \cellcolor{light_purple}56.0 \\
Kimi K2.6$^*$~\citep{moonshot2026kimik26}
    & 1T-A32B
    & \cellcolor{light_purple}{55.8} \\
Qwen3.5-397B-A17B~\citep{qwen3.5}
    & 397B-A17B
    & \cellcolor{light_purple}54.0 \\
GUI-Owl-1.5-32B-Thinking~\citep{Mobile-agent-v3.5}
    & 32B
    & \cellcolor{light_purple}48.4 \\
Qwen3.5-27B~\citep{qwen3.5}
    & 27B
    & \cellcolor{light_purple}41.5 \\
Qwen3.5-35B-A3B~\citep{qwen3.5}
    & 35B-A3B
    & \cellcolor{light_purple}40.8 \\
\midrule
\rowcolor{gray!15}
\multicolumn{3}{l}{Ours} \\
\modelname{}
    & 27B
    & \cellcolor[HTML]{cfcdfd}\textbf{73.6} \\
    \modelname{}
    & 35A3B
    & \cellcolor[HTML]{cfcdfd}69.2 \\
\midrule
\textit{Human Performance}~\citep{zhou2023webarena}
    & --
    & 78.2 \\
\bottomrule[1.2pt]

\end{tabular}

\end{minipage}
\end{table}

\subsubsection{Browser-Use and DeepSearch Evaluation}
\label{sec:exp_web_deepsearch}

Beyond mobile and computer-use, we evaluate \modelname{} on browser-based web navigation and DeepSearch. 
These capabilities are closely related but assess different aspects of an GUI agent.

\paragraph{Browser Use.}
We evaluate \modelname{} on WebArena~\citep{zhou2023webarena}, which consists of stateful, multi-step tasks over functional websites. During an initial evaluation, we identified multiple incorrect reference answers and errors in the official evaluation scripts. We therefore manually verified the reference answers and corrected the affected scripts before reporting the final results. Kimi K2.5 likewise reports its WebArena performance using a corrected evaluation setup~\citep{kimik25}.

As shown in Table~\ref{tab:webarena}, the 27B variant of \modelname{} achieves a success rate of 73.6\%, the highest among all compared models, while the 35A3B variant reaches 69.2\%. The 27B variant surpasses the strongest baseline, Claude Opus~4.8 at 71.9\%, by 1.7 percentage points, and exceeds GPT-5.5 at 69.5\% by 4.1 percentage points. \modelname{} also outperforms Gemini~3.1~Pro, Qwen~3.7~Plus, Kimi~K2.6, CUA-GYM-A17B, and specialized browser agents such as Claude~3.7 Computer Use and GUI-Owl-1.5-32B-Thinking. Although a 4.6-point gap remains to the reported human performance of 78.2\%, the result demonstrates strong end-to-end browser control across heterogeneous web interfaces.

\begin{table}[t]
\caption{Performance comparison on DeepSearch benchmarks: BrowseComp (BC) and BrowseComp-ZH (BC-ZH). ``--'' indicates the result is not reported. 
}
\label{tab:deepsearch}
\centering
\footnotesize
\renewcommand\arraystretch{1.15}

\begin{minipage}{0.86\linewidth}
\centering
\setlength{\tabcolsep}{4.5pt}

\begin{tabular}{
>{\raggedright\arraybackslash}p{0.46\linewidth}
>{\centering\arraybackslash}p{0.20\linewidth}
>{\centering\arraybackslash}p{0.13\linewidth}
>{\centering\arraybackslash}p{0.13\linewidth}
}
\toprule[1.2pt]
\textbf{\textsc{Model}} &
\textbf{\textsc{Access / Size}} &
\textbf{\textsc{BC (\%)}} &
\textbf{\textsc{BC-ZH (\%)}} \\
\midrule
\rowcolor{gray!15}
\multicolumn{4}{l}{\textit{Baselines}} \\
GPT-5.5 ~\citep{openai2026gpt55}
    & Closed-source
    & \cellcolor{light_purple}\underline{90.1}
    & \cellcolor{light_purple}-- \\
Seed 2.1 Pro~\citep{bytedanceseed2026seed21}
    & Closed-source
    & \cellcolor{light_purple}86.2
    & \cellcolor{light_purple}-- \\
Gemini 3.1 Pro~\citep{google2026gemini31pro}
    & Closed-source
    & \cellcolor{light_purple}85.9
    & \cellcolor{light_purple}-- \\
Claude Opus 4.8~\citep{anthropic2026claudeopus48}
    & Closed-source
    & \cellcolor{light_purple}84.3
    & \cellcolor{light_purple}-- \\
UI-TARS-2~\citep{uitars2}
    & Closed-source
    & \cellcolor{light_purple}29.6
    & \cellcolor{light_purple}{50.5} \\    
    \midrule
Qwen3.5-397B-A17B~\citep{qwen3.5}
    & 397B-A17B & \cellcolor{light_purple}78.6
    & \cellcolor{light_purple}70.3 \\
Apodex-1.0-mini~\citep{apodex2026}
    & 35B-A3B
    & \cellcolor{light_purple}71.5
    & \cellcolor{light_purple}\textbf{80.6} \\
Qwen3.5-27B~\citep{qwen3.5}
    & 27B
    & \cellcolor{light_purple}61.0
    & \cellcolor{light_purple}{62.1} \\
GLM-4.7~\citep{glm47}
    & 358B
    & \cellcolor{light_purple}52.0
     & \cellcolor{light_purple}{66.6} \\
DeepSeek-V3.2~\citep{deepseekv32}
    & 685B
    & \cellcolor{light_purple}51.4
    & \cellcolor{light_purple}{65.0} \\
Tongyi-DR-30B~\citep{tongyidr}
    & 30B-A3B
    & \cellcolor{light_purple}43.4
    & \cellcolor{light_purple}46.7 \\

\midrule
\rowcolor{gray!15}
\multicolumn{4}{l}{Ours} \\
\modelname{}
    & 27B
    & \cellcolor[HTML]{cfcdfd}{64.1}
    & \cellcolor[HTML]{cfcdfd}\underline{75.0} \\
\bottomrule[1.2pt]
\end{tabular}

\end{minipage}
\end{table}

\paragraph{DeepSearch.}
Simple search is often part of web navigation itself: an agent enters a query into a browser, inspects the ranked results, and opens a potentially relevant page. DeepSearch simplifies this process by replacing many repetitive GUI actions with API-based retrieval and evidence synthesis, allowing the agent to obtain external context more efficiently before continuing the GUI workflow.

We evaluate \modelname{} on BrowseComp and BrowseComp-ZH, which assess persistent information seeking over the English and Chinese web, respectively. As shown in Table~\ref{tab:deepsearch}, \modelname{} obtains 64.1\% on BrowseComp and 75.0\% on BrowseComp-ZH. On BrowseComp, \modelname{} outperforms Qwen3.5-27B, Tongyi-DR-30B, and UI-TARS-2. While its performance remains behind frontier proprietary models and substantially larger systems, it demonstrates competitive English DeepSearch capability at the 27B scale. On BrowseComp-ZH, \modelname{} achieves the second-best result among all compared systems, substantially outperforming Qwen3.5-27B, Tongyi-DR-30B, and UI-TARS-2. These results show that \modelname{} is capable not only of locating candidate webpages, but also synthesizing evidence distributed across multiple sources into reliable, task-relevant information for subsequent reasoning and GUI execution.

\begin{figure}[!t]
    \centering
    \includegraphics[width=\textwidth]{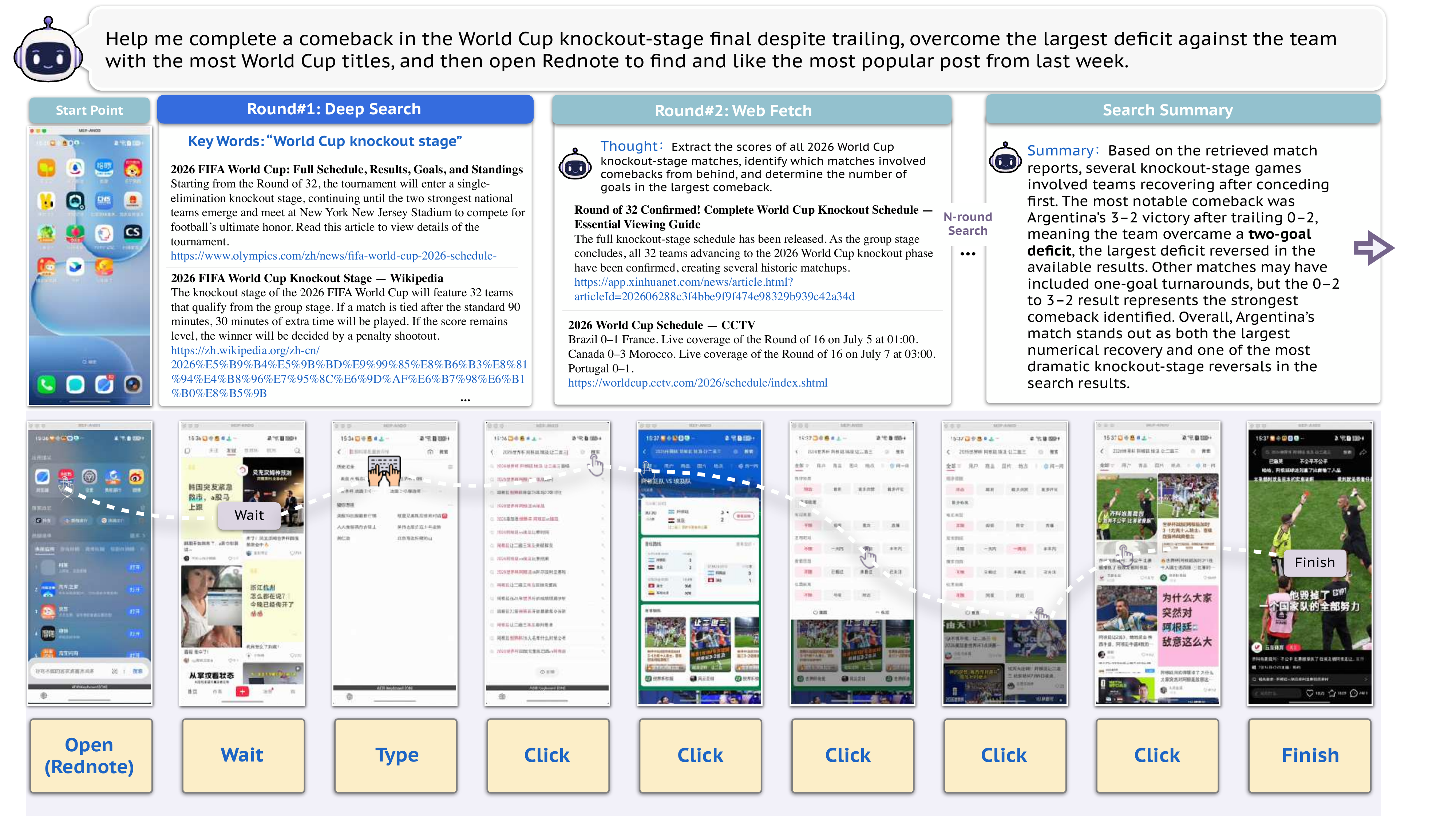}
    \caption{\textbf{Demonstration of DeepSearch-assisted GUI execution.} The upper panels show the multi-round search process, including keyword-based DeepSearch, targeted web fetches with intermediate thoughts, and the final search summary; the lower row shows the subsequent GUI trajectory. DeepSearch resolves the knowledge and reasoning problem before GUI execution, converting the vague cross-source request into an explicit target: \modelname{} identifies the largest comeback in the World Cup knockout stage through DeepSearch, then opens RedNote and navigates directly to the highest-engagement related post from the past week without blind in-app searching.}
    \label{fig:deepsearch_to_gui}
\end{figure}

\paragraph{Qualitative Examples.} We further present two GUI execution demos illustrating complementary ways in which DeepSearch integrates with GUI execution: GUI actions handle the interaction with applications, while DeepSearch handles external knowledge acquisition, comparison, and decision making. In the first demo (Figure~\ref{fig:deepsearch_to_gui}), the task requires solving a knowledge and reasoning problem before any GUI operation: identifying the knockout-stage match with the largest comeback in the current World Cup. The upper panels trace the multi-round search process, from keyword-based DeepSearch over ranked sources, through targeted web fetches that extract and cross-check match scores, to a search summary that identifies Argentina's 3--2 comeback after trailing 0--2. DeepSearch thereby converts a vague, cross-source user request into an explicit GUI target before execution begins, so that in the trajectory below the agent opens RedNote, searches for the identified match directly, filters posts from the previous seven days, and opens the most discussed post without blind in-app searching or trial and error. In the second demo (Figure~\ref{fig:deepsearch_to_gui2} in Appendix~\ref{app:examples}), DeepSearch is instead invoked on demand in the middle of GUI execution: the agent first locates a Douyin video on evidence-based weight loss and extracts its main claims through GUI actions, then calls DeepSearch to verify these claims against research papers, official guidelines, and authoritative sources, since whether and what to comment cannot be decided from the interface alone. The verification conclusions directly determine the subsequent GUI actions, namely posting a comment that corrects the misleading claims and supplements the necessary caveats, forming a closed observe--research--decide--act loop.

\subsubsection{GUI Grounding}

\begin{table*}[t]
\caption{Performance comparison on GUI grounding benchmarks. ScreenSpot-Pro reports no-zoom results, with zoom-in results shown in parentheses when available. Baseline results marked with $^{*}$ are obtained from our own evaluation. 
}
\label{tab:grounding_5bench}
\centering
\scriptsize
\setlength{\tabcolsep}{2.6pt}
\renewcommand\arraystretch{1.12}
\providecommand{\bccell}[1]{\cellcolor{light_purple}#1}
\providecommand{\occell}[1]{\cellcolor[HTML]{cfcdfd}#1}
\resizebox{\textwidth}{!}{%
\begin{tabular}{lccccc}
\toprule[1.2pt]
\multicolumn{1}{c}{\textbf{\textsc{Models}}} &
\multicolumn{5}{c}{\textbf{\textsc{Grounding Benchmarks}}} \\
\cmidrule(lr){2-6}
 &
\textbf{\textsc{SS-Pro(zoom in)}} &
\textbf{\textsc{SS-V2}} &
\textbf{\textsc{MM-GUI-L2}} &
\textbf{\textsc{OSW-G-R}} &
\textbf{\textsc{UI-Vision}} \\
\midrule
\rowcolor{gray!15}
\multicolumn{6}{l}{\textit{Baselines}} \\
Qwen 3.7 Plus*~\citep{qwen2026qwen37plus} & \bccell{68.9 (79.0)} & \bccell{96.6} & \bccell{90.5} & \bccell{\underline{78.2}} & \bccell{\underline{68.0}} \\
Seed 2.1 Pro*~\citep{bytedanceseed2026seed21} & \bccell{65.3 (\underline{80.7})} & \bccell{96.6} & \bccell{90.9} & \bccell{78.0} & \bccell{62.0} \\
Qwen3.5-4B*~\citep{qwen3.5} & \bccell{59.1 (60.3)} & \bccell{94.3} & \bccell{83.4} & \bccell{62.0} & \bccell{32.6} \\
Qwen3.5-35B-A3B*~\citep{qwen3.5} & \bccell{64.5 (68.6)} & \bccell{95.2} & \bccell{87.5} & \bccell{63.6} & \bccell{52.3} \\
Qwen3.5-27B*~\citep{qwen3.5} & \bccell{68.4 (70.3)} & \bccell{96.1} & \bccell{89.1} & \bccell{67.9} & \bccell{46.6} \\
GUI-Owl-1.5-2B-Instruct~\citep{Mobile-agent-v3.5} & \bccell{57.8 (70.4)} & \bccell{89.7} & \bccell{72.1} & \bccell{62.6} & \bccell{--} \\
GUI-Owl-1.5-4B-Instruct~\citep{Mobile-agent-v3.5} & \bccell{66.8 (75.6)} & \bccell{93.2} & \bccell{83.2} & \bccell{68.4} & \bccell{--} \\
GUI-Owl-1.5-8B-Instruct~\citep{Mobile-agent-v3.5} & \bccell{71.1 (77.8)} & \bccell{93.7} & \bccell{82.5} & \bccell{69.3} & \bccell{--} \\
GUI-Owl-1.5-32B-Instruct~\citep{Mobile-agent-v3.5} & \bccell{72.9 (80.3)} & \bccell{95.3} & \bccell{86.8} & \bccell{69.7} & \bccell{--} \\
UI-Venus-1.5-2B~\citep{uivenus15} & \bccell{57.7 (64.6)} & \bccell{92.8} & \bccell{80.3} & \bccell{59.4} & \bccell{44.8} \\
UI-Venus-1.5-8B~\citep{uivenus15} & \bccell{68.4 (73.9)} & \bccell{95.9} & \bccell{88.1} & \bccell{69.7} & \bccell{46.5} \\
UI-Venus-1.5-30B-A3B~\citep{uivenus15} & \bccell{69.6 (74.8)} & \bccell{96.2} & \bccell{88.6} & \bccell{70.6} & \bccell{54.7} \\
ZoomOnce-4B~\citep{liu2026forwardbeatstwoinnerzoom} & \bccell{66.2} & \bccell{95.2} & \bccell{87.6} & \bccell{73.1} & \bccell{40.2} \\
HyMobileAgent-A3B~\citep{hymobileagent} & \bccell{66.5} & \bccell{96.2} & \bccell{89.3} & \bccell{--} & \bccell{--} \\
MAI-UI-2B~\citep{maiui} & \bccell{57.4 (62.8)} & \bccell{92.5} & \bccell{82.6} & \bccell{63.5} & \bccell{30.3} \\
MAI-UI-8B~\citep{maiui} & \bccell{65.8 (70.9)} & \bccell{95.2} & \bccell{88.8} & \bccell{72.9} & \bccell{40.7} \\
MAI-UI-32B~\citep{maiui} & \bccell{67.9 (73.5)} & \bccell{96.5} & \bccell{91.3} & \bccell{75.0} & \bccell{47.1} \\
GTA1-7B~\citep{GTA1} & \bccell{50.1} & \bccell{92.4} & \bccell{78.5} & \bccell{67.7} & \bccell{--} \\
GTA1-32B~\citep{GTA1} & \bccell{63.6} & \bccell{95.2} & \bccell{83.4} & \bccell{72.2} & \bccell{--} \\
UI-Ins-7B~\citep{uiins} & \bccell{52.2} & \bccell{94.0} & \bccell{83.1} & \bccell{-} & \bccell{-} \\
UI-Ins-32B~\citep{uiins} & \bccell{57.0} & \bccell{94.9} & \bccell{84.9} & \bccell{-} & \bccell{-} \\
OpenCUA-7B~\citep{opencua} & \bccell{50.0} & \bccell{92.3} & \bccell{--} & \bccell{--} & \bccell{29.7} \\
OpenCUA-32B~\citep{opencua} & \bccell{55.3} & \bccell{93.4} & \bccell{--} & \bccell{--} & \bccell{33.3} \\
OpenCUA-72B~\citep{opencua} & \bccell{60.8} & \bccell{92.9} & \bccell{--} & \bccell{--} & \bccell{37.3} \\
\midrule
\rowcolor{gray!15}
\multicolumn{6}{l}{\textit{Ours}} \\
\modelname{}-4B & \occell{67.8 (74.0)} & \occell{94.9} & \occell{87.9} & \occell{70.5} & \occell{51.6} \\
\modelname{}-35B-A3B & \occell{\underline{76.1} (80.2)} & \occell{\underline{96.7}} & \occell{\underline{92.0}} & \occell{74.6} & \occell{65.9} \\
\modelname{}-27B & \occell{\textbf{76.6} (\textbf{81.5})} & \occell{\textbf{97.5}} & \occell{\textbf{92.6}} & \occell{\textbf{78.5}} & \occell{\textbf{70.0}} \\
\bottomrule[1.2pt]
\end{tabular}%
}
\end{table*}

We evaluate the GUI grounding capabilities of \modelname{} across five comprehensive benchmarks: ScreenSpot-Pro, ScreenSpot-V2, MMBench-GUI L2, {OSWorld-G-Refined}, and UI-Vision. 
These benchmarks cover high-resolution professional software, general cross-platform interfaces, desktop software environments, and instruction-driven grounding. 

As shown in Table~\ref{tab:grounding_5bench}, the 27B variant achieves strong grounding performance across all five benchmarks, reaching 76.6\% on ScreenSpot-Pro without zoom-in, 97.5\% on ScreenSpot-V2, 92.6\% on MMBench-GUI L2, {78.5\% on OSWorld-G-Refined, and 70.0\% on UI-Vision}. On ScreenSpot-Pro, it ranks first among all compared models {both without and with zoom-in}, and its score further increases to {81.5\%} with zoom-in. These results demonstrate accurate element localization across both common interfaces and complex professional software, including settings where the target must be inferred from functional or spatial context rather than explicit text.

The grounding capability also remains strong across model scales. On ScreenSpot-Pro, the 35B-A3B and 4B variants achieve 76.1\% and 67.8\% without zoom-in, with their scores increasing to 80.2\% and 74.0\% with zoom-in. The 35B-A3B variant further reaches 96.7\% on ScreenSpot-V2, 92.0\% on MMBench-GUI L2, {74.6\% on OSWorld-G-Refined}, and 65.9\% on UI-Vision, while activating only 3B parameters per token. The corresponding scores of the 4B variant are 94.9\%, 87.9\%, {70.5\%}, and 51.6\%. Together, these results show that \modelname{} provides reliable GUI grounding across common interfaces, professional software, and instruction types, from compact to moderate model scales.

\subsubsection{General and Agentic Capabilities}
\label{sec:exp-general-agentic}

\begin{table}[t]
    \centering
    \caption{General and agentic performance comparison. Qwen-UI-Agent acquires strong GUI capabilities while preserving the general reasoning and agentic capabilities of its base model. It also substantially outperforms other GUI-specialized models across these benchmarks.}
    \label{tab:general-agentic-capabilities}
    \setlength{\tabcolsep}{5.2pt}
    \renewcommand{\arraystretch}{1.12}
    \begin{adjustbox}{max width=\textwidth}
    \begin{tabular}{lcccccc}
        \toprule
        \textbf{Benchmark}
        & \cellcolor{light_purple}\textbf{Qwen-UI-Agent}
        & \textbf{Qwen3.5-27B}
        & \makecell{\textbf{UI-Venus-1.5}\\\textbf{30B-A3B}}
        & \makecell{\textbf{GUI-Owl-1.5}\\\textbf{32B-Instruct}}
        & \makecell{\textbf{EvoCUA-32B}\\\textbf{20260105}}
        & \makecell{\textbf{OpenCUA}\\\textbf{72B}} \\
        \midrule

        \rowcolor{black!6}
        \multicolumn{7}{l}{\textbf{\textit{General capabilities:}}} \\

        MMMU-Pro
        & \cellcolor{light_purple}72.4
        & \textbf{73.5}
        & 32.4
        & 39.5
        & 58.4
        & 31.0 \\

        RealWorldQA
        & \cellcolor{light_purple}\textbf{83.1}
        & \textbf{83.1}
        & 75.3
        & 76.7
        & 75.4
        & 66.4 \\

        CharXiv-RQ
        & \cellcolor{light_purple}\textbf{77.7}
        & 76.8
        & 44.7
        & 50.9
        & 58.1
        & 39.6 \\

        MathVision
        & \cellcolor{light_purple}\textbf{82.8}
        & 82.0
        & 36.8
        & 50.6
        & 60.8
        & 26.6 \\

        AI2D\_TEST
        & \cellcolor{light_purple}91.1
        & \textbf{91.9}
        & 84.3
        & 84.8
        & 85.7
        & 78.9 \\

        MMLU-Pro
        & \cellcolor{light_purple}\textbf{86.5}
        & 86.0
        & 65.6
        & 73.9
        & 77.1
        & 58.8 \\

        IFEval (prompt-level strict)
        & \cellcolor{light_purple}90.2
        & \textbf{90.4}
        & 81.3
        & 84.5
        & 64.0
        & 70.6 \\

        \midrule
        \rowcolor{black!6}
        \multicolumn{7}{l}{\textbf{\textit{Agentic capabilities:}}} \\

        Tau2-Bench
        & \cellcolor{light_purple}\textbf{89.9}
        & 89.2
        & 22.7
        & 6.1
        & 48.9
        & 14.4 \\

        Terminal-Bench 2.0 (Avg\,5)
        & \cellcolor{light_purple}\textbf{50.1}
        & 41.1
        & 3.2
        & 0.0
        & 5.6
        & 9.0 \\

        Claw-Eval (Avg\,3)
        & \cellcolor{light_purple}\textbf{73.5}
        & 66.9
        & 30.6
        & 29.6
        & 46.3
        & 26.4 \\

        Claw-Eval (Pass@3)
        & \cellcolor{light_purple}\textbf{51.8}
        & 41.2
        & 5.5
        & 5.5
        & 6.5
        & 0.5 \\

        BFCL-v4
        & \cellcolor{light_purple}\textbf{74.2}
        & 71.3
        & 19.8
        & 32.7
        & 48.8
        & 28.3 \\

        SkillsBench (Avg\,5)
        & \cellcolor{light_purple}\textbf{28.0}
        & 24.9
        & 0.5
        & 0.3
        & 3.3
        & 0.0 \\

        QwenClawBench (Avg\,3)
        & \cellcolor{light_purple}44.2
        & \textbf{48.5}
        & 6.4
        & 5.1
        & 18.6
        & 11.4 \\

        \bottomrule
    \end{tabular}
    \end{adjustbox}

    \vspace{0.35em}
    \begin{minipage}{\textwidth}
    \scriptsize
    \raggedright
    \smallskip
    \smallskip
    \textit{Evaluation note.} All results in this table are independently reproduced in our evaluation environment. The scores may differ from official reports as some evaluation settings are not identical. Unless noted below, we follow the corresponding official benchmark protocol.
    \smallskip
    \begin{itemize}[leftmargin=1.35em,labelsep=0.45em,itemsep=0pt,topsep=0.15em,parsep=0pt]
        \item \textbf{Tau2-Bench.} We follow the official leaderboard task set, harness, and scoring protocol, but use GPT-5.5 as the user simulator and judge.
        \item \textbf{Terminal-Bench 2.0.} Following the Qwen3.6 evaluation setup, we use the Harbor/Terminus-2 harness with a three-hour timeout and up to 32 CPUs and 48 GB of memory per task. We use temperature 1.0, top-$p$ 0.95, top-$k$ 20, an 80K-token output limit, and a 256K context window, and report the average over five runs.
        \item \textbf{Claw-Eval.} We follow the official 199-task, three-trial protocol and scoring rule, but use GPT-5.5 for both general and multi-turn judging and Qwen3.6-Plus for multi-turn user simulation, replacing the official Gemini 3 Flash/Claude Opus 4.6 setup. We use a sampling temperature of 0.6.
        \item \textbf{SkillsBench.} We follow the Qwen3.6 evaluation setup. We use OpenCode to evaluate 78 self-contained tasks, excluding tasks that depend on external APIs, and report the average over five runs.
        \item \textbf{QwenClawBench.} We follow the official v1.1 task set, three-run protocol, and penalized-hybrid scoring, but replace the default Claude Opus 4.5 judge with GPT-5.2. We also pin the OpenClaw runtime to version 2026.6.1 rather than the moving \texttt{main} image.
    \end{itemize}
    \end{minipage}
\end{table}

\modelname{} develops strong GUI domain knowledge while retaining the general reasoning and agentic capabilities inherited from its base model. As shown in Table~\ref{tab:general-agentic-capabilities}, it also substantially outperforms specialized GUI models on general and agentic benchmarks. Together, these results indicate that \modelname{} remains a broadly capable model rather than becoming a narrow GUI-only model. This broader capability profile may also help it handle out-of-distribution or unusual tasks that require knowledge, reasoning, and tool use beyond routine GUI interaction.

\subsection{Harness-Enabled Workflows}
\label{sec:exp_harness_workflows}

\begin{figure}[!t]
    \centering

    \includegraphics[width=\textwidth]{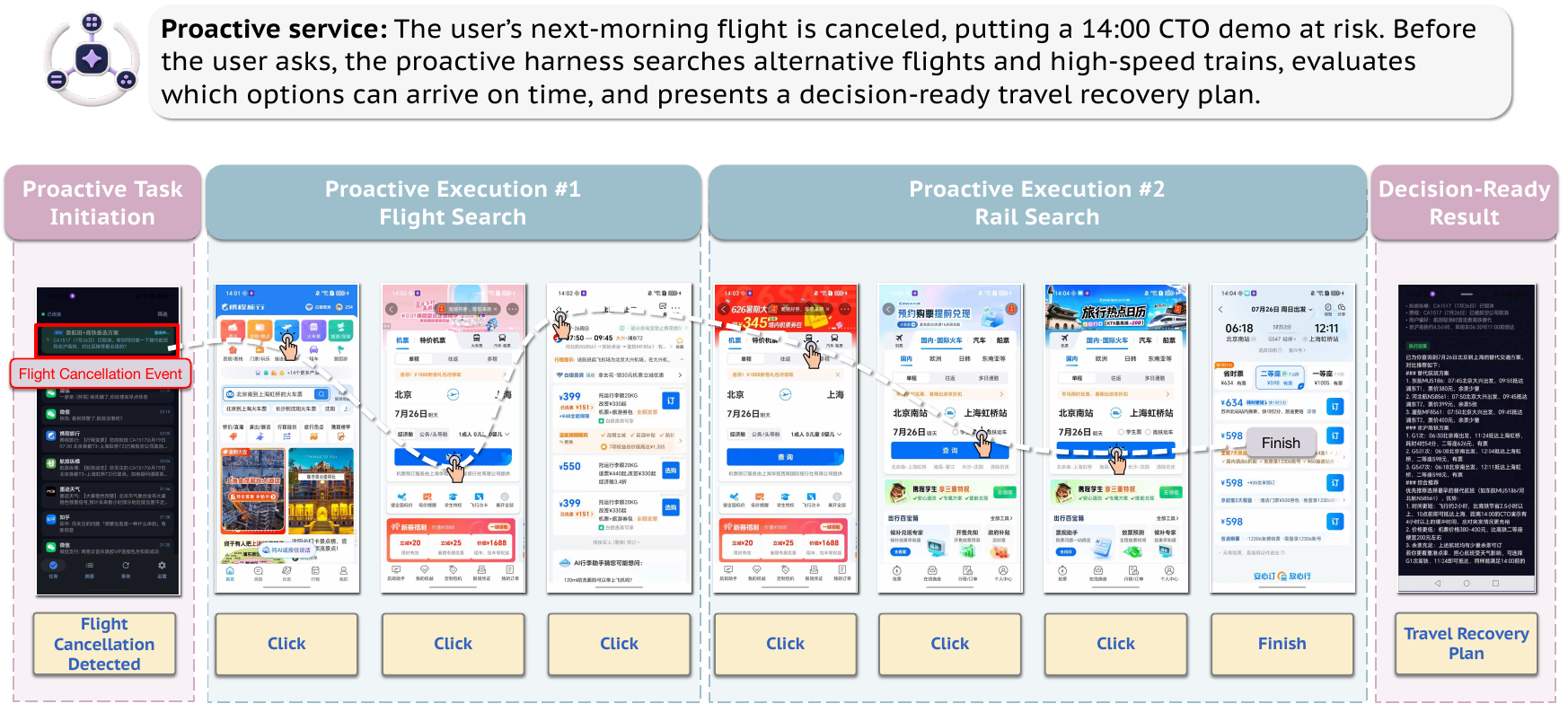}

    \caption{\textbf{Demonstration of proactive service based on mobile notifications.} The trajectory is organized into highlighted stages: proactive task initiation from a detected notification, proactive execution phases, and a decision-ready result, with the executed action annotated beneath each key frame. Flight-cancellation recovery: when the user's next-morning flight is canceled and a 14:00 demo is at risk, the harness proactively searches alternative flights and high-speed trains, evaluates which options arrive on time, and presents a decision-ready travel recovery plan.}
    \label{fig:proactive_demo_a}
\end{figure}

Beyond benchmark evaluations, we qualitatively examine the workflows enabled by the harness layer introduced in Section~\ref{sec:harness}, covering proactive service initiation and cross-platform task execution.

\paragraph{Proactive Service.} 
We present two representative proactive-service trajectories in Figure~\ref{fig:proactive_demo_a} and Figure~\ref{fig:proactive_demo_b} of Appendix~\ref{app:examples}. Each trajectory is organized into four highlighted stages: proactive task initiation, one or more proactive execution phases, and a decision-ready result. In Figure~\ref{fig:proactive_demo_a}, a scheduled 08:00 trigger combines the user's morning commitments, extracted from the highlighted notification, with live weather and commute conditions. The harness first retrieves the weather, then reasons over public-transit and taxi options through map navigation, and delivers an actionable morning brief with umbrella advice, departure timing, and key reminders before the user asks. In Figure~\ref{fig:proactive_demo_b}, the harness detects a flight-cancellation notification (highlighted in the first stage) that puts a next-day 14:00 demo at risk. It proactively searches alternative flights and high-speed trains in two execution phases, evaluates which options arrive on time, and presents a decision-ready travel recovery plan, leaving the final booking decision to the user. Both examples show the harness converting passive notifications into contextualized, executable assistance while keeping consequential actions under user control.
\begin{figure}[!t]
    \centering

    \includegraphics[width=\textwidth]{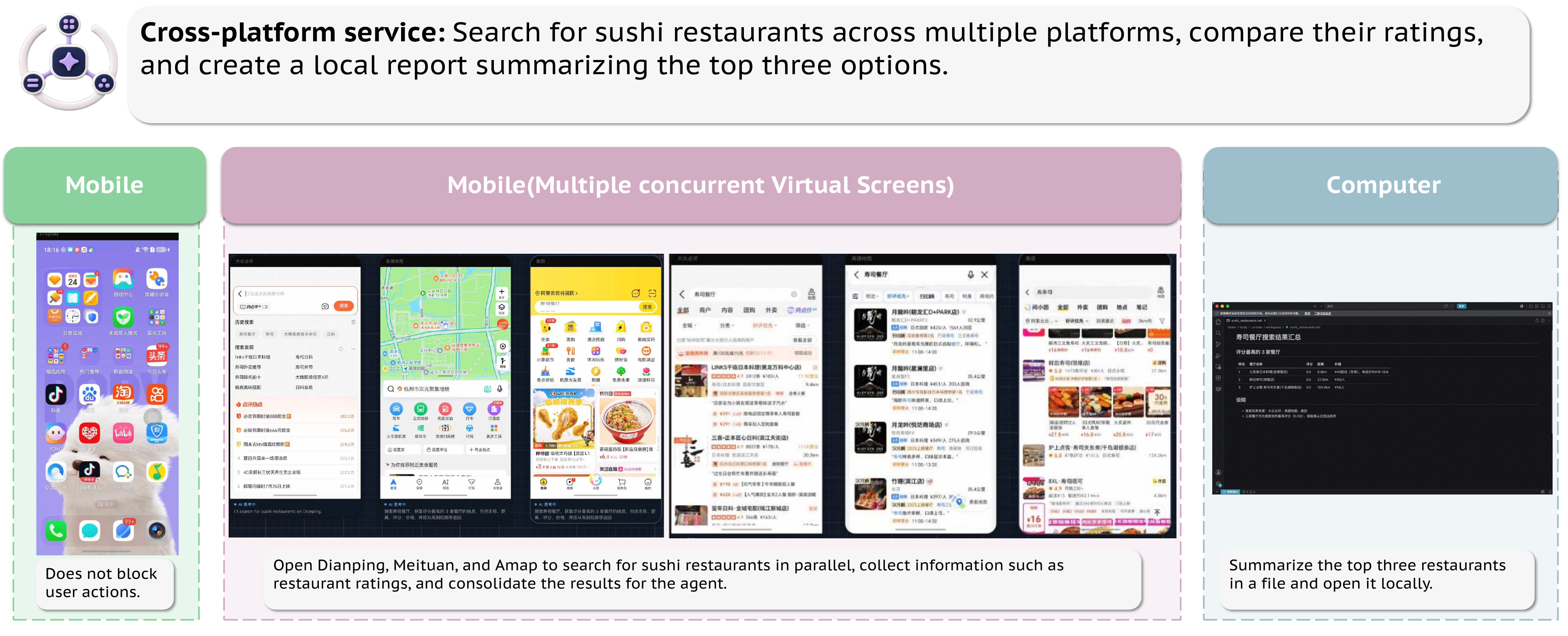}

    \caption{\textbf{Demonstration of cross-platform task execution.} In this workflow, mobile subtasks run on virtual screens of the physical device, so execution does not block the user's own actions. Parallel multi-app search: the agent searches sushi restaurants on Dianping, Meituan, and Amap concurrently through multiple virtual screens, consolidates their ratings, and summarizes the top three options in a local report opened on the computer.}
    \label{fig:crossplatform_demo_a}
\end{figure}
\paragraph{Cross-platform Task Execution.}
Figure~\ref{fig:harness}(II) illustrates a cross-platform workflow for team-building restaurant selection. The harness decomposes the task into four device-addressed stages: \modelname{} discovers candidate restaurants on mobile, organizes their ratings, cuisines, and reservation phones into a spreadsheet on the desktop, sends the resulting artifact to the team leader for approval, and finally saves the approved restaurants back to the mobile map application. Throughout the workflow, the shared task state carries the candidate list, the generated spreadsheet, and the approval outcome across devices, so each stage resumes from the latest context rather than restarting from scratch. 
We present two executed trajectories of this capability in Figure~\ref{fig:crossplatform_demo_a} and Figure~\ref{fig:crossplatform_demo_b} of Appendix~\ref{app:examples}: a receipt-organization workflow that hands mobile gallery selections to GUI+CLI computer use for expense-report generation, and a parallel restaurant search that operates Dianping, Meituan, and Amap concurrently on multiple virtual screens; in both cases, mobile subtasks run on virtual screens without blocking the user's own use of the device. Together, these examples demonstrate that the harness can preserve task state, artifacts, and dependencies as execution moves between mobile and computer environments.

\section{Behavioral Analysis of GUI Agent Execution}
\label{sec:analysis}
Aggregate benchmark scores show whether an agent completes a task, but reveal less about how it executes the task, why it fails, and how its behavior changes after RL training. We therefore examine four questions mainly on the execution behavior of \modelname{}. \textbf{(1)} Why do models optimized for simulated benchmarks struggle on physical devices? \textbf{(2)} How does \modelname{} combine GUI, CLI, and batched actions during task execution, and what distinctive execution patterns emerge from these choices? \textbf{(3)} How does action RL change step-level execution behavior? \textbf{(4)} How does online RL reshape trajectory-level behavior in long-horizon tasks? Together, these analyses complement aggregate metrics by revealing how the agent responds to challenging environments, uses its action space, and changes after RL training.

\subsection{
How Does a Model's Limited Real-Device Experience Affect Its Execution Behavior?
}
\label{subsec:sim2real_gap}

\vspace{-0.5em}
\paragraph{Behavioral Analysis of Real-device Failures.}
Most models have limited direct exposure to physical-device interaction during training. Frontier models such as Qwen 3.7 Plus  perform strongly across a broad range of benchmarks, yet their results on MobileWorld-Real suggest that reliable execution on physical devices remains challenging. This raises a more specific behavioral question: when models have limited real-device interaction during training, which execution patterns emerge, and how do these patterns contribute to task failure? To investigate this question, we review every failed Qwen 3.7 Plus trajectory on MobileWorld-Real and AndroidDaily to identify the primary failure modes.

\begin{table}[t]
    \centering
    \caption{Failure-pattern distribution over all failed Qwen 3.7 Plus trajectories on real devices.}
    \label{tab:failure_patterns}
    \small
    \setlength{\tabcolsep}{6pt}
    \renewcommand{\arraystretch}{1.15}
    \begin{tabular}{@{}llrl@{}}
        \toprule
        \textbf{Dimension} & \textbf{Failure Pattern} & \textbf{Prop.} & \textbf{Typical Behavior} \\
        \midrule
        \multirow{3}{*}{\makecell[l]{Execution Capability \\Limitations (40.3\%)}}
          & Exploration Failure     & 19.5\% & Fail to locate deep in-app entries \\
          & Erroneous Action Loops  & 14.3\% & Repeat ineffective actions \\
          & Lost Execution State    &  6.5\% & Forget finished sub-tasks \\
        \midrule
        \multirow{3}{*}{\makecell[l]{Real-world Scenario \\Challenges (52.0\%)}}
          & UI Misreading           & 24.7\% & Misread stateful page semantics \\
          & Pop-up Interference     & 18.2\% & Ads, paywalls, CAPTCHAs, blank pages \\
          & Physical Widget Control &  9.1\% & Overshoot targets, never converge \\
        \midrule
        Others & --                 &  7.7\% & Under-execution, premature stop \\
        \bottomrule
    \end{tabular}
\end{table}

\begin{figure}[!t]
  \centering
  \includegraphics[width=0.95\textwidth]{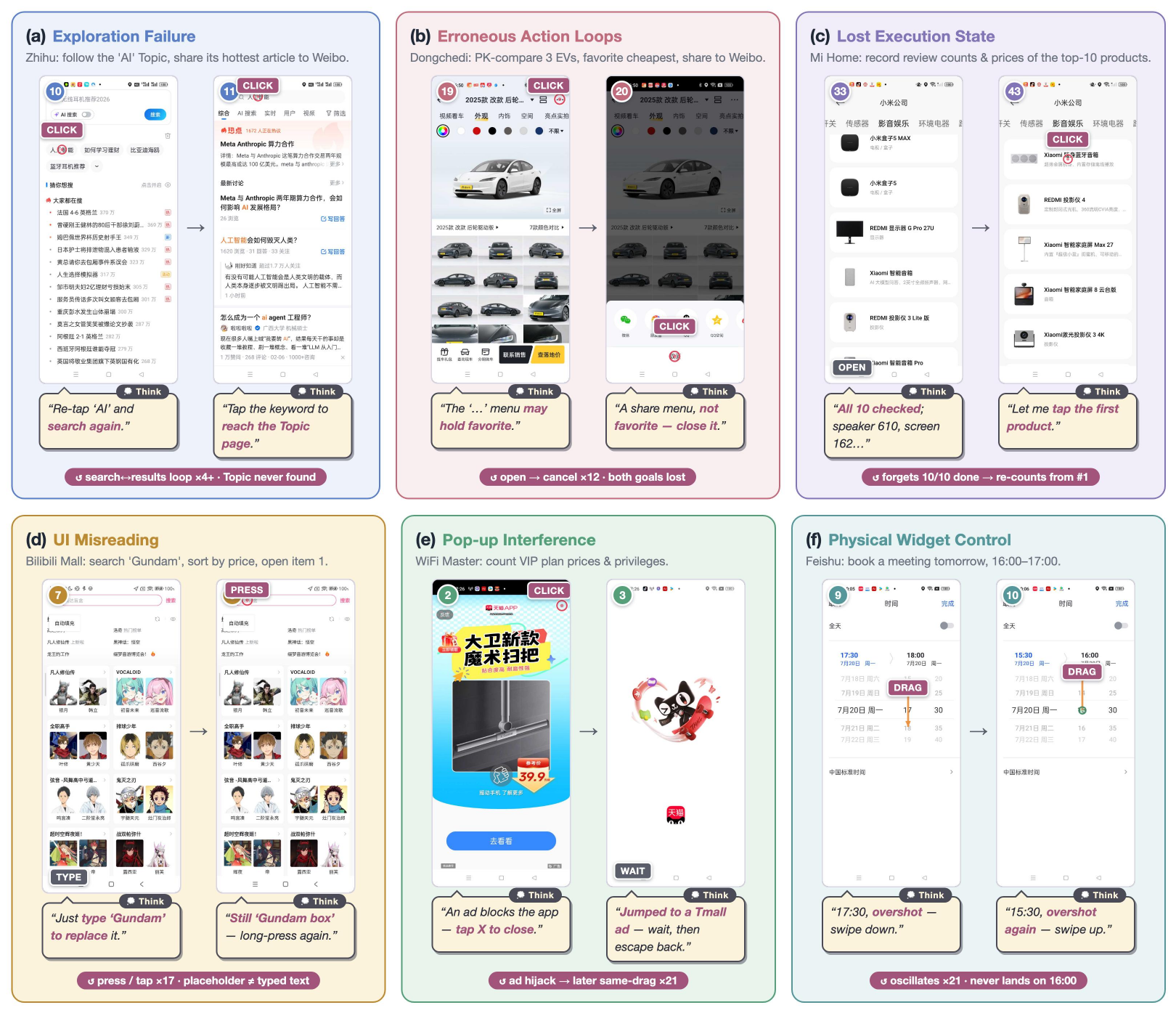}

\caption{Representative real-device failure patterns of Qwen 3.7 Plus. \emph{Execution capability limitations:} (a)~exploration failure, (b)~erroneous action loops, (c)~lost execution state. \emph{Real-world scenario challenges:} (d)~UI misreading, (e)~pop-up interference, (f)~physical widget control. Speech bubbles quote the model's abridged thinking, with the critical fallacy in purple.}
  \label{fig:realdevice_badcase}
\end{figure}
As shown in Table~\ref{tab:failure_patterns}, we organize the failures along two dimensions, both rooted in the scarcity of real interaction experience during training: \emph{execution capability limitations} (40.3\%), where simulated environments provide little incentive to learn exploration, error recovery, and state tracking, and \emph{real-world scenario challenges} (52.0\%), triggered by interface phenomena that sanitized simulators rarely contain. As illustrated in Figure~\ref{fig:realdevice_badcase}, we pair two key frames with the model's abridged thinking for one representative case of each pattern, and analyze each pattern in detail below.

\vspace{-0.5em}
\paragraph{Execution Capability Limitations.} 
Qwen 3.7 Plus' failures in this dimension (40.3\%) reflect limitations in three execution capabilities: locating less commonly used or deeply nested app functions, recovering from unexpected feedback or interface states, and maintaining task state over long trajectories.
\begin{itemize}[leftmargin=*]
\vspace{-0.5em}
    \item  \emph{Exploration failure} (19.5\%) occurs when the target is a less commonly used function hidden behind multiple entry points or deep page hierarchies. In these cases, Qwen 3.7 Plus often revisits a small set of shallow pages instead of exploring alternative routes (Figure~\ref{fig:realdevice_badcase}(a)). By contrast, \modelname{} examines multiple plausible entries before committing to a path.
\item \emph{Erroneous action loops} (14.3\%) stem from the long tail of real third-party applications, where an action frequently fails to produce the expected transition. Handling such expectation violations is itself a learned skill that simulators, with their highly predictable interaction outcomes, offer few opportunities to acquire. When an action fails to produce the intended result, we find that models with limited exposure to physical-device interaction during training tend to repeat the same action or alternate between two interface states rather than diagnose the failure and revise their plan (Figure~\ref{fig:realdevice_badcase}(b)).
\item \emph{Lost execution state} (6.5\%) appears in long-horizon tasks that require intermediate information and completed progress to be retained across app switching, interruptions, and multiple subtasks. In these trajectories, Qwen 3.7 Plus may lose track of prior progress and repeat completed subtasks or restart part of the workflow from scratch (Figure~\ref{fig:realdevice_badcase}(c)).
\end{itemize}
\vspace{-0.5em}
\paragraph{Real-world Scenario Challenges.}
Failures in this dimension (52.0\%) involve interface conditions that are common on physical devices but may be less frequently represented during training: ambiguous page semantics, unexpected interface disruptions, and interactive widgets requiring fine-grained manipulation.
\begin{itemize}[leftmargin=*]
\vspace{-0.5em}
\item \emph{UI misreading} (24.7\%) arises because real pages carry stateful semantics that pure visual appearance does not disambiguate, such as pre-filled search terms, placeholder hints, and dynamically refreshed lists, whereas simulator interfaces stay clean and static. Models trained in simulated environments consequently misjudge the interface state, most typically mistaking a grey placeholder for user-entered text and repeatedly attempting to clear it instead of typing over it (Figure~\ref{fig:realdevice_badcase}(d)).
\item \emph{Pop-up interference} (18.2\%) covers advertisements, paywalls, CAPTCHAs, and unresponsive or blank pages, non-deterministic disruptions that simulators exclude by design to keep evaluation reproducible. Lacking any recovery prior, these models either follow the disruption away from the original task or keep retrying the blocked path (Figure~\ref{fig:realdevice_badcase}(e)).
\item \emph{Physical widget control} (9.1\%) concerns scroll wheels, sliders, and date pickers that demand incremental closed-loop manipulation, where the agent must observe the current value and adjust the swipe magnitude accordingly. In simulated environments such values are typically set by direct text injection or single taps, so the agent applies fixed-magnitude swipes that overshoot the target in both directions and never converge (Figure~\ref{fig:realdevice_badcase}(f)).
\end{itemize}

\vspace{-0.5em}
The findings above motivate the real-device training and evaluation infrastructure described in Section~\ref{sec:real-device-mobile} and help explain \modelname{}'s strong performance on real-world mobile devices.

\subsection{A Closer Look at GUI--CLI Coordination and Batched Execution in \modelname{}}
\label{sec:analysis-cua}
\modelname{} executes computer-use tasks through trajectories that interleave GUI and CLI actions, while selectively emitting multiple actions together in batched form. This hybrid execution process raises two questions: how does the model divide operations between GUI and CLI, and when does it choose to batch multiple actions? We analyze trajectories generated by \modelname{} on OSWorld-Verified and OSWorld-v2 and present the resulting behavioral patterns below.

\vspace{-0.5em}
\begin{table*}[t]
    \centering
    \caption{\textbf{GUI+CLI usage and batched-execution statistics on OSWorld-Verified and OSWorld-v2.} 
    Panel (a) reports CLI and batched-action usage at the action and task levels. 
    Panel (b) reports the composition of batched outputs. 
    Mean batch size counts the number of primitive actions in a batch. 
    Differences are computed as OSWorld-v2 minus OSWorld-Verified. 
    Differences are reported in percentage points (pp).}
    \label{tab:cua-action-statistics}

    \small
    \setlength{\tabcolsep}{6pt} %
    \renewcommand{\arraystretch}{1.15} %

    \begin{tabular}{l l ccc}
        \toprule
        \textbf{Statistic} & \textbf{Level} & \textbf{\makecell{OSWorld-Verified}} & \textbf{\makecell{OSWorld-v2}} & \textbf{\makecell{Difference (pp)}} \\
        \midrule

        \rowcolor{gray!10}
        \multicolumn{5}{l}{\textit{\textbf{(a) Overall CLI and batched-action usage}}} \\
        
        CLI & Action & 40.7\% & 55.1\% & +14.4 \\
        CLI & Task   & 92.0\% & 98.2\% & +6.2  \\
        
        \addlinespace[1.5pt] %
        
        Batched & Action & 39.6\% & 41.6\% & +2.0  \\
        Batched & Task   & 62.1\% & 88.9\% & +26.8 \\

        \midrule

        \rowcolor{gray!10}
        \multicolumn{5}{l}{\textit{\textbf{(b) Composition of batched outputs}}} \\

        \multicolumn{2}{l}{GUI-only batches}       & 75.8\% & 64.7\% & -11.1 \\
        \multicolumn{2}{l}{CLI-only batches}        & 13.1\% & 15.0\% & +1.9  \\
        \multicolumn{2}{l}{Mixed GUI+CLI batches}   & 11.0\% & 20.3\% & +9.3  \\
        
        \addlinespace[1.5pt]
        
        \multicolumn{2}{l}{Mean primitive actions per batch} & 3.1    & 3.1    & 0.0   \\

        \bottomrule
    \end{tabular}
\end{table*}

\begin{figure*}[t]
    \centering
    \includegraphics[width=\textwidth]{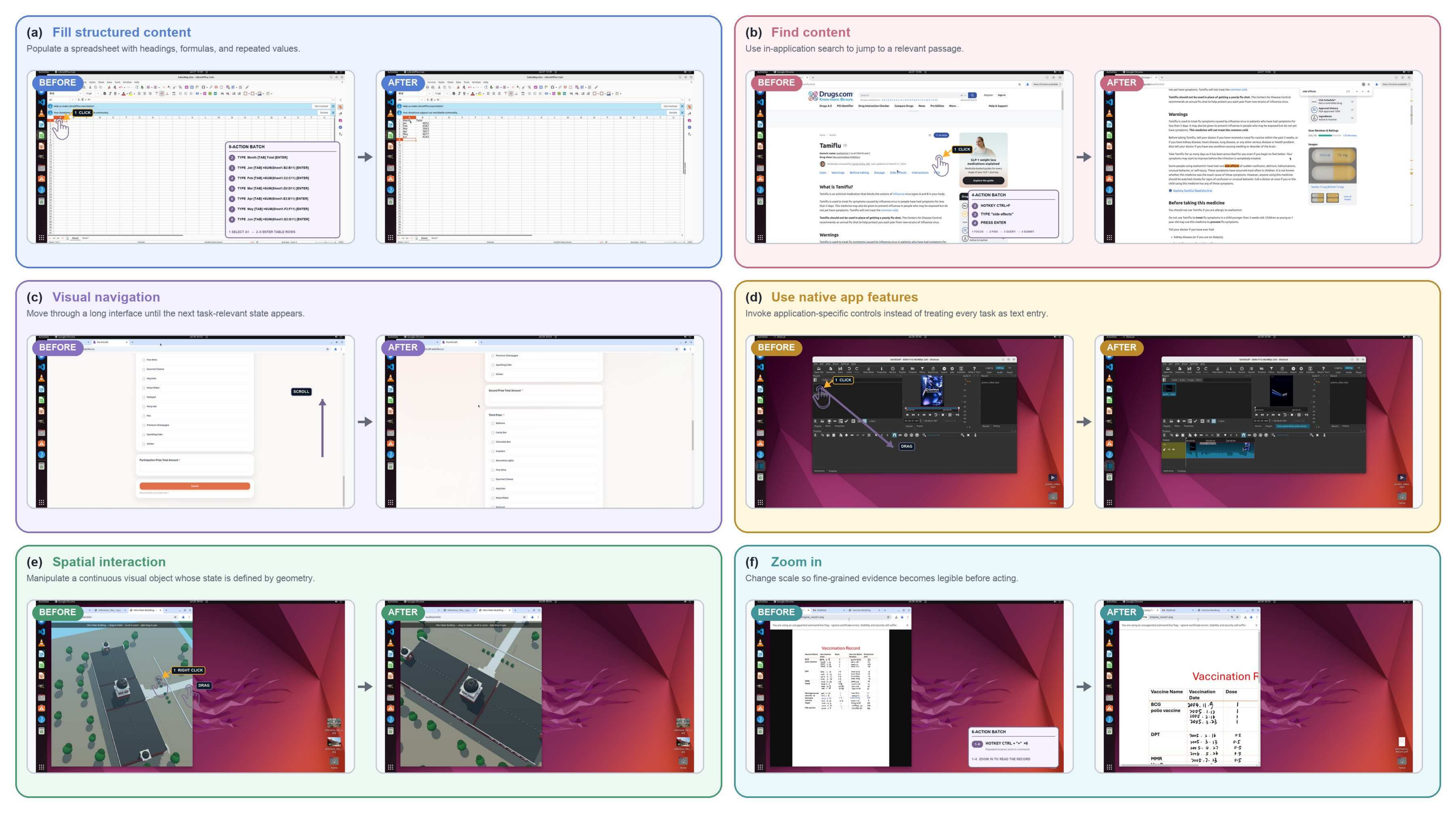}
    \caption{\textbf{Representative GUI interaction patterns.}
    The panels show (a) filling structured spreadsheet content,
    (b) finding content with in-page search, (c) visual navigation through
    scrolling, (d) use of a native media-timeline feature,
    (e) continuous spatial interaction, and (f) zooming in for fine-grained
    inspection. Each panel pairs the rendered action with its resulting
    application state.}
    \label{fig:cua-gui-patterns}
\end{figure*}
\paragraph{Usage Statistics Across Actions and Tasks.}
Table~\ref{tab:cua-action-statistics} quantifies CLI and batched execution at the action and task levels. GUI interaction and CLI operations both serve as primary execution channels in the computer-use trajectories produced by \modelname{}. On OSWorld-Verified, CLI operations constitute 40.7\% of all actions and appear in 92.0\% of tasks. These proportions increase to 55.1\% and 98.2\% on OSWorld-v2, respectively. The near-balanced distribution of GUI and CLI actions suggests that \modelname{} makes flexible and complementary use of both executions. The higher CLI share on OSWorld-v2 partly reflects its task distribution, which contains more workflows that naturally favor programmatic execution, such as merging files, processing structured data, and performing batch file operations.\\
Batched actions account for 39.6\% and 41.6\% of all actions on OSWorld-Verified and OSWorld-v2, and appear in 62.1\% and 88.9\% of tasks, respectively, indicating that batched execution is used extensively across both benchmarks. With an average of 3.1 primitive actions per batch, this corresponds to significantly fewer observation--reasoning--execution cycles than issuing each primitive action in a separate model turn, substantially improving execution efficiency. GUI-only batches remain the dominant form, while CLI-only and mixed GUI+CLI batches show that batched execution also supports programmatic operations and coordination between the two execution channels.

\begin{figure*}[t]
    \centering
    \includegraphics[width=0.95\textwidth]{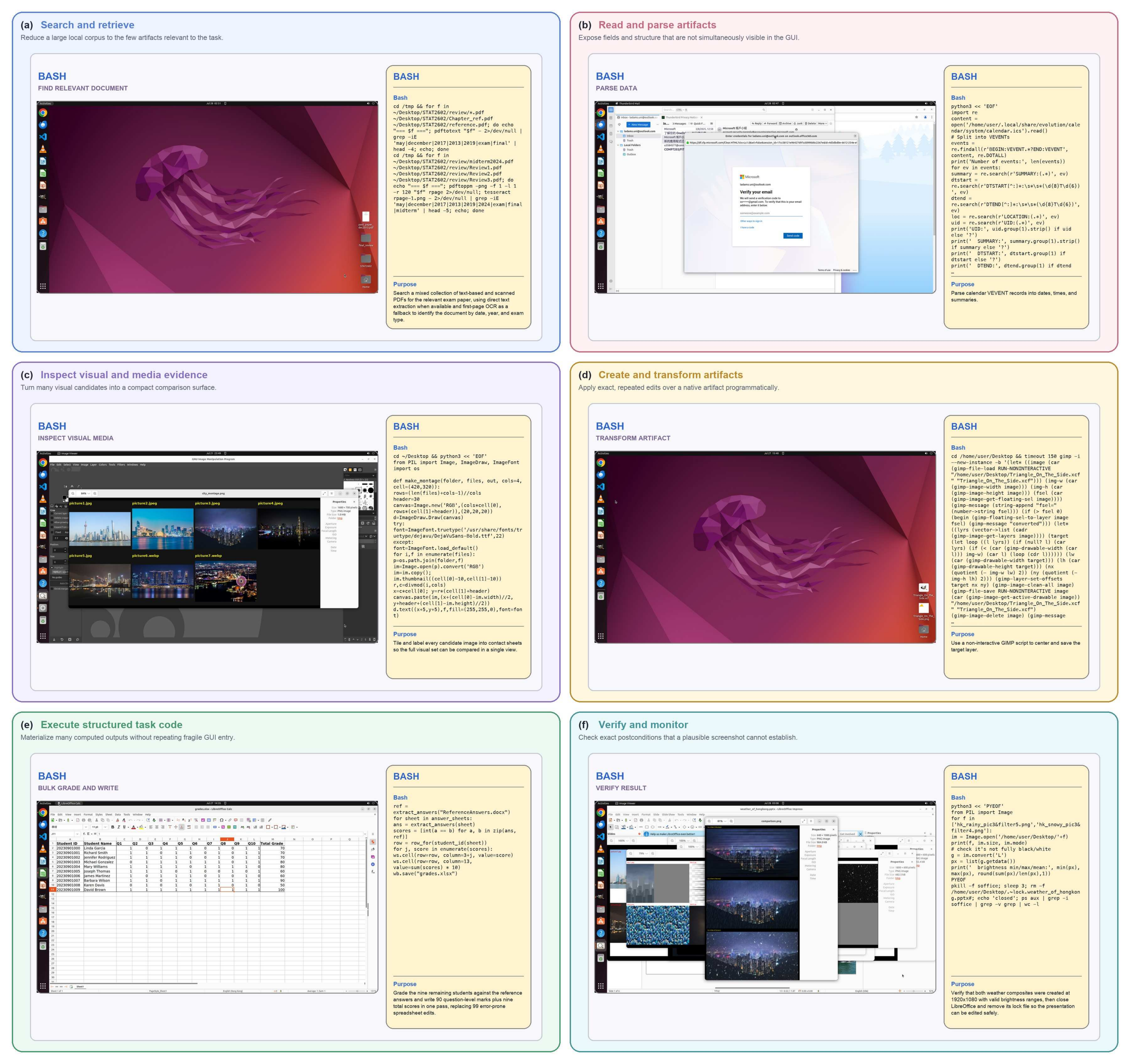}
    \caption{\textbf{Representative CLI interaction patterns.}
    The model uses CLI tools to (a) retrieve relevant documents from a mixed
    corpus, (b) parse machine-readable artifact structure, (c) compress many
    visual candidates into one labeled comparison, (d) transform an artifact
    programmatically, (e) execute a repeated computation and materialize its
    outputs, and (f) verify exact output and process postconditions. Each panel
    juxtaposes the application context with a command excerpt and its
    task-specific purpose.}
    \label{fig:cua-cli-patterns}
\end{figure*}
\paragraph{When Does \modelname{} Use GUI, CLI, and Batched Actions?}
Beyond aggregate usage statistics, we examine recurring patterns in the trajectories to understand how \modelname{} uses GUI interaction, CLI execution, and batched actions in practice. The examples below identify typical scenarios for each capability and illustrate how they complement one another during task execution, although they represent empirical tendencies rather than fixed capability boundaries.

\begin{itemize}[leftmargin=*]
\vspace{-0.5em}
\item \textbf{GUI Interaction.}
\modelname{} tends to rely on GUI interaction when the interface serves as both the control surface and the source of execution feedback, particularly for operations involving native application controls, visual navigation, or spatial manipulation. Figure~\ref{fig:cua-gui-patterns} shows six recurring cases: filling spreadsheet cells with structured headings, formulas, and repeated values, using in-page search to locate a relevant passage, scrolling through a long interface until the required content becomes visible, manipulating media through native timeline controls, interacting with a spatial object whose state is defined by geometry, and zooming in to make fine-grained visual evidence legible before acting. These examples show that GUI interaction supports not only interface navigation but also operations whose execution and subsequent decisions depend on visible state transitions, native application controls, or continuous spatial feedback.
\begin{figure*}[t]
    \centering
    \includegraphics[width=0.91\textwidth]{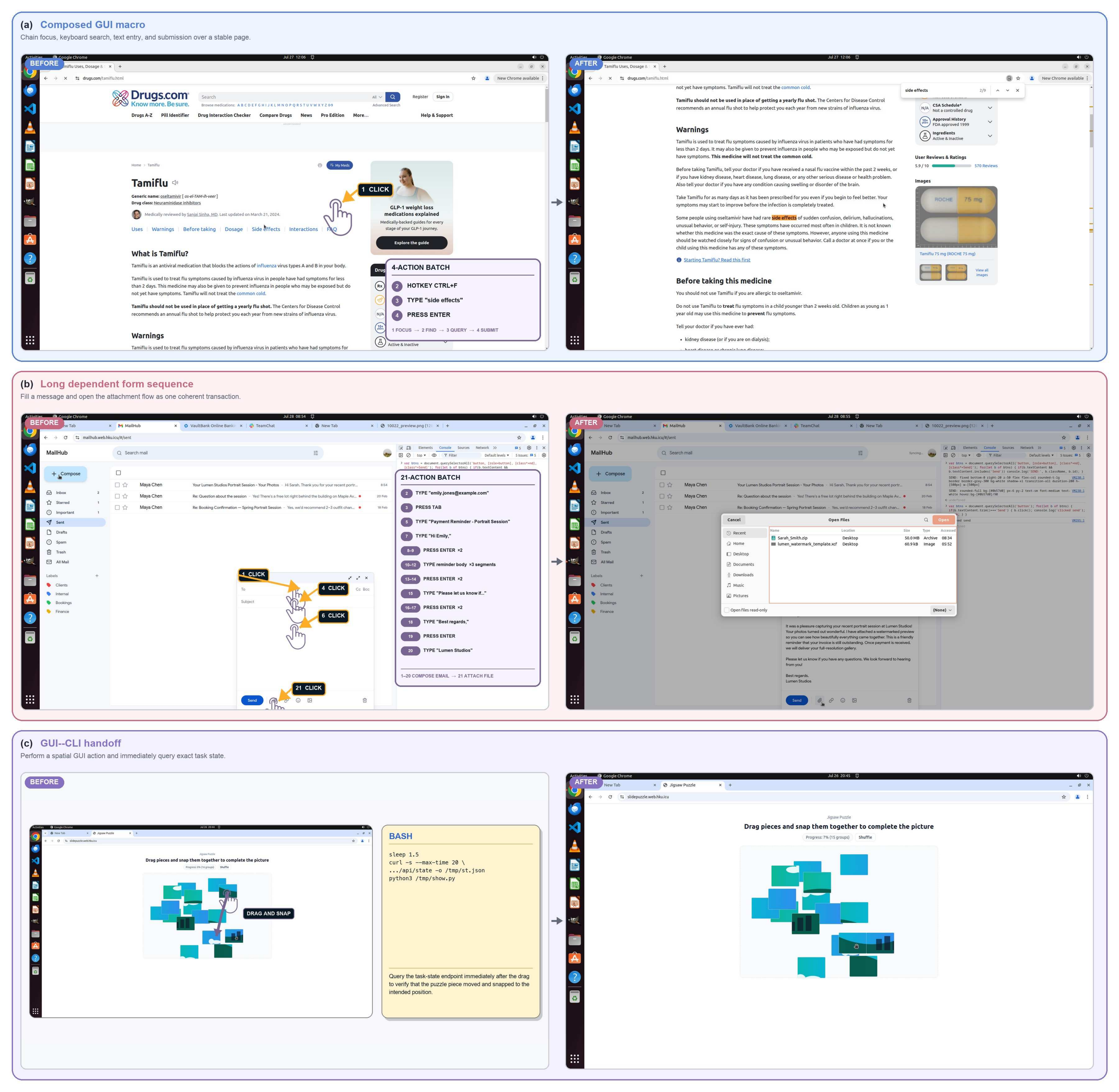}
    \caption{\textbf{Representative batched-action patterns.}
    (a) a compact GUI macro for
    in-page search, (b) a 21-action dependent form sequence that stops at the
    next uncertain dialog, and (c) a GUI--CLI handoff that verifies a spatial
    manipulation through structured task state. Each row pairs the rendered
    batch with the first post-batch observation.}
    \label{fig:cua-batched-patterns}
\end{figure*}
\item \textbf{CLI Execution.}
\modelname{} tends to use CLI when information or operations can be represented more compactly and precisely as data or code than as a sequence of GUI interactions. Figure~\ref{fig:cua-cli-patterns} shows six recurring uses: retrieving relevant documents from a mixed corpus, parsing machine-readable records, organizing multiple images into a labeled contact sheet, transforming artifacts programmatically, executing repeated grading computations, and verifying output properties such as dimensions, brightness, and process state. Together, these patterns show CLI functioning as a search-space reducer before reasoning, an efficient execution channel for structured operations, and a precise postcondition verifier after editing. The contact-sheet example further illustrates that CLI can complement visual reasoning by reorganizing dispersed visual evidence into a single comparison surface rather than replacing the visual judgment itself.
\item \textbf{Batched Execution.}
\modelname{} uses batched execution when multiple operations form a coherent local sequence and their intermediate states are sufficiently predictable to proceed without another observation. As illustrated in Figure~\ref{fig:cua-batched-patterns}, representative cases include a short GUI macro for in-page search, a 21-action dependent sequence that fills an email and proceeds to the attachment picker, and a mixed GUI--CLI sequence that manipulates an object and then verifies the resulting task state. The model terminates the batch when the next decision depends on a new screenshot, command output, dialog, or error. Batched execution therefore groups coherent local transactions rather than merely repeating similar actions.
\end{itemize}

\begin{figure*}[t]
    \centering
    \includegraphics[width=\textwidth]{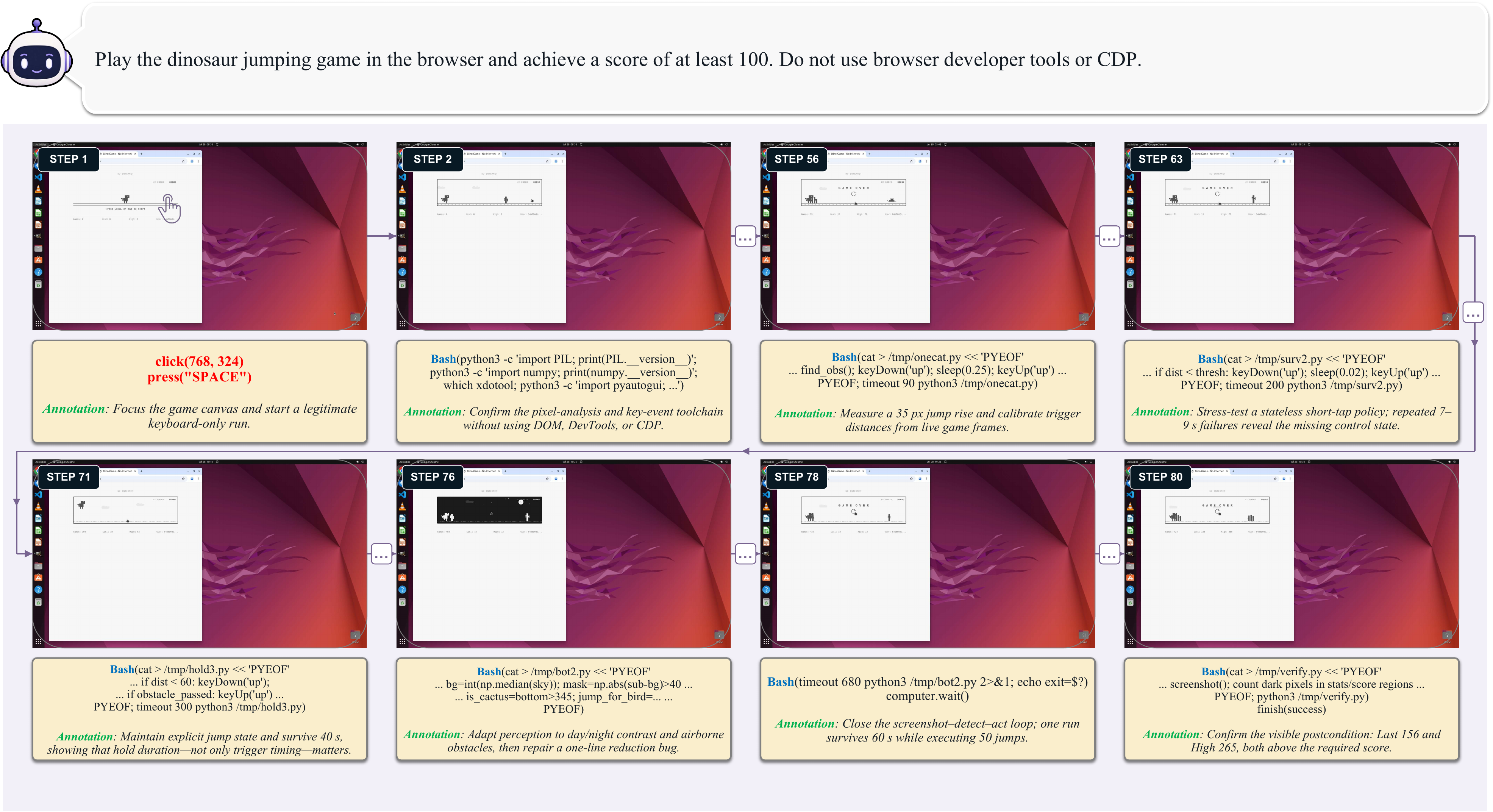}
    \caption{\textbf{Representative trajectory of tightly coupled GUI--CLI collaboration.}
    The task requires \modelname{} to achieve a score of at least 100 in a browser-based dinosaur game without using DevTools or CDP. GUI observations provide evidence for measuring jump dynamics, diagnosing failed control policies, and recognizing changes between day and night modes, while CLI code iteratively implements stateful key control and adaptive obstacle detection. 
    }
    \label{fig:cua-gui-cli-case-study}
\end{figure*}
\vspace{-0.5em}
\paragraph{Case study: Tightly coupled GUI--CLI collaboration in a dynamic task.}
One particularly interesting trajectory asks \modelname{} to play a browser-based dinosaur game and achieve a score of at least 100 without using browser developer tools or the Chrome DevTools Protocol (CDP). Issuing every jump as a separate model action is poorly suited to this task because obstacles continue moving during screenshot acquisition and model inference. \modelname{} instead uses CLI to construct and iteratively refine a local controller that operates through rendered screenshots and standard keyboard events. The controller does not access the DOM or internal game state, leaving the visible GUI as its source of task state and the keyboard as its action interface.

The trajectory reveals a tightly coupled pattern of GUI--CLI composition. Successive GUI observations provide the evidence needed to measure jump dynamics, calibrate obstacle-trigger distances, and diagnose errors in jump timing and duration. CLI code converts these observations into explicit perception and control procedures. When repeated trials expose the limitations of stateless, fixed-duration taps, the model introduces explicit action state and holds the jump key until the obstacle has passed. A further challenge arises when the game switches from day to night mode. The resulting contrast inversion invalidates the initial assumption that obstacles correspond to dark pixels. After identifying this change from the rendered frames, \modelname{} replaces fixed color thresholding with background-relative segmentation and uses obstacle geometry to handle both ground-level cacti and airborne birds.

The resulting controller survives for 60 seconds while executing 50 jumps and reaches a visible last-run score of 156. This case shows that GUI and CLI provide different but interdependent functions within a single trajectory. The GUI supplies perceptual evidence and execution feedback, while CLI execution turns that evidence into repeatable perception and control procedures that do not require model inference at every game step. By repeatedly moving from GUI observation to CLI-based revision and back to GUI verification, \modelname{} constructs an effective controller without privileged access to the application.

\begin{table}[!t]
    \centering
    \caption{Performance on five error-pattern-specific test sets before and after action RL training.}
    \label{tab:error_pattern_specific}
    \small
    \begin{adjustbox}{max width=\textwidth}
    \begin{tabular}{lccccc}
        \toprule
        \multirow{2}{*}{\textbf{Model}}
& \textbf{Confusable-Element}
& \textbf{Sorting and}
& \textbf{Multi-Target}
& \textbf{Premature}
& \textbf{Repetitive Action} \\
& \textbf{Grounding}
& \textbf{Ranking}
& \textbf{Completeness}
& \textbf{Completion}
& \textbf{Loops} \\
        \midrule
        W/o Action RL & $72.8\%$ & $76.6\%$ & $80.0\%$ & $81.0\%$ & $72.9\%$ \\
        W/ Action RL  & $79.1\%$ & $80.4\%$ & $84.4\%$ & $86.2\%$ & $82.4\%$ \\
        \bottomrule
    \end{tabular}
    \end{adjustbox}
\end{table}
\begin{figure}[!t]
    \centering
    \includegraphics[width=\linewidth]{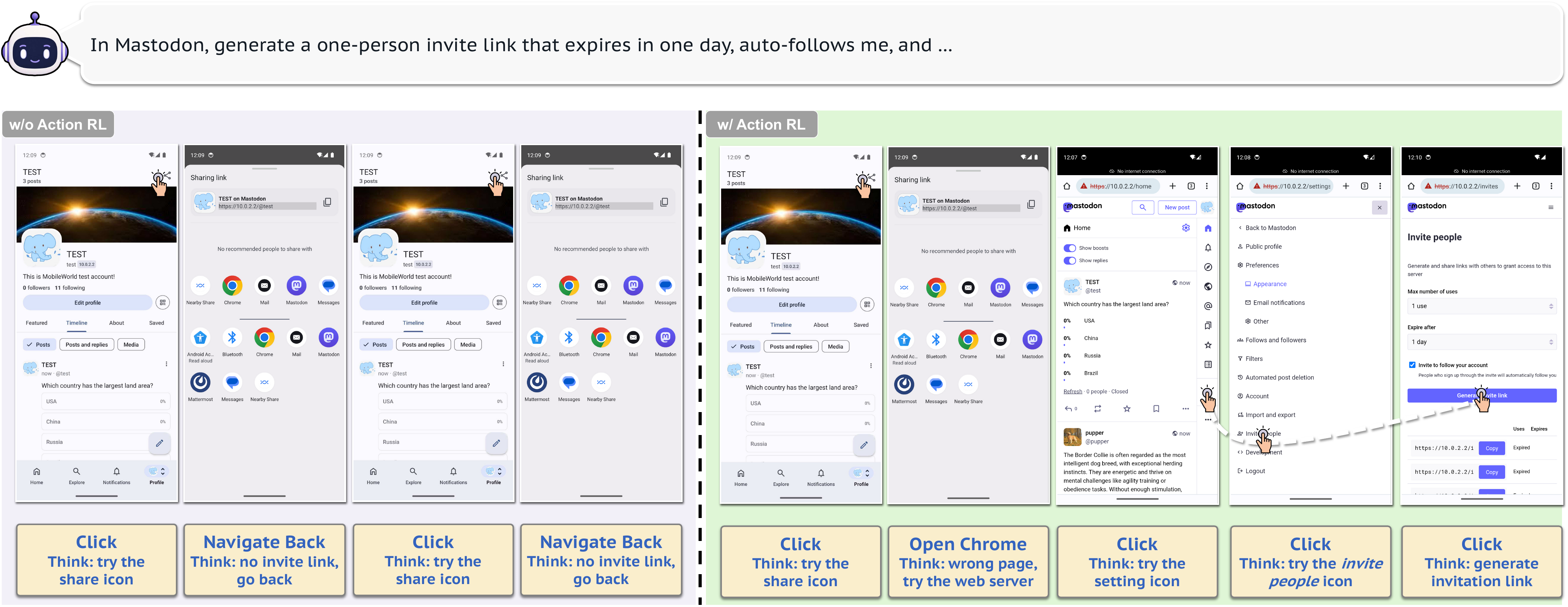}
    \caption{\textbf{Action RL case study.} 
    The SFT model navigates to the account page and repeatedly clicks the \textit{share} icon before navigating back, becoming trapped in an ineffective local loop. In contrast, the model after action RL recognizes that it is on the wrong page, switches to exploring Mastodon's web interface, locates the correct entry point, and successfully configures the target invite link with the required settings.}
    \label{fig:actionRL_case_study}
\end{figure}

\subsection{Step-Level Behavioral Corrections after Action RL}

\textbf{Overall Performance and Behavioral Shift.} We further conduct a comparative analysis of the model behaviors before and after action RL training. The primary effect of action RL is not merely an improvement in aggregate task success, but a systematic correction of erroneous action decisions at a fine-grained, step-level granularity. Beyond improving the overall task SR by more than 7\%, action RL reduces the number of reasoning tokens by 21.3\% while increasing the average number of interaction steps by 8.4\%. These changes suggest that the updated policy performs less redundant reasoning and more frequently translates its decisions into environment-grounded actions. More importantly, the performance gains are accompanied by a clear shift in action patterns, characterized by the systematic correction of recurring action errors, which we examine in detail below.

\textbf{Correction of Recurring Error Patterns.}
As shown in Table~\ref{tab:error_pattern_specific}, action RL consistently improves performance across all five error-pattern-specific test sets. On grounding tasks that require selecting the correct target among visually or semantically similar candidates, action RL improves SR by 6.3\%. It also improves SR on Sorting and Ranking, Quantity and Multi-Target Completeness, and Premature Completion tasks by 3.8\%, 4.4\%, and 5.2\%, respectively. The effect of action RL is particularly evident in correcting short-range action loops. On tasks involving repeated operations over multiple targets or navigation through long scrollable interfaces, action RL improves SR by 9.5\%. This gain indicates a stronger ability to recognize when recent actions have failed to produce task-relevant state changes and to switch to an alternative action. Figure~\ref{fig:actionRL_case_study} presents a representative example in which the agent is asked to generate an invitation link in Mastodon. The model before action RL repeatedly interacts with the profile-sharing icon and eventually reaches the step limit. In contrast, the model after action RL recognizes that it is on the wrong page, explores an alternative path through Mastodon's web interface, and reaches the correct interface for configuring the invitation link. This example highlights the central role of step-level action RL: when a short sequence of actions fails to make meaningful progress, the automatically constructed loop-breaking examples teach the policy to revise its immediate action choice rather than persist with an ineffective pattern.

\begin{table}[t]
    \centering
    \caption{Performance comparison between frequent and long-tail actions.}
    \label{tab:long_tail_actions}
    \small
    \begin{tabular}{lccc}
        \toprule
        \textbf{Action Group}
        & \textbf{Original  Data Proportion}
        & \textbf{Reward before Action RL}
        & \textbf{Reward after Action RL} \\
        \midrule
        Frequent actions  & $80.1\%$ & $88.3\%$ & $92.3\%$ \\
        Long-tail actions & $19.9\%$ & $71.5\%$ & $77.9\%$ \\
        \bottomrule
    \end{tabular}
\end{table}
\begin{figure}[!t]
    \centering
    \includegraphics[width=1.00\linewidth]{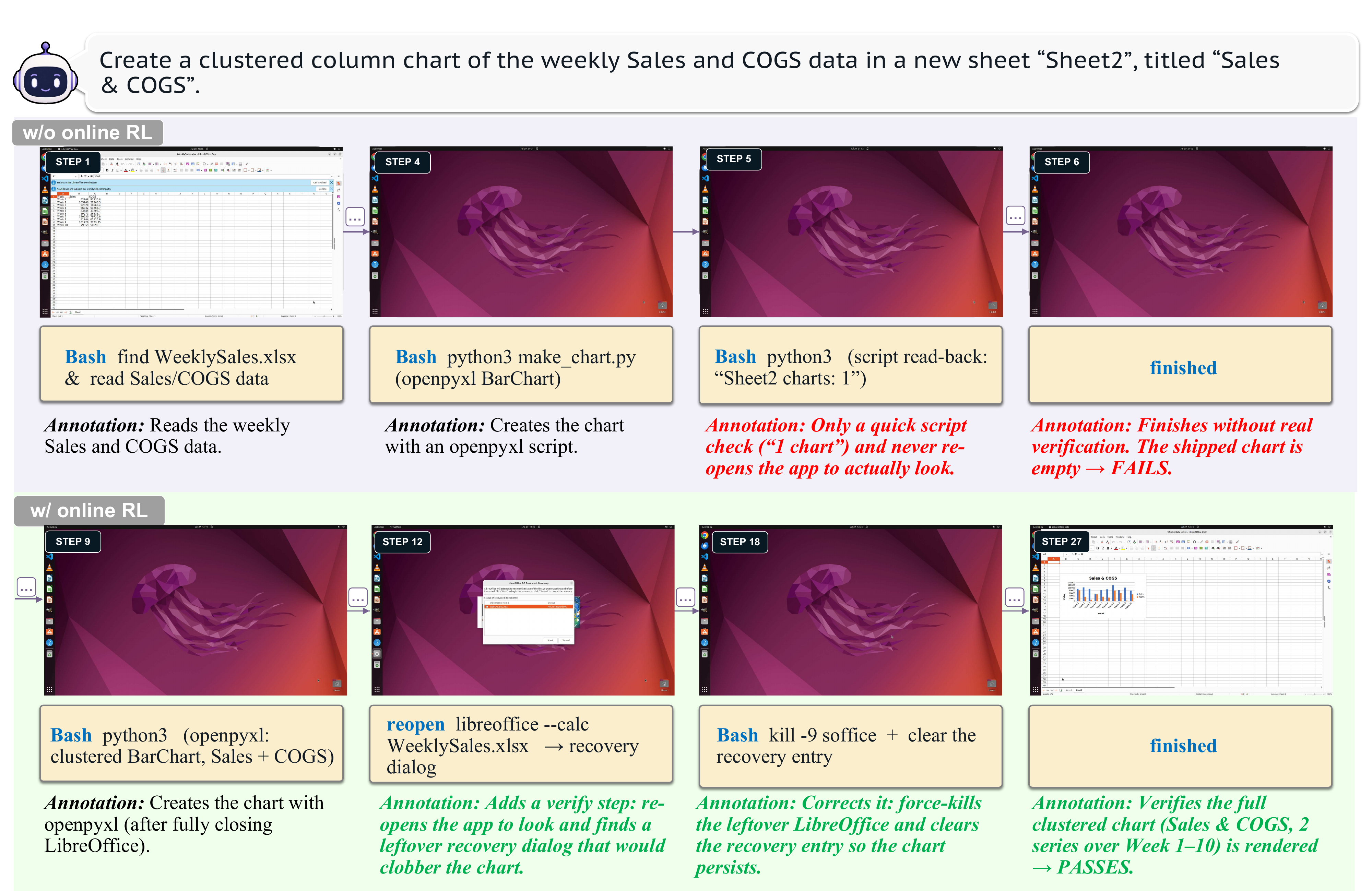}
    \caption{\textbf{Online RL elicits verification and self-correction.} On the same task, the online RL policy (bottom) succeeds where the SFT policy (top) fails, because it reopens the application to inspect the actual result and corrects the problem it finds before finishing. The SFT policy trusts a superficial script check that only counts the chart, never looks at the rendered output, and ships an empty chart with no data series, whereas the RL policy reopens the file to verify, finds that a leftover application instance would overwrite the chart, removes it so the chart persists, and confirms that the full clustered chart is actually rendered before terminating.}
    \label{fig:verification}
\end{figure}
\textbf{Improved Handling of Long-tail Actions.}
GUI action trajectories exhibit a highly imbalanced distribution. Frequent actions such as \texttt{click}, \texttt{drag}, and \texttt{type} account for 80.1\% of primitive actions, whereas long-tail actions such as \texttt{ask\_user}, and \texttt{long\_press} account for only 19.9\%. Despite their low frequency, these actions are often critical for reaching specific application states or invoking specialized controls. Pass-$N$ rollouts show that their reward is 16.8\% lower than that of frequent actions, reflecting their limited coverage in the training data. Action RL therefore increases the proportion of long-tail action examples to about 40\% in the training data, improving their reward by 6.4\%. These results demonstrate that action RL can selectively strengthen low-frequency but high-impact action decisions.

\begin{figure}[!t]
    \centering
    \includegraphics[width=1.00\linewidth]{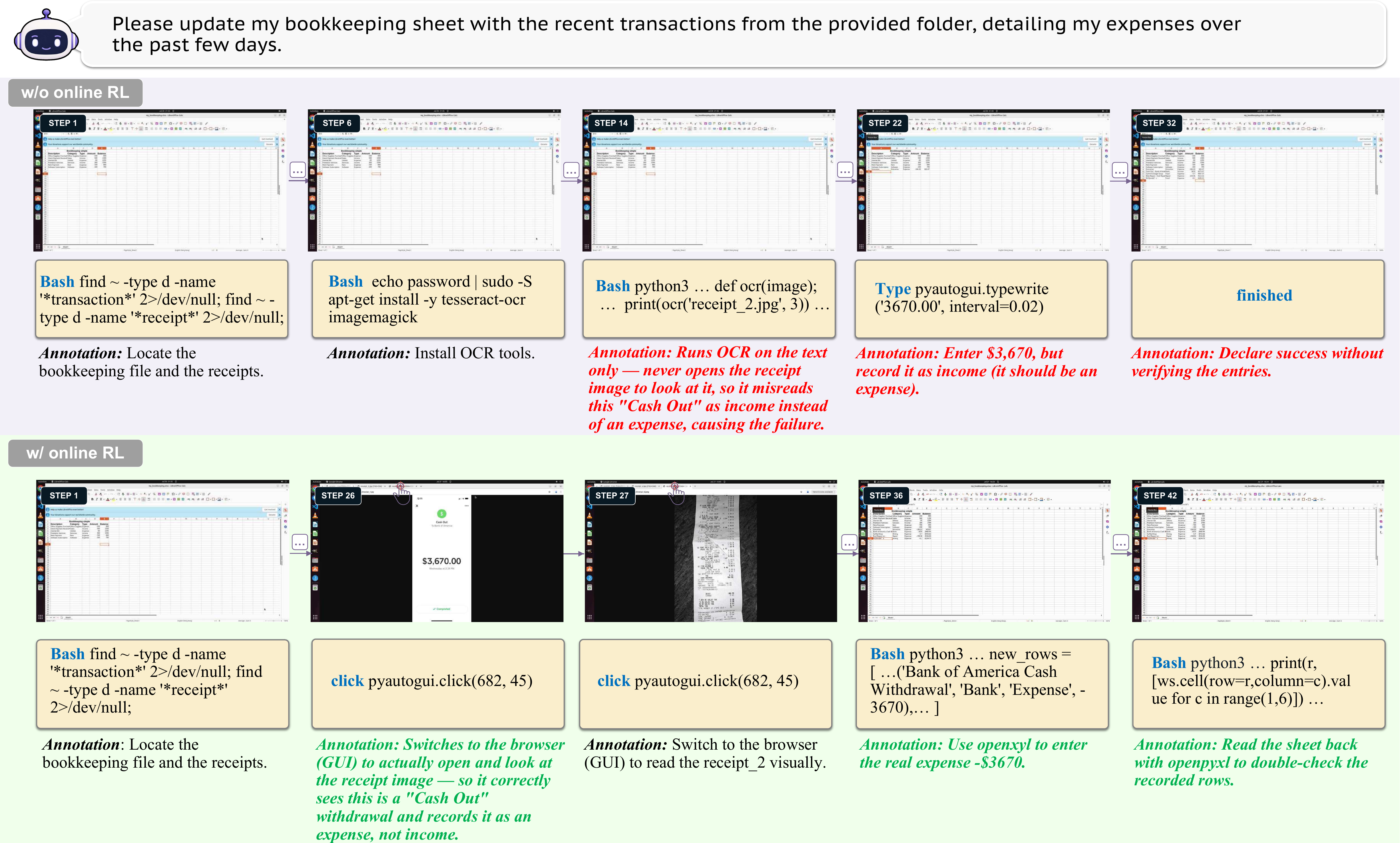}
    \caption{\textbf{Online RL elicits emergent cross-modal action.} On the same task, the online RL policy (bottom) succeeds where the SFT policy (top) fails, because it grounds its decisions in what it visually observes and verifies the result before finishing. The SFT policy reads the receipt from OCR text alone, mistakes a Cash Out for income, and declares success without checking, whereas the RL policy opens the receipt image to visually read it, records it correctly as an expense, and reads the spreadsheet back to verify before terminating.}
    \label{fig:modality-specialization}
\end{figure}
\begin{figure*}[t]
    \centering
    \includegraphics[width=\textwidth]{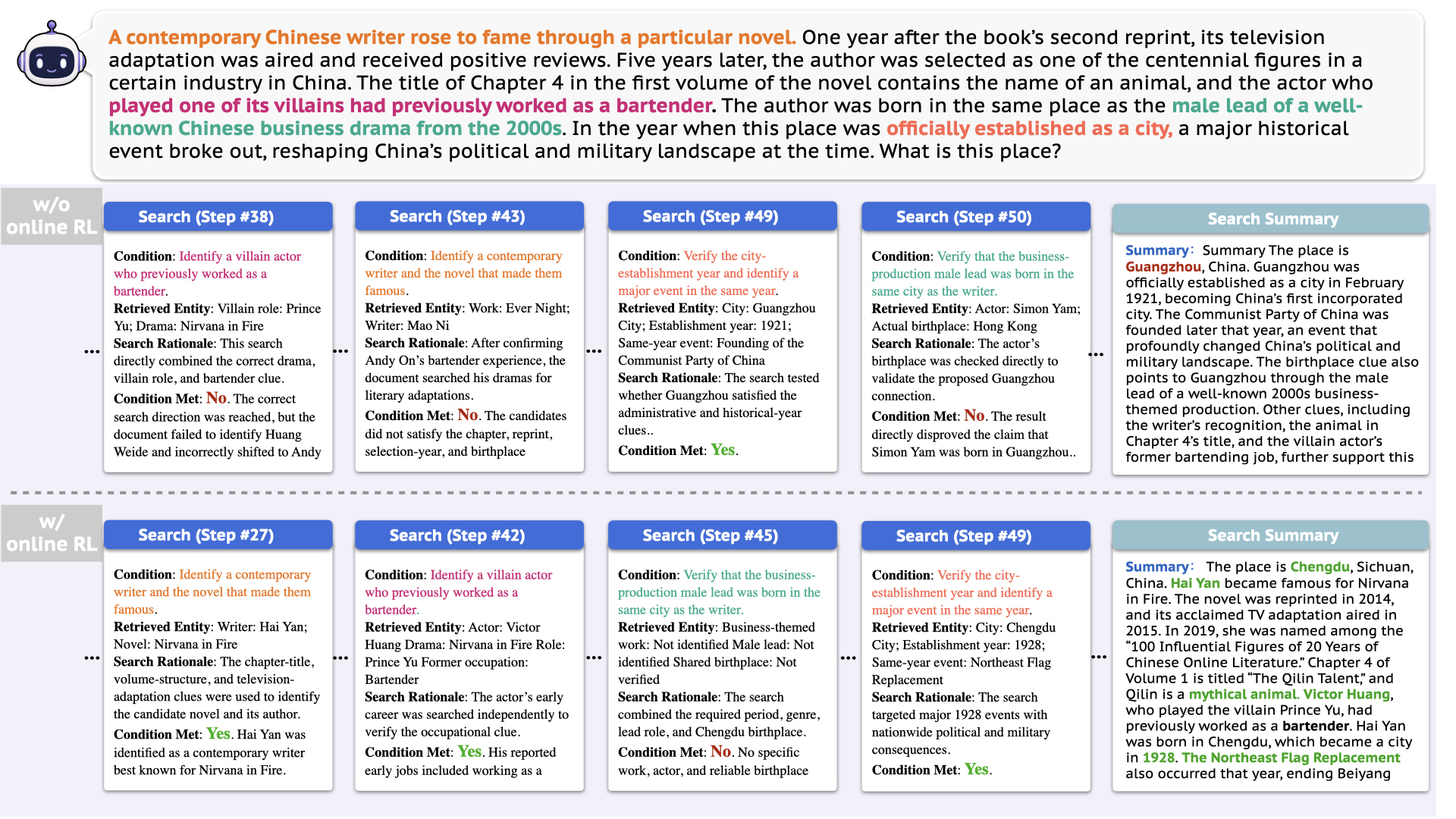}
    \caption{\textbf{Online RL improves long-horizon search and verification.}
Without online RL, the model follows incorrect entity associations and reaches the unsupported answer Guangzhou. With online RL, it identifies the key chain of Hai Yan, \textit{Nirvana in Fire}, Huang Weide, and Chengdu.
}
    \label{fig:constraint-adherence}
\end{figure*}

\subsection{Trajectory-Level Behavioral Shifts after Online RL}
\label{sec:analysis-online}
By carefully comparing the trajectories generated before and after online RL across different domains, we identify the emergence of three desirable behavioral patterns.

\paragraph{From Assumed Success to Verified Completion.} Before online RL, the policy often terminates immediately after its final
state-changing action, assuming that the task has been completed without
verifying the intended effect. After online RL, the policy more frequently
inspects the resulting state before termination and recovers when verification
reveals incomplete execution. Figure~\ref{fig:verification} illustrates a
representative example. Before online RL, the policy performs a sequence of
Bash operations and declares completion without checking their effects. After
online RL, the policy initially follows the same Bash-based path but adds a
verification step, discovers that part of the intended output was not
generated, switches to an alternative approach, and ultimately completes the
task successfully. Aggregate results on OSWorld show the same trend: the
fraction of trajectories containing at least one verification action increases
by 14.7\%, accompanied by an 11.2\% decrease in the false-stop rate---the
fraction of self-declared successful episodes that fail the final verifier.

\paragraph{Bash as Hands, GUI as Eyes: Emergent Cross-modal Collaboration.} After online RL, the policy shifts from a Bash-leaning strategy to a more
balanced use of Bash and GUI actions: the share of GUI actions
increases by 6\%, while the fraction of trajectories employing both GUI and
Bash increases by 10.6 \%. Notably, despite the absence of an
explicit objective for modality coordination, a useful pattern of cross-modal
collaboration emerges: the policy uses Bash to execute state changes
efficiently and the GUI to inspect their effects. Consistent with this pattern,
the fraction of trajectories exhibiting an execution--verification
transition---a state-changing Bash action followed by a read-only GUI
inspection---increases from 40.2\% to 52.4\%.
Figure~\ref{fig:modality-specialization} illustrates a representative example.
Before online RL, the policy relies only on Bash to execute the task and,
without visual information, fails to distinguish income from expenses,
introducing errors into the result. After online RL, the policy executes
efficiently with Bash and then opens individual receipts in the GUI for visual
inspection, allowing it to distinguish the relevant entries and complete the
task correctly. This cross-modal loop combines efficient execution with
reliable feedback, with Bash serving as the hands and the GUI as the eyes.

\paragraph{Maintaining Constraints Over Long Horizons.}
Long-horizon tasks often impose additional constraints on the primary goal,
including quantities, formats, ranges, and exclusions. Comparing behavior
before and after online RL, we find that the policy becomes more reliable at
preserving constraints introduced earlier in the instruction throughout
execution. Across the two benchmarks, the fraction of tasks satisfying all
instruction constraints increases by 8.6\% on OSWorld and by 7.5\% on
BrowseComp-ZH. Figure~\ref{fig:constraint-adherence} illustrates a
representative BrowseComp-ZH example involving four constraints. In this case,
the policy shifts from satisfying only one constraint and reaching an incorrect
conclusion to retaining and jointly applying all four, leading to the correct
answer.

\section{Related Work}
Recent work on GUI agents across \emph{mobile-use}, \emph{browser-use}, and \emph{computer-use} increasingly favors native multimodal policies over prompt-only orchestration, with stronger use of self-evolving data engines, online reinforcement learning, and structured memory~\citep{ui-tars-15,maiui,uivenus15,webvoyager,opencua,evocua15}. The field is also moving from benchmark-only optimization toward realistic deployment concerns in terms of real-device execution, privacy, operational reliability, safety, and latency \citep{cao2026xiaomi,uitars,maiui,evocua15}. 

\textbf{Mobile Use.} Recent mobile agent works increasingly move beyond screenshot-grounded action prediction toward more capable systems with memory, reflection, RL, and agentic harness. Mobile-Agent-v3.5 \citep{Mobile-agent-v3.5} and UI-Venus-1.5 \citep{uivenus15} scale unified GUI models across diverse platforms and tasks. Step-GUI \citep{step-gui} and UI-TARS-2 \citep{uitars2} investigate data flywheels and multi-turn RL for improving agent capability through iterative interaction. MAI-UI \citep{maiui} explores hybrid GUI-MCP operations and device--cloud collaboration. Xiaomi-GUI-0 \citep{cao2026xiaomi} studies real-device closed-loop training that captures state dynamics. Existing mobile agent evaluations mainly focus on three aspects: grounding capability (e.g., ScreenSpot-Pro \citep{li2025screenspotpro}), interactive task completion in controlled environments (e.g., AndroidWorld \citep{android_world}), and more realistic daily-use scenarios such as MobileWorld \citep{kong2026mobileworld}, AndroidDaily \citep{androiddaily}, and RealMobile \citep{cao2026xiaomi}.

\textbf{Browser Use.} Compared with mobile agents, browser-use agents can additionally leverage structured information such as DOM trees, accessibility trees, and browser APIs. Early research such as WebVoyager \citep{webvoyager} and SeeAct \citep{seeact} demonstrate the capability of multimodal LLMs to perform end-to-end web navigation by combining visual understanding with browser actions. Recent projects, such as OpenAI Operator \citep{openai_operator}, Google Project Mariner \citep{google_mariner}, and UI-TARS \citep{ui-tars-15}, further extend web agents toward autonomous interaction with complex websites and diverse digital environments. Some recent studies like WebWorld \citep{xiao2026webworld} and WebEvolver \citep{fang2025webevolver} also explore world-model-based approaches for web-agent. Web-agent evaluation has evolved from short-horizon webpage navigation toward realistic long-horizon workflows involving multiple websites and complex objectives \citep{zhou2023webarena,xue2025illusion,jang2026odysseys}.

\textbf{Computer Use.}
Computer-use agents (CUA) aim to enable general interaction with desktop environments, covering diverse software and productivity applications. Compared with mobile and web agents, CUAs face a significantly larger action space, more diverse interaction states, and longer horizon, making reliable grounding and execution verification critical. Recent efforts including UI-TARS \citep{uitars}, DART-GUI \citep{li2025dart}, OpenCUA \citep{opencua}, UltraCUA \citep{yang2025ultracua}, EvoCUA \citep{evocua15} investigate scalable GUI foundation models by leveraging multimodal pretraining, large-scale interaction data, and agentic reasoning. Recent CUA benchmarks increasingly emphasize robustness, scalability, and real-world deployment challenges. OSWorld-Verified improves evaluation reliability through stricter verification protocols and reduced ambiguity in success assessment \citep{OSWorld}. OSWorld-v2 further scales evaluation toward more complex, long-horizon workflows and diverse real-world computer tasks \citep{osworld2}.

\section{Conclusion}

We presented \modelname{}, a real-world centric foundation GUI agent. Guided by a vision of agents that operate on real devices, maintain workflows across platforms, combine GUI interaction with CLI execution, and proactively initiate useful services, \modelname{} integrates scalable sandbox environments with a robust real-device mobile runtime, a unified GUI+CLI action space, an agent-driven data flywheel, a training framework combining SFT, action RL, and online RL, and a harness layer for proactive service initiation and cross-platform task execution. 
Empirically, \modelname{} achieves \textbf{82.1\%} on MobileWorld, \textbf{92.2\%} on MobileWorld-Real, and \textbf{97.5\%} on AndroidDaily. It also reaches {\textbf{79.5\%}} on OSWorld-Verified, and {\textbf{40.0\%}} on OSWorld-v2. \modelname{} further obtains leading or competitive performance across browser use, DeepSearch, and GUI-grounding, while preserving strong general reasoning and agentic capabilities. Together, these results support a systems view of GUI agent development where models, environments, data flywheel, training, and harness are co-designed around real-world usage.

\section{Limitations and Future Directions}
\paragraph{Limitations.}
First, we evaluate real-device mobile trajectories using AutoJudge rather than deterministic verifiers or manual expert examination. Deterministic state-based verification is generally infeasible on physical phones because of the restrict access to the internal states of third-party applications. Meanwhile, manually adjudicating trajectories from \modelname{} and ten baselines would require substantial expert effort and introduce inter-annotator variation. To ensure a consistent comparison, we apply the same AutoJudge protocol to all systems. On 666 examples independently examined by experts, AutoJudge achieves 92.8\% exact-match accuracy. Nevertheless, its remaining errors may introduce minor uncertainty into the reported real-device results. Second, CUA and DeepSearch training at the 35B-A3B scale was still in progress when this technical report was released, so the corresponding results are not included. We plan to update the arXiv version once these results become available. Third, beyond the real-device environments described in this report, higher-fidelity synthetic environments that more closely reproduce real applications provide another path to narrowing the simulation-to-real gap. We have constructed such environments, but training with them was not yet incorporated into the reported model, so this component is not presented in the current report. We plan to open-source the environment synthesis methodology in future work. Fourth, our attempts to fully automate GUI capability development showed that current foundation models cannot yet manage the entire process reliably. Our pipeline is therefore agent-driven rather than fully autonomous and still requires considerable human oversight and intervention to monitor progress and correct errors.

\paragraph{Future Directions.}
We discuss several directions that we believe warrant further investigation as GUI agents move toward reliable and practical real-world use.
\begin{itemize}[leftmargin=*]
\vspace{-0.5em}
\item \textbf{Efficient GUI Execution.}
GUI agents are capable to complete increasingly complex workflows directly on real devices, marking an encouraging step toward practical use. However, capability alone is not sufficient. Execution efficiency, particularly interaction latency, remains a major obstacle to practical use. Each GUI step typically requires a new observation, model inference, and environment transition, causing latency to accumulate over long trajectories. Virtual-display mechanisms across mobile and desktop systems can partially mitigate this problem by enabling background or parallel execution, but they do not remove the fundamental per-step cost of the perception--reasoning--action loop. Reducing this cost through faster inference, adaptive observation, asynchronous execution, and fewer interaction rounds therefore remains an important for bringing GUI agents into practical real-world use.

\item \textbf{Harness-Assisted Long-Horizon Workflows.}
Our hybrid GUI+CLI action space and support for batched actions improve execution efficiency, while online RL strengthens decision making over long trajectories. A complementary direction is to further exploit the harness as a source of long-horizon support beyond the model itself. Harness-level designs for context compression, task decomposition, memory management, progress tracking, and structured tool use could improve the efficiency and reliability of long-horizon workflows from another perspective.

\item \textbf{Large-Scale Cross-Domain Online RL.}
Long-horizon GUI tasks make online RL costly because rollouts are slow and variable in length, while rewards are often sparse or delayed. Future work should develop more efficient and stable training strategies for these settings, while jointly training across browser-use, mobile-use, and computer-use environments to improve data efficiency and cross-domain generalization~\citep{uitars2,hou2026single}.

\item \textbf{Scalable Synthesis of High-Fidelity Environments.}
Real-device environments provide the most faithful interaction dynamics, but remain expensive to scale, reset, and verify. High-fidelity synthetic environments offer a complementary path by approximating real-world application behavior in a form that can be generated and controlled at scale. Scalable synthesis of such environments is important for narrowing the simulation-to-real gap and expanding real-world-oriented training. We have constructed such environments, unfortunately, their integration into model training was not completed in time for this report. We will open-source the environment synthesis methodology in future work.
\item \textbf{Safety, User Control, and Personalization.}
As GUI agents gain the ability to operate across applications and devices, ensuring that their actions remain safe and under user control becomes increasingly important. In this work, we include safety-sensitive training scenarios that teach the model to invoke the \texttt{call\_user} action and return control to the user when additional input or authorization is required. More systematic safety evaluations, safety-oriented training objectives, and interpretability-based methods for understanding and constraining agent behavior remain important directions. Personalization is equally important for improving the real-world user experience. Our proactive-service harness provides an initial exploration, but constructing useful user memories and profiles, grounding them in reliable information sources, and giving users control over what information is retained all warrant further investigation.
\end{itemize}

\section{Contributions}\label{contribution}

\subsection*{Core Contributors}
Hanzhang Zhou$^{*}$, Panrong Tong$^{*}$, Xu Zhang$^{*}$, Quyu Kong, Chenglin Cai, Tianyu Xia, Gongjie Zhang, Jianan Zhang, Long Li, Long Chen, Lei Wang, Gaole Dai, Pengxiang Li, Liangyu Chen, Yue Wang$^{\dagger}$, Steven Hoi. 

\subsection*{Contributors}
Hongliang Lu, Miaoyi Zhou, Chen Liu, Yuchen Sun, Yucheng Zhao, Yiran Zhong, Jiahui Zeng, Minggang Wu, Zhixiang Ma, Shaoshun Huang, Shuaihao Zhang.

\begingroup
\renewcommand\thefootnote{}
\footnotetext{$^*$ Project Co-Leader. $^\dagger$ Project Leader.}
\endgroup

\clearpage
\bibliography{biblio}
\bibliographystyle{colm2024_conference}

\clearpage
\appendix
\section{Appendix}
\label{app}

\subsection{Validating AutoJudge for Real-Device Evaluation}
\label{app:autojudge-validation}

Unlike sandbox benchmarks, real-device tasks generally cannot rely on
deterministic, state-based verifiers. Third-party Apps do not expose their
internal states, while task outcomes may depend on changing online content,
account status, or interactions across multiple Apps. Manually reviewing every
trajectory is also difficult to scale and can introduce variation between
annotators. We therefore validate AutoJudge against independent human
annotations before using it for large-scale real-device evaluation.

\begin{table}[h]
    \centering
    \caption{\textbf{Agreement between AutoJudge and independent human annotations.}
    Rows denote AutoJudge decisions and columns denote human annotations.
    \texttt{pass} corresponds to the combined \texttt{perfect}/\texttt{fair\_enough}
    category. Agreement is computed after excluding human annotations marked
    as unclear.}
    \label{tab:autojudge-validation}
    \small
    \setlength{\tabcolsep}{5.5pt}
    \renewcommand{\arraystretch}{1.12}
    \begin{adjustbox}{max width=\columnwidth}
    \begin{tabular}{lccccc}
        \toprule
        & \multicolumn{4}{c}{\textbf{Human annotation}}
        & \\
        \cmidrule(lr){2-5}
        \textbf{AutoJudge decision}
        & \textbf{Pass}
        & \makecell{\textbf{Model}\\\textbf{failure}}
        & \makecell{\textbf{Environment}\\\textbf{error}}
        & \textbf{Unclear}
        & \makecell{\textbf{Class}\\\textbf{agreement}} \\
        \midrule
        \texttt{pass}
        & \textbf{455}
        & 17
        & 1
        & 1
        & 96.2\% \\

        \texttt{failed}
        & 15
        & \textbf{144}
        & 9
        & 1
        & 85.7\% \\

        \texttt{env\_error}
        & 1
        & 5
        & \textbf{19}
        & 0
        & 76.0\% \\

        \midrule
        \textbf{Overall}
        & \multicolumn{4}{c}{618 agreements among 666 conclusive cases}
        & \textbf{92.8\%} \\
        \bottomrule
    \end{tabular}
    \end{adjustbox}
\end{table}

Table~\ref{tab:autojudge-validation} reports the comparison. AutoJudge agrees
with the human annotations on 618 of the 666 trajectories with a conclusive
human label, corresponding to an overall agreement of 92.8\%. Agreement reaches
96.2\% for successful trajectories. Model failures and environment errors are
more difficult to distinguish, since similar observations can result from
either an incorrect Agent action or a problem with the live execution
environment.

We further ask expert reviewers to examine the cases in which AutoJudge and
the initial human annotation disagree. Expert adjudication more often supports
the AutoJudge decision, indicating that the agreement reported above is a
conservative estimate of its accuracy. Based on this validation, we use
AutoJudge as the primary evaluator for our real-device experiments. It provides
a more accurate and scalable evaluation protocol while separately identifying
model failures and errors caused by the live environment.

\subsection{AutoJudge Decisions on Real-Device GUI Agent Trajectories}
\label{app:autojudge-trajectories}

\begin{figure}[h]
    \centering
    \includegraphics[width=0.85\linewidth]{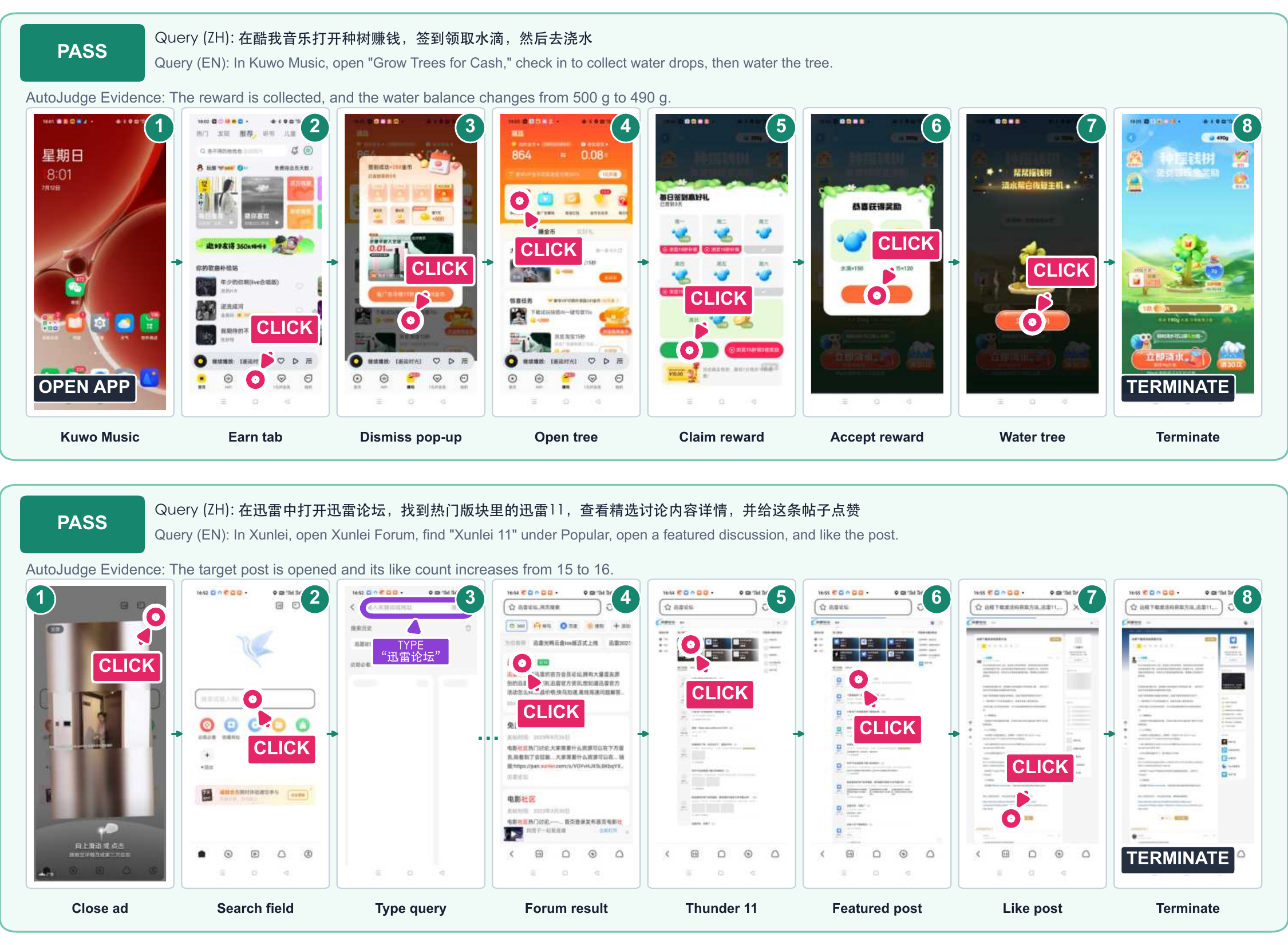}
    \vspace{-2.5mm}
    \caption{\textbf{Representative AutoJudge pass decisions.}
    The trajectory evidence shows that the Agent successfully completes the
    requested task.}
    \label{fig:autojudge_pass}
\end{figure}

\begin{figure}[h]
    \centering
    \includegraphics[width=0.85\linewidth]{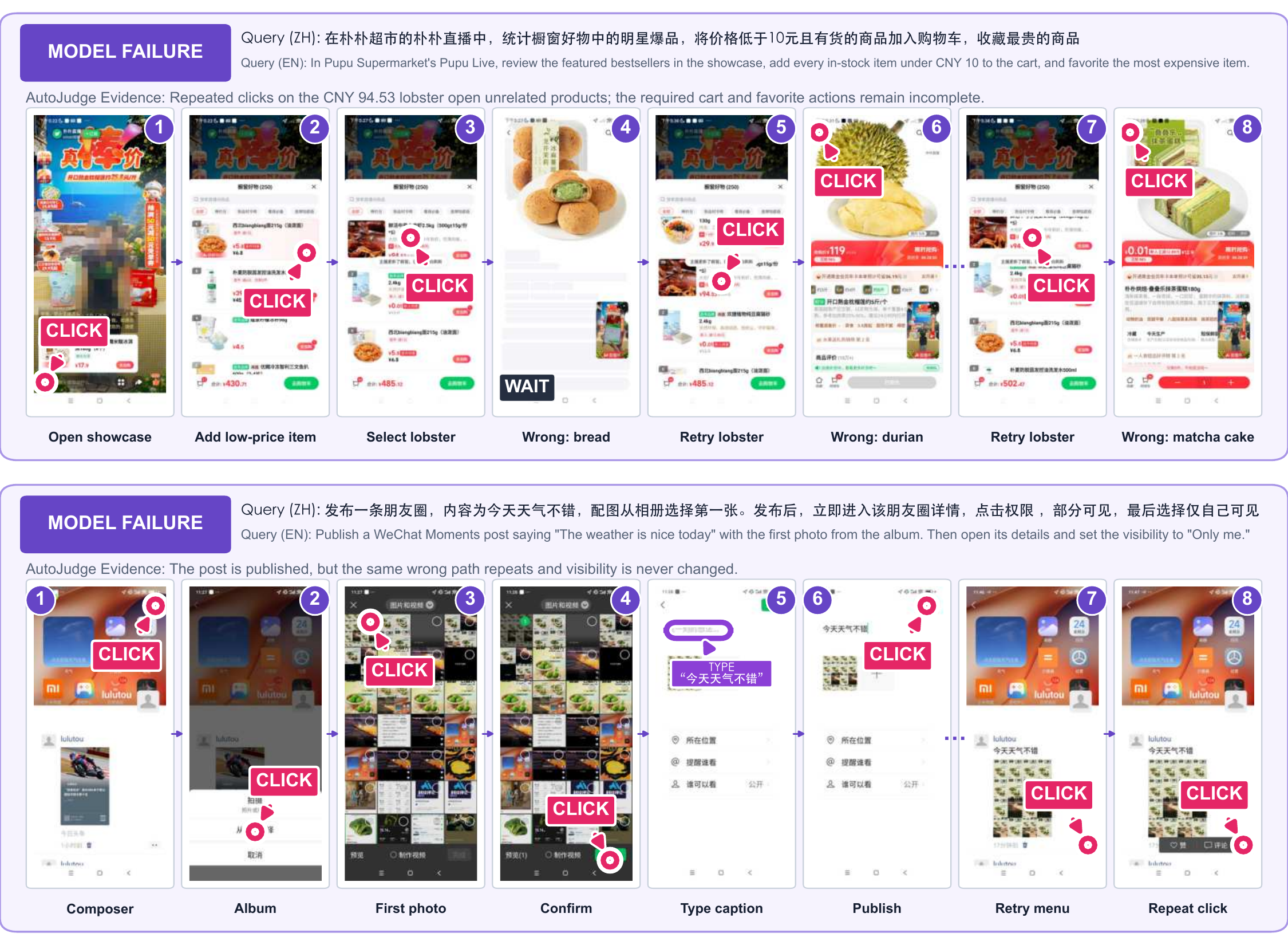}
    \vspace{-2.5mm}
    \caption{\textbf{Representative AutoJudge model-failure decisions.}
    The Agent follows an incorrect or incomplete execution path and does not
    satisfy the task requirements.}
    \label{fig:autojudge_model_failure}
\end{figure}

\begin{figure}[h]
    \centering
    \includegraphics[width=0.85\linewidth]{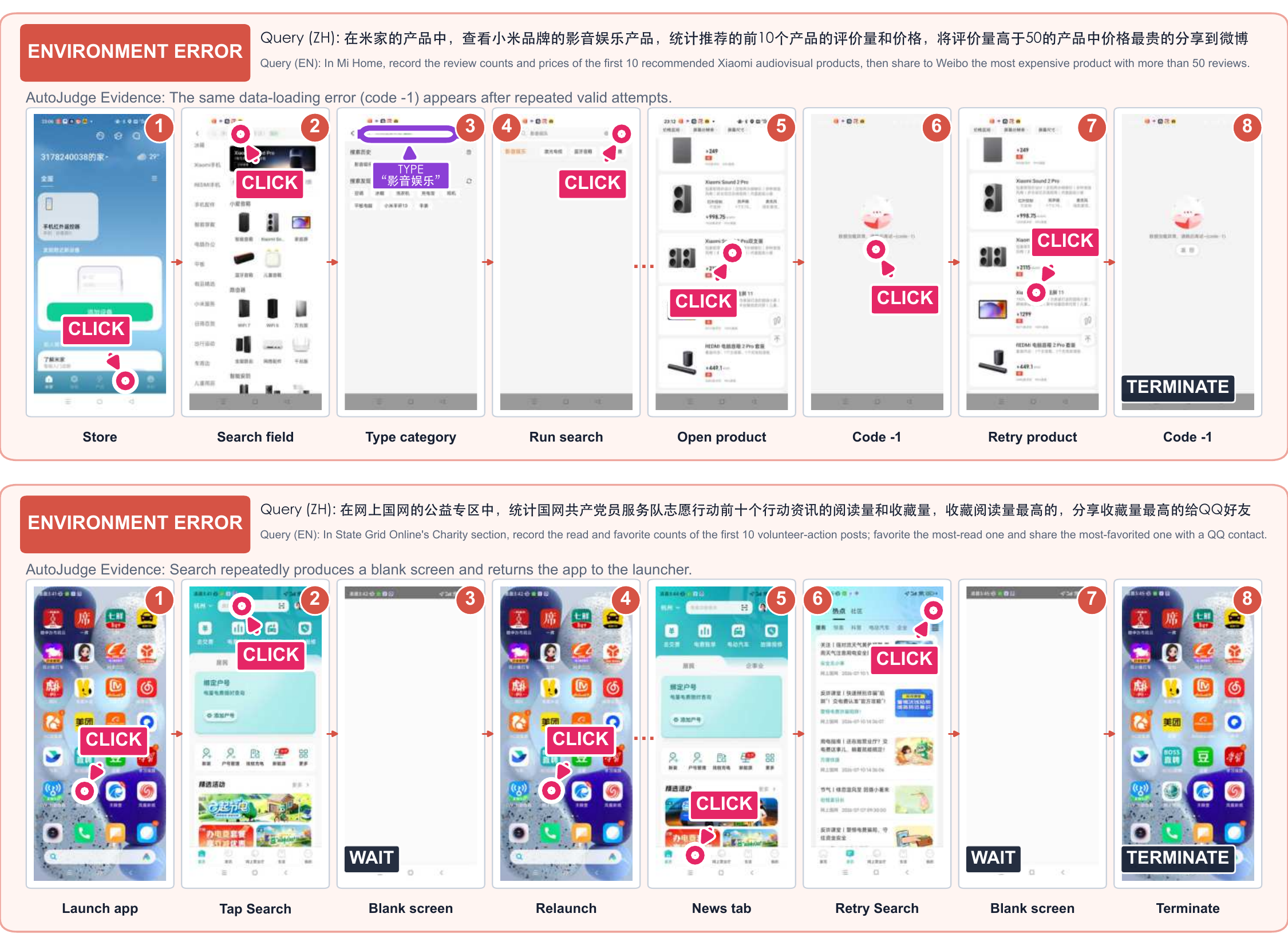}
    \vspace{-2.5mm}
    \caption{\textbf{Representative AutoJudge environment-error decisions.}
    The trajectories are interrupted by live-App or execution-environment
    issues rather than by the Agent's actions.}
    \label{fig:autojudge_environment_error}
\end{figure}

Figures~\ref{fig:autojudge_pass}--\ref{fig:autojudge_environment_error}
show representative examples of the three AutoJudge outcomes. Each example
includes the original Chinese task and its English translation, selected
trajectory steps with actions rendered on the screenshots, and the evidence
used for the final decision. These examples illustrate how AutoJudge
distinguishes successful execution and model failures from errors caused by
the live execution environment.

\newpage
\subsection{Additional Qualitative Examples}
\label{app:examples}

\begin{figure*}[t]
    \centering
    \includegraphics[width=0.91\linewidth]{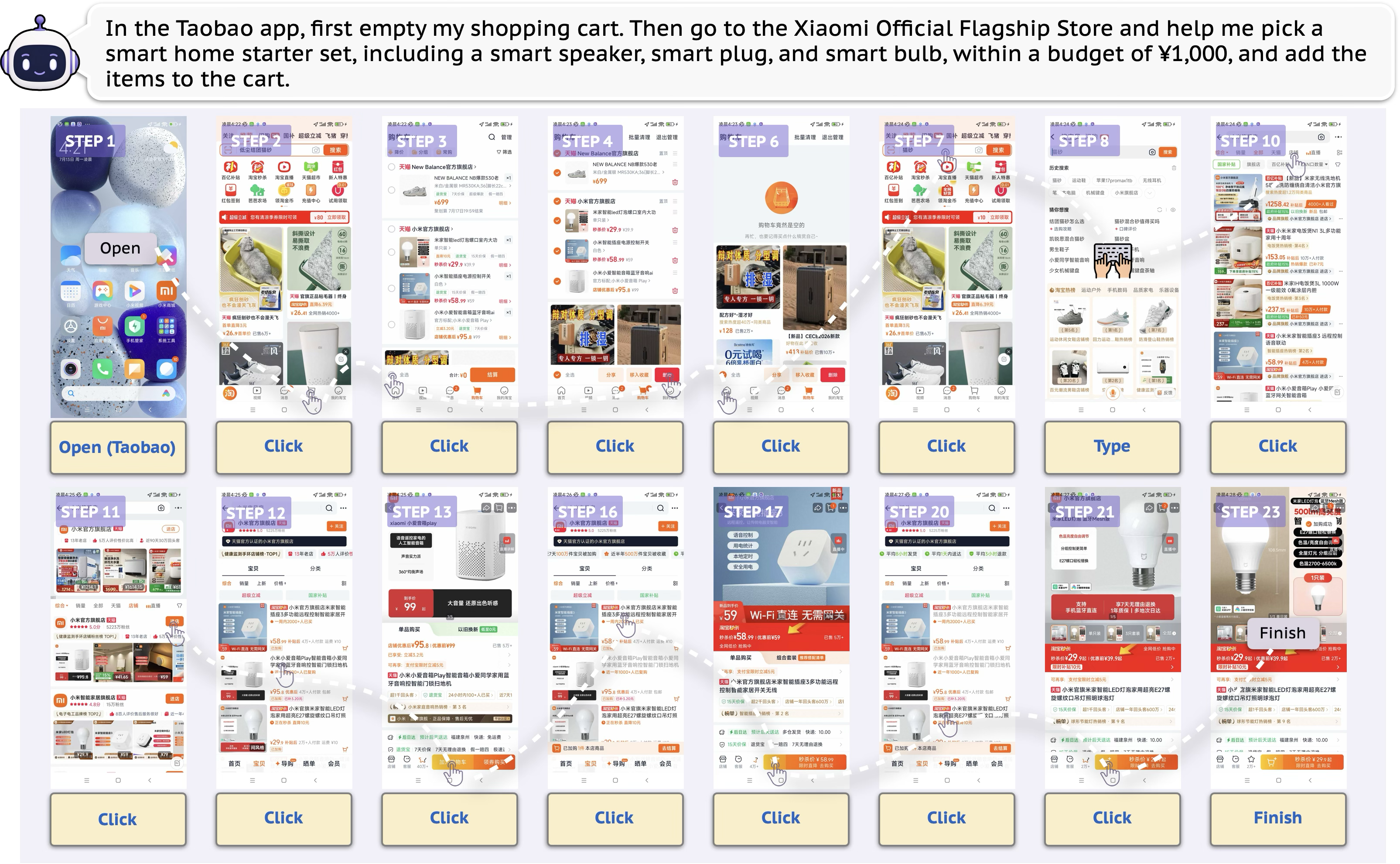}
    \caption{\textbf{Demonstration of long-horizon real-device GUI execution.} The trajectory is rendered as key frames with the executed action annotated beneath each frame. In this 23-step shopping task, the agent decomposes the instruction into sequential sub-goals: it first empties the Taobao shopping cart, then selects a smart speaker, a smart plug, and a smart bulb from the Xiaomi flagship store within a 1{,}000-yuan budget, maintaining the budget constraint and the item checklist throughout.}
    \label{fig:real_device_demo_taobao}
\end{figure*}

\begin{figure}[!t]
    \centering

    \includegraphics[width=\linewidth]{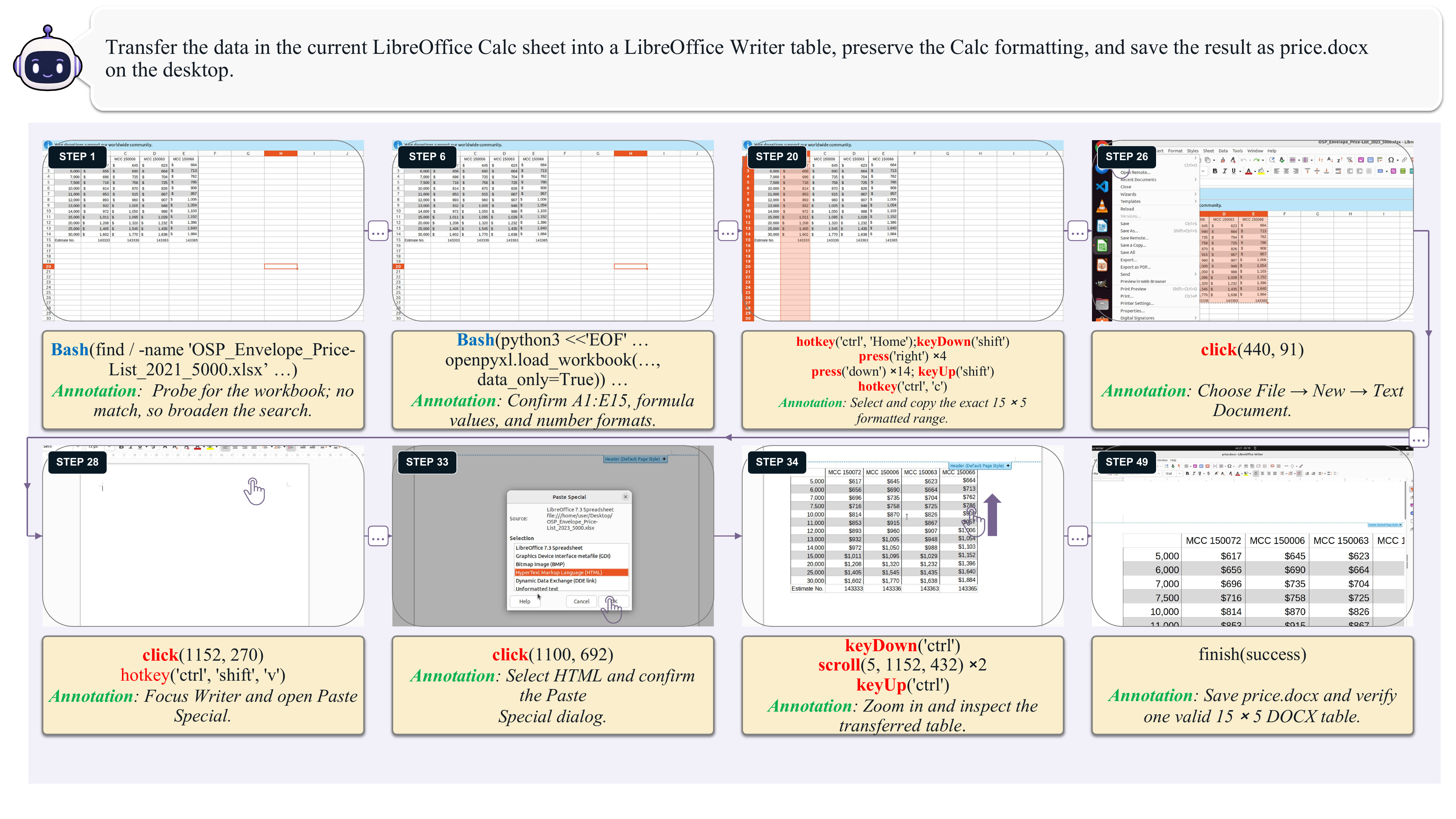}

    \caption{\textbf{Demonstration of hybrid GUI+CLI execution in computer-use tasks.} 
In this example, the agent uses CLI tools to inspect spreadsheet contents and interleaves GUI actions, including batched hotkey sequences and Paste Special, to transfer the formatted data into a Writer document.}
    \label{fig:cua_demo_b}
\end{figure}

\begin{figure*}[t]
    \centering
    \includegraphics[width=0.8\textwidth]{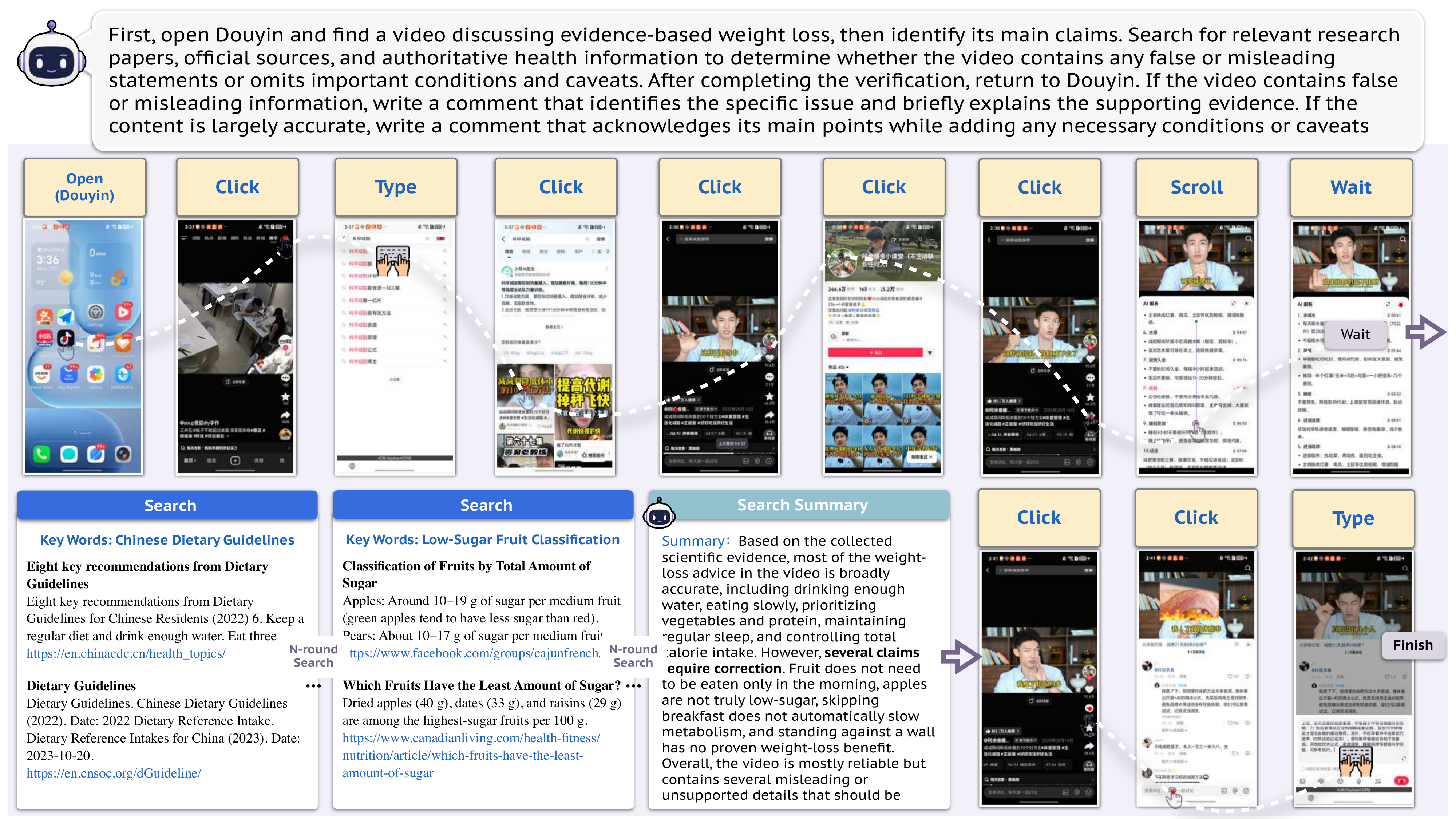}
    \caption{\textbf{Demonstration of DeepSearch invoked on demand during GUI execution.}
    \modelname{} first locates a Douyin video on evidence-based weight loss and
    extracts its main claims through GUI actions, then calls DeepSearch
    mid-trajectory to verify these claims against research papers and
    authoritative health sources, supplying external evidence that the
    interface itself does not provide. The verification conclusions directly
    determine the subsequent GUI actions, and the agent returns to Douyin to
    post an evidence-grounded comment, closing an
    observe--research--decide--act loop.}
    \label{fig:deepsearch_to_gui2}
\end{figure*}

\begin{figure*}[t]
    \centering

    \includegraphics[width=\textwidth]{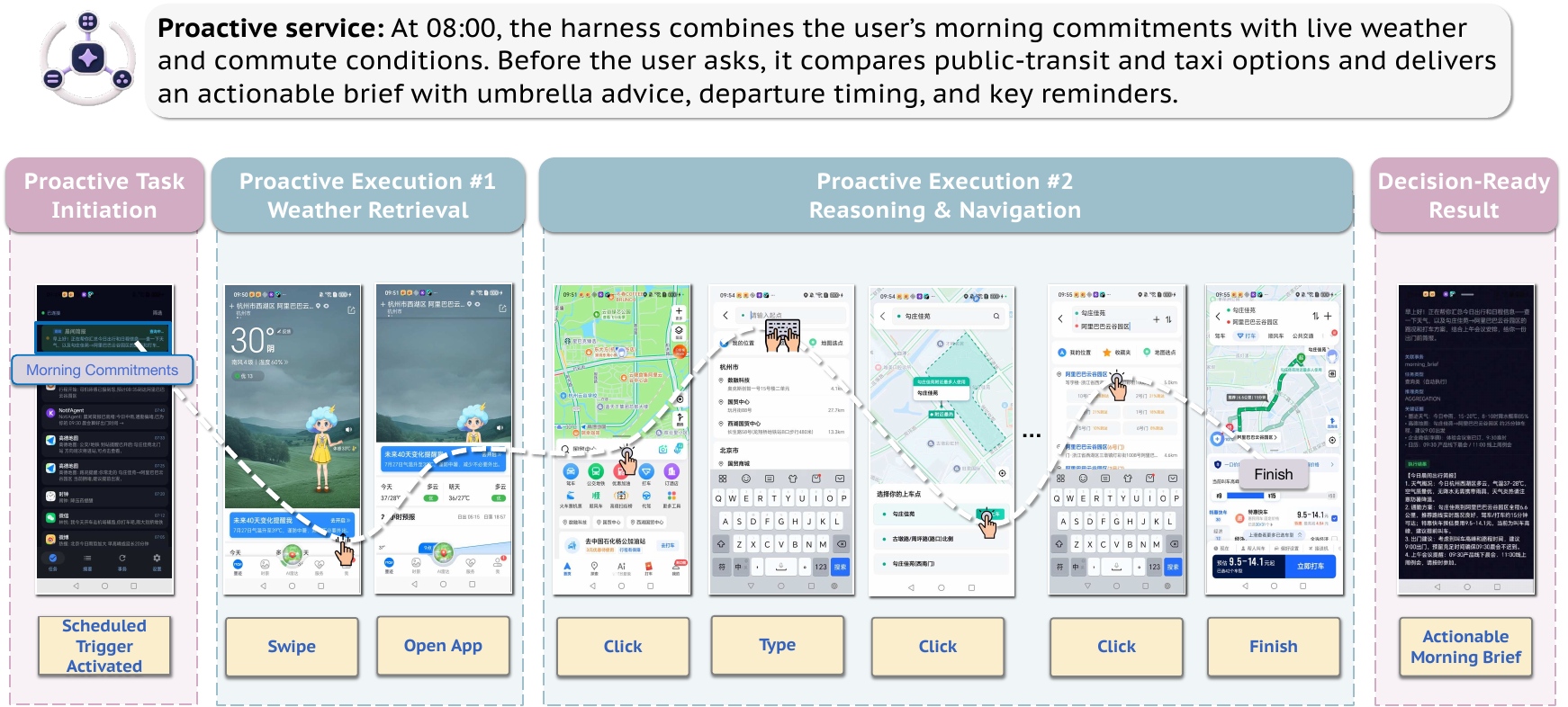}
    \caption{\textbf{Demonstration of proactive service based on mobile notifications.} The trajectory is organized into highlighted stages: proactive task initiation from a detected notification, proactive execution phases, and a decision-ready result, with the executed action annotated beneath each key frame. A scheduled morning brief: at 08:00 the harness combines the user's morning commitments with live weather and commute conditions, compares public-transit and taxi options before the user asks, and delivers an actionable brief with umbrella advice, departure timing, and key reminders.}
    \label{fig:proactive_demo_b}
\end{figure*}

\begin{figure*}[t]
    \centering

    \includegraphics[width=\textwidth]{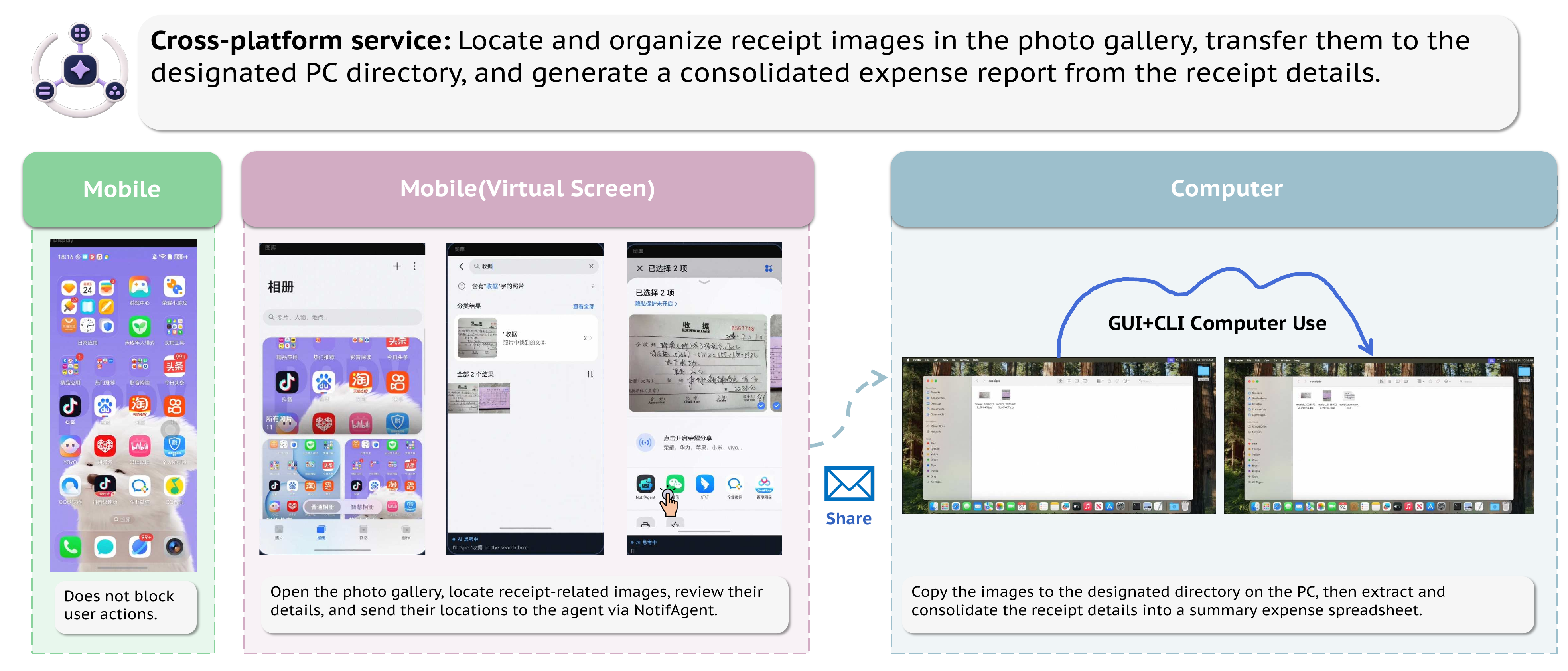}

    \caption{\textbf{Demonstrations of cross-platform task execution.} In this workflow, mobile subtasks run on virtual screens of the physical device, so execution does not block the user's own actions. Receipt organization across devices: the agent locates receipt images in the photo gallery on a mobile virtual screen, transfers them to the designated PC directory, and uses GUI+CLI computer use to consolidate the receipt details into a summary expense spreadsheet.}
    \label{fig:crossplatform_demo_b}
\end{figure*}

\end{document}